\documentclass[journal]{IEEEtran} 
\usepackage{graphicx}
\usepackage{graphics} % for pdf, bitmapped graphics files
\usepackage{epsfig} % for postscript graphics files
\usepackage{booktabs}
\usepackage{array}
\usepackage{times} % assumes new font selection scheme installed
\usepackage{amsmath} % assumes amsmath package installed
\usepackage{amssymb}  % assumes amsmath package installed
\usepackage{enumerate}
\usepackage{mathrsfs}
\usepackage{multirow}
\usepackage{subfigure}
\usepackage{latexsym}
\usepackage{bm}
\usepackage{dsfont}
\usepackage{multicol}
\usepackage{multirow}
\usepackage{color}
\usepackage{url}
\usepackage{cite}
\usepackage{algorithmic}
\usepackage{soul}
\usepackage[ruled]{algorithm2e}
\usepackage{comment}
\usepackage{picinpar}
\usepackage{flushend}
\usepackage{colortbl}
\usepackage{soul}
\usepackage{multirow}
\usepackage{pifont}
\usepackage{alltt}
\usepackage[hidelinks]{hyperref}
\usepackage{siunitx}
\usepackage{breakurl}
\usepackage{pbox}
\usepackage{footmisc} 
\usepackage{threeparttable}
\def\BibTeX{{\rm B\kern-.05em{\sc i\kern-.025em b}\kern-.08em
		T\kern-.1667em\lower.7ex\hbox{E}\kern-.125emX}}
{Xinglong Zhang \MakeLowercase{\textit{et al.}}: Toward Scalable Multirobot control: Fast Policy Learning in Distributed MPC}
\newtheorem{remark}{Remark}
\newtheorem{assumption}{Assumption}

\newtheorem{definition}{Definition}

\newtheorem{theorem}{Theorem}

{\renewcommand{\arraystretch}{2.0}
	\begin{bmatrix}}%
	{\end{bmatrix}
	\renewcommand{\arraystretch}{1.2}}

\def\build#1_#2^#3{\mathrel{\mathop{\kern0pt#1}\limits_{#2}^{#3}}}%
\def\argmin#1{\build{\rm argmin}_{#1}^{}}

\def\build#1_#2^#3{\mathrel{\mathop{\kern0pt#1}\limits_{#2}^{#3}}}%

\makeatletter
\renewcommand*\env@matrix[1][*\c@MaxMatrixCols c]{%
	\hskip -\arraycolsep
	\let\@ifnextchar\new@ifnextchar
	\array{#1}}
\makeatother

\begin{document}
	
	\title{Toward Scalable Multirobot Control: \\Fast Policy Learning in Distributed MPC}
	\author{
		%\vskip 1em		
		Xinglong Zhang, Wei Pan,  Cong Li, Xin Xu, Xiangke Wang,  Ronghua Zhang,  Dewen Hu
		\thanks{
			%
			%Manuscript received Month xx, 2xxx; revised Month xx, xxxx; accepted Month x, xxxx.
			%The work was supported by the National Natural Science Foundation of China under Grant  61825305, 62003361.	
			
			Xinglong Zhang, Cong Li, Xin Xu, Xiangke Wang,  Ronghua Zhang, and Dewen Hu are with the College of Intelligence Science and Technology, National University of Defense Technology, Changsha, 410073, China. (e-mail: \{zhangxinglong18, lc, xinxu,  xkwang, zhangronghua19, dwhu\}@nudt.edu.cn) 
			
			Wei Pan is with the Department of Cognitive Robotics, Delft University of Technology, The Netherlands (e-mail: panweihit@gmail.com). 
			
			Cong Li was with the Chair of Automatic Control
			Engineering, Technical University of Munich, Munich, 80333, Germany. 
			
			Corresponding author: Xinglong Zhang, Xin Xu.
			
			The paper will appear in the IEEE Transactions on Robotics. Personal use of this material is permitted.  Permission from IEEE must be obtained for all other uses, in any current or future media, including reprinting/republishing this material for advertising or promotional purposes, creating new collective works, for resale or redistribution to servers or lists, or reuse of any copyrighted component of this work in other works.
			
		}
	}
	
	\maketitle
	\vspace{-8mm}	
	\begin{abstract}
		
		Distributed model predictive control (DMPC) is promising in achieving optimal cooperative control in multirobot systems (MRS). However, real-time DMPC implementation relies on numerical optimization tools to periodically calculate local control sequences online. This process is computationally demanding and lacks scalability for large-scale, nonlinear MRS. This article proposes a novel distributed learning-based predictive control (DLPC) framework for scalable multirobot control. Unlike conventional DMPC methods that calculate open-loop control sequences, our approach centers around a computationally fast and efficient distributed policy learning algorithm that generates explicit closed-loop DMPC policies for MRS without using numerical solvers. The policy learning is executed incrementally and forward in time in each prediction interval through an online distributed actor-critic implementation. The control policies are successively updated in a receding-horizon manner, enabling fast and efficient policy learning with the closed-loop stability guarantee. The learned control policies could be deployed online to MRS with varying robot scales, enhancing scalability and transferability for large-scale MRS. Furthermore, we extend our methodology to address the multirobot safe learning challenge through a force field-inspired policy learning approach. We validate our approach's effectiveness, scalability, and efficiency through extensive experiments on cooperative tasks of large-scale wheeled robots and multirotor drones. Our results demonstrate the rapid learning and deployment of DMPC policies for MRS with scales up to 10,000 units. Source codes and multimedia materials are available online at \href{https://sites.google.com/view/pl-dpc/}{https://sites.google.com/view/pl-dpc/}.		
	\end{abstract}
	
	%\markboth{IEEE TRANSACTIONS ON ROBOTICS}%
	{}
	
	\IEEEdisplaynontitleabstractindextext
	\IEEEpeerreviewmaketitle
	
	\section{Introduction}
	
	Multirobot systems (MRS) represent a collective of autonomous robots interconnected through communication networks~\cite{quan2023robust}, enabling collaborative control tasks. This networked structure endows MRS with the capability to pursue global objectives that exceed the capabilities of individual robots. However, achieving optimal coordination in MRS often involves optimizing a global performance index, which poses a significant large-scale optimal control problem~\cite{Xuan2023}. 
	Centralized solutions to the above problems may struggle to adequately address the complexities arising from interactions between multiple robots~\cite{farina2018hierarchical}. 
	Consequently, recent decades have witnessed considerable attention in developing distributed optimal control approaches for MRS~\cite{hujinwen2020,Fathian2021,liu2018,ren2020,Santilli2022,Qingbiao,Liangming}.
	Among them, distributed model predictive control (distributed MPC, DMPC) is a primary methodology for multirobot control under constraints~\cite{hou2021distributed,todorovic2020distributed,shen2020distributed,7546918,navsalkar2023data}, formulating control problems as optimization tasks over prediction horizons to achieve optimized performance.

	%\begin{figure}[t!]
	%	\centerline{\includegraphics[scale=0.42]{figure/jiqun}}
	%	\caption{ Fast policy generation for robot swarm. Our approach online learns DMPC policies for MRS with scales up to 10,000, and the learned policy is online deployed to MRS with varying robot scales up to 1,000.}
	%	\label{fig:Communication diagram0} 
	%\end{figure}
	
	{In DMPC, each robot calculates the local control sequences online by solving the optimization problem with numerical solvers~\cite{xinglongzhang2022_robust,Ferranti,Wei2021,el2015distributed}, which could be computationally intensive for nonlinear MRS.  In real-world applications such as small-size mobile robots, the computational efficiency and real-time performance optimization hinge on several influencing factors: i) Onboard computing resources are inherently limited in scale and processing capability; ii) The nonlinear dynamics and complex interactions among robots compound the computational load. %In many real-world applications, the computational efficiency and real-time optimization hinge on several influencing factors: i) The onboard computing resources are necessarily small-scale and have low processing capability; ii) the robots' dynamics and their mutual interactions are nonlinear and probably large-scale. 
		%face challenges in computational efficiency and real-time optimization due to several factors
		Consequently, %balancing the trade-off between computational complexity and real-time performance optimization poses a challenge
		the real-time resolution of large-scale DMPC problems presents significant challenges~\cite{Ferranti,Wei2021}, deemed inapplicable for real-world large-scale yet small-sized MRS. This motivates us to propose a computationally fast and efficient policy learning approach to generate explicit closed-loop DMPC policies, rather than calculating the open-loop control sequences, i.e., implicit policies, with numerical solvers. To our knowledge, no previous work has developed policy learning techniques for designing the DMPC policies.}
	
	As a class of policy learning techniques, reinforcement learning (RL) has made significant progress for robot control (cf. \cite{Dieter2022,wei2020continuous,odekunle2020reinforcement,Han2022RL}). RL enables the acquisition of control policies directly from data~\cite{Dieter2022}, or through model predictions~\cite{wei2020continuous}, to improve sample efficiency. 
	In the context of multirobot control, numerous approaches have been proposed leveraging multiagent RL paradigms~\cite{Sartoretti2019,wang2018model,Yang2022databased}. Despite the prevalence of deep RL frameworks, such as asynchronous advantage actor-critic~\cite{Sartoretti2019}, challenges persist in training scalability, sample efficiency, and the absence of closed-loop guarantees in policy learning.
	These challenges highlight the crucial need for scalable policy learning with stability guarantees. Our approach achieves this goal from two perspectives. First, we design a distributed online actor-critic learning algorithm in which the training process is executed incrementally to generate control policies efficiently. Second, we introduce a policy training approach grounded in control theory, integrating the receding horizon optimization strategy into policy updates. This ensures closed-loop stability and improves learning efficiency.
	
	In addition to guaranteed stability, safety constraints, such as collision avoidance, must be met to ensure the persistent and reliable operation of MRS. However, ensuring control safety in RL remains a nontrivial task, even in the model-based scenario~\cite{emam2021safe,cheng2019end,annurev-control-042920-020211,Motoya8746143}. Recent efforts in safe learning control have focused primarily on the centralized control structure~\cite{annurev-control-042920-020211}.  Few works have been dedicated to the safe multiagent policy learning~\cite{lu2021distributed,Zhang9812083}, drawing inspiration from cost-shaping-based RL \cite{emam2021safe,cheng2019end}. However, the cost-shaping design would lead to weight divergence in the actor-critic framework~\cite{Paternain9718160}, which lacks online learning ability and safety guarantees. Gaining insights from interior point optimization~\cite{boyd2004convex}, we design a novel safe policy learning algorithm with a force field-inspired policy structure. This design can balance the objective and constraint-associated forces acting on the MRS during policy learning, ensuring safe learning with physical interpretations.
	
	%In this article, we propose a distributed learning-based predictive control (DLPC) framework for scalable performance optimization in MRS. First, a distributed fast policy learning algorithm is designed to generate closed-loop DMPC policies. Specifically, the optimization problem over each prediction horizon is decomposed into several sequential subproblems and solved through policy learning. 
	%The control policy for each robot is a parameterized function of the neighbor states, which are updated incrementally and forward in time with an efficient distributed actor-critic implementation. Compared with numerical DMPC, our approach significantly improves computational efficiency through fast policy learning and facilitates scalability and rapid adaptability in large-scale MRS via agile policy deployment. %Second, we extend the proposed approach to address the safe policy learning challenge through a novel force field-inspired policy design. We design a force field-inspired control structure to balance the objective force and constraint force acting on the MRS in the policy learning process.
	%	We test our approach through extensive simulations and real-world experiments on the formation control of wheeled mobile robots and multirotor drones.
	The contributions of this article are summarized below.
	
	\begin{enumerate}[i)]
		\item We propose a novel distributed learning-based predictive control (DLPC) framework for large-scale MRS. In contrast to conventional numerical DMPC, which calculates open-loop control sequences, our approach generates closed-loop DMPC policies without relying on numerical solvers. %In contrast to conventional numerical DMPC methods that  calculate the open-loop control sequences, our approach generates closed-loop DMPC policies without using numerical solvers. 
		The optimization problem of DMPC within each prediction interval is decomposed into several sequential subproblems and solved by policy learning. 
		The control policies are composed of parameterized functions capable of online learning and deployment to scenarios with varying robot scales, enhancing scalability and transferability for large-scale MRS. 
		
		\item  A computationally fast and efficient distributed policy learning algorithm is developed, integrating the receding horizon optimization strategy into policy updates.  In each prediction interval, policy learning is executed forward in time rather than backward in time with a distributed incremental actor-critic implementation, enabling fast online policy updates. The control policies generated from each prediction interval are successively refined in subsequent intervals to improve learning efficiency, fundamentally different from the common independent problem-solving paradigm of DMPC in different prediction intervals. %The control policies are successively updated in a receding horizon manner, enabling fast and efficient policy learning with closed-loop stability guarantee. Compared with numerical DMPC, 
		%As a result, our approach significantly improves computational efficiency through fast policy learning, and facilitates scalability and rapid adaptability in large-scale MRS via agile policy deployment.
		Compared with numerical DMPC, our approach significantly reduces the computational load through fast and efficient policy learning while maintaining closed-loop stability.
		\item  We further address the challenge of safe policy learning in MRS through a force field-inspired policy design, which has clear physical interpretations to balance the objective force and the constraint force acting in MRS, enabling online policy learning and policy deployment with safety and robustness guarantees.
		
		\item 	We numerically and experimentally validate our method's superior sim-to-real transferability and scalability in large-scale multirobot control. 
		%In particular, we have shown for the first time that our approach efficiently learns near-optimal formation control policies for MRS with scales up to 10,000. 
		In particular, we have shown on different computing platforms, i.e., a laptop and a Raspberry PI 5, that our approach efficiently learns near-optimal formation policies for MRS with scales up to 10,000, and the computational load grows linearly with robot scales in both platforms.  To our knowledge, no optimization-based control approach has realized distributed control on such a large scale. Moreover, the policy trained with two robots is well deployed to robots with scales up to 1,000.

		%            different computing platforms, i.e., a Laptop and a Raspberry PI 5, that our approach could efficiently learn the DMPC policies for MRS with scales up to 10,000. 
		%the preservation and transformation of formation configurations, along with
		
	\end{enumerate}

	This article is a novel development of our previous conference work~\cite{9811604}. In this article, we design a fast policy learning approach toward scalable multirobot control, which is beyond the scope of our previous work~\cite{9811604}. Hence, the detailed techniques, theoretical insights, and experimental validations presented here differ substantially from that in~\cite {9811604}.  %we additionally: i) introduce control constraints into the problem formulation for practical concerns, resulting in a new policy structure and learning algorithm; ii) develop a novel learning control extension by including proper designs on the terminal cost and constraint in each prediction interval; iii) prove the closed-loop theoretical properties of the algorithm in both nominal and perturbed scenarios; iv) design a distributed dual heuristic programming (DHP) method to accelerate the convergence of actor-critic learning; v) demonstrate that our control policy can be transferred to a greater number of MRS without fine-tuning and evaluate our approach in more sophisticated real-world control scenarios.
	
	%The article is structured as follows.  Section II reviews the related work on DMPC and multiagent RL. In Section III, we present the dynamical models of MRS and the formulations of the distributed DMPC problem. The proposed policy learning framework for DMPC is presented in Section IV, while Section V derives the extension to distributed safe learning for DMPC. Section VI demonstrates the simulated and real-world experimental results on the formation control of wheeled robots and multirotor drones. Conclusions are drawn in Section VII. Additional simulation results and the theoretical analysis are given in the Appendix.
	
	The article is structured as follows.  Section II reviews the related work. Section III presents the dynamical models of MRS and the formulation of the DMPC problem. The proposed policy learning framework for DMPC is presented in Section IV, while Section V derives the extension to safe policy learning. Section VI demonstrates the simulated and experimental results. Conclusions are drawn in Section VII. The main theoretical and auxiliary numerical results are given in Appendix~\ref{sec:32}, while additional theoretical results for safe policy learning are referred to in the attached materials.
	
	\textbf{Notation:}  We use $\mathbb{R}$ and $\mathbb{R}^{+}$ to denote the sets of real numbers and positive real numbers, respectively;
	$\mathbb{R}^{n}$ to denote the Euclidean space of the $n$-dimensional real vector;
	$\mathbb{R}^{n \times m}$ to denote the Euclidean space of $n \times m$ real matrices. Denote $\mathbb{N}$ as the set of integers and denote $\mathbb{N}_{l_1}^{l_2}$ as the set of integers $l_1,l_1+1,\cdots,l_2$. For a group of vectors $z_i\in\mathbb{R}^{n_i}$, $i\in\mathbb{N}_{1}^{M}$, we use $\text{col}_{i\in\mathbb{N}_1^M}(z_i)$ or $(z_1,\cdots,z_M)$ to denote $[z_1^{\top},\cdots,z_M^{\top}]^{\top}$, where $M\in\mathbb{N}$. We use $\bm{u}(k)$  to represent a control policy formed by the control sequence $u(k),\cdots,u(k+N-1)$, where $k,\,N\in\mathbb{N}$. For a general function $h(z(k))$ in a variable $z(k)$, we use $h(k)$ to represent $h(z(k))$ for simplicity. Given a function $f(x)$ with argument $x$, we define $\triangledown f(x)$ and $\triangledown^2 f(x)$  as the gradient and Hessian to $x$, respectively. We use ${\rm {Int}}(\mathcal{Z}_i)$ to represent the interior of the set $\mathcal{Z}_i$. For a matrix $P\in\mathbb{R}^{n\times n}$, $P\succ 0$ means that it is positive definite. Given two general sets $\mathcal{A}$ and $\mathcal{B}$, the pontryagin difference of $\mathcal{A}$ and $\mathcal{B}$ is denoted as $\mathcal{A}\ominus\mathcal{B}=\{c|c+b\in \mathcal{A},\, \forall b\in\mathcal{B}\}$. For a vector $x\in\mathbb{R}^{n}$, we denote $\|x\|_Q^2$ as $x^{\top}Qx$ and $\|x\|$ as the Euclidean norm.
	
	{\section{Related Work}
		% \emph{DMPC for linear MRS:} 
		%	Several DMPC approaches have been proposed for linear MRS. Among them, a fast dual gradient-based DMPC approach was proposed in~\cite{wang2018accelerated} for discrete-time linear systems with coupled constraints.
		%A flexible DMPC approach was developed in~\cite{Giuseppe2021} for leader-follower formation control, incorporating collision avoidance through time-varying topology reconfiguration. A distributed MPC algorithm for MRS' motion planning problem was investigated in~\cite{Carlos2020}, where limited information exchanges among local robots allowed mutual collision avoidance.   In~\cite{Zhou2019}, the MRS' motion planning problem was decomposed into several independent MPC problems using the neighbor's rough state prediction, significantly simplifying the optimization problem. However, these approaches assume linear system dynamics, which may lead to suboptimality or degraded performance in real-world scenarios with nonlinear dynamics, as noted in~\cite{Saeed2019}. 
		
		\emph{Nonlinear DMPC:} Numerous DMPC approaches have been proposed for nonlinear MRS, and the most relevant ones are discussed here. In particular, {\color{black}a DMPC approach under unidirectional communication topologies was designed in~\cite{7546918} for platoon control of intelligent vehicles.} A DMPC algorithm was proposed in \cite{Ferranti} for trajectory optimization of MRS. The cooperative optimization problem was solved by the nonconvex alternating direction method.  In~\cite{Wei2021}, a Lyapunov-based DMPC approach was developed for the formation control of autonomous robots under exogenous disturbances. In addition, a DMPC framework was developed in \cite{el2015distributed} for autonomous vehicles with limitations in communication bandwidth and transmission delays. %However, the stability argument relies on the small-gain condition, which might not hold for strongly coupled subsystems. 
		It should be noted that the above works~\cite{7546918,Ferranti,Wei2021,el2015distributed} resort to nonlinear optimization solvers for online computation. {\color{black} At each time instant, the local controller in numerical DMPC aims to find an optimal numerical solution by optimizing the local performance index over the entire prediction horizon. However, this approach can be computationally intensive for large-scale MRS, particularly when dealing with long prediction horizons and limited onboard computing capabilities. In contrast, our method addresses the local optimization problem incrementally, advancing step by step within each prediction interval through an efficient distributed policy learning approach. Moreover, the learned local control policies can be deployed directly without the need for online retraining or fine-tuning.} 
		
		\emph{Explicit DMPC:} Explicit MPC was initially proposed in \cite{BEMPORAD20023} to generate explicit control laws for linear systems. It involved offline computation and online deployment of a collection of explicit piecewise control laws. Although explicit MPC can reduce the online computation time, the complexity of offline computation grows exponentially with the system's orders, and the control performance demands model accuracy.
		The extension to explicit DMPC was developed through system-level synthesis in~\cite{alonso2020explicit}. Still, this work is suitable only for linear systems and relies on the separability assumption of systems.  In contrast to~\cite{BEMPORAD20023,alonso2020explicit}, this article learns explicit closed-loop control policies for nonlinear, large-scale MRS. 
		
		%
		%\subsection{The Integration of RL and MPC}\label{sec:rl4dmpc}
		\emph{Integration of RL and MPC:}  %Several approaches have been proposed to integrate RL and MPC~\cite{Li2021learning,Song2022,xu2018learning,xinglongzhang2022_robust}. 	
		As RL can design control policies from data, optimizing the high-level decision variables of MPC is straightforward to improve control performance~\cite{Song2022}. In~\cite{Li2021learning}, an RL algorithm was used to model the maximum entropy as a penalty function in MPC.
		Recent works~\cite{xu2018learning,xinglongzhang2022_robust} incorporated the receding horizon strategy into the RL training process and proposed actor-critic learning algorithms to generate MPC's policies. However, these approaches are centralized in nature and designed for small-scale systems. %In contrast, our work presents a distributed learning control approach for large-scale MRS, addressing challenges such as safety and stability guarantees under distributed actor-critic learning. 
		
		\emph{Multiagent RL (MARL):} Several MARL approaches have emerged for MRS using various policy learning methods,  including policy iteration \cite{jiang2020cooperative,wei2020continuous}, policy gradient \cite{Zhang2021,Yang2022databased}, asynchronous advantage actor-critic \cite{Sartoretti2019}.  Yet, these approaches cannot learn online with stability guarantees. The promising work \cite{bhattacharya2023multiagent} demonstrated, with a decision-making example in discrete space, the potential of multistep lookahead rollout in performance improvement. 
		
		\emph{MARL under safety constraints:} Previous MARL approaches~\cite{lu2021distributed,Zhang9812083} for multirobot collision avoidance utilized potential functions for cost shaping~\cite{emam2021safe,cheng2019end}. However, this design may face weight divergence within the actor-critic framework~\cite{Paternain9718160}. In~\cite{fan2020distributed}, a deep RL approach was proposed to navigate MRS safely, incorporating a hybrid control structure to improve the robustness of policy deployment. An extension to a multiagent RL approach was presented in~\cite{Han2022RL} with a reward design based on reciprocal velocity obstacles. Nonetheless, the development of MARL, with the ability to learn policies online and ensure safety, remains unresolved. We address this challenge through a force field-inspired safe policy design with an efficient actor-critic implementation.
		\section{Control Problem Formulation}
		In this section, we begin by introducing the dynamical models of MRS. Next, we present the formulation of the cooperative DMPC based on preliminary work~\cite{conte2016distributed}. 
		\subsection{Dynamical Models of MRS}
		
		Consider the formation control of $M$ mobile robots with collision avoidance.
		The dynamical model of the $i$-th robot is given as
		\begin{equation}\label{Eqn:kinematic model}
			\dot{q}_i =(v_i\cos \theta _i,v_i\sin \theta _i,\omega _i,a_i),
		\end{equation}
		where $q_i=({p}_{x,i},{p}_{y,i},{\theta}_i,{v}_i)\in\mathbb{R}^{n_i}$, $n_i=4$, $\text{(}p_{x,i},p_{y,i}\text{)}$ is the coordinate of the $i$-th robot in Cartesian frame, $\theta _i$ and $v_i$ are the yaw angle and the linear velocity; $(a_i,\omega_i)\in\mathbb{R}^{m_i}$ with $m_i=2$ are the acceleration and yaw rate. 
		The formation error of the $i$-th robot in the local coordinate frame is defined as 
		\begin{equation}\label{Eqn:formation error model}
			\begin{array}{ll}
				e_i =&T_i(\sum_{j=1}^Mc_{ij}\left(\Lambda_1(q_j-q_i)+\varDelta h_{ji} ) +\right.\\
				&\left.s_i(\Lambda_1(q_r -q_i)+\varDelta h_{ri})+\Lambda_2(q_r-q_i)\right),
			\end{array}
		\end{equation}	
		where $c_{ij}$ represents the connection status, $c_{ij}=1$ for $j\in\mathcal{N}_i$ and $c_{ij}=0$ otherwise, $\mathcal{N}_i$ is the set of all neighbors of robot $i$ (including robot $i$ itself); $s_{i}$ represents the pinning gain, $s_{i}=1$ if the robot $i$ receives the position information of the leader, $\Lambda_1=\text{diag}\{1,1,0,0\}$, $\Lambda_2=\text{diag}\{0,0,1,1\}$, $q_r$ is the reference state received from the leader. The last term in~\eqref{Eqn:formation error model} is used for guiding the consensus of linear velocity and yaw angle of each robot;  $\varDelta h_{ji}$ and $\varDelta h_{ri} $  are coordinate correction variables, which are determined by the formation shape and size; the coordinate transformation matrix is
		\begin{equation*}%\label{Eqn:T}
			T_i=\left[ \begin{matrix}
				\cos \theta _i&		\sin \theta _i&		0\\
				-\sin \theta _i&		\cos \theta _i&		0\\
				0&		0&			I_2\\
			\end{matrix} \right]\in\mathbb{R}^{n_i\times n_i}. 
		\end{equation*}
		
		Let $u_i=(w_r-w_i,a_r-a_i)$ be the control input associated with robot $i$, where $w_r,\,a_r$ are the reference acceleration and yaw rate received from the leader, and denote $e_{\scriptscriptstyle \mathcal{N}_i}\in\mathbb{R}^{n_{\mathcal{N}_i}}$ as the collection of all neighboring error states (including $e_i$), i.e., $e_{\scriptscriptstyle \mathcal{N}_i}={\rm col}_{j\in{\scriptscriptstyle\mathcal{N}_i}}e_j$. By discretizing~\eqref{Eqn:kinematic model} under~\eqref{Eqn:formation error model} over a sampling interval $\varDelta t$, we write the local formation error model for the $i$-th robot as an input-affine form through a straightforward derivation process deferred in Appendix~\ref{sec:model}. The concise form follows:
		\begin{equation}\label{Eqn:LL-nei}
			\begin{array}{ll}
				e_{i}(k+1)=f_{i}(e_{\scriptscriptstyle \mathcal{N}_i}(k))+g_{i}(e_i(k))u_{i}(k), \ \ \ i\in\mathbb{N}_1^M,
			\end{array}
		\end{equation}
		where $k\in\mathbb{N}$ is the discrete-time index, the mappings $f_{i}: \mathbb{R}^{n_{\scriptscriptstyle \mathcal{N}_i}}\rightarrow\mathbb{R}^{n_i}$ and $g_{i}: \mathbb{R}^{n_i}\rightarrow \mathbb{R}^{n_i\times m_i}$ are smooth state transition and input mapping functions respectively, and $f_{i}(0)=0$; $e_{\scriptscriptstyle \mathcal{N}_i}\in\mathcal{E}_i\subseteq{\mathbb{R}}^{n_{\scriptscriptstyle \mathcal{N}_i}}$ and $u_{i}\in\mathcal{U}_i\subseteq {\mathbb{R}}^{m_{i}}$,  the sets $\mathcal{E}_i$ and $\mathcal{U}_i$ are
		\begin{equation}\label{eqn:constraints}
			\begin{array}{ll}
				\mathcal{E}_i=\{e_{\scriptscriptstyle \mathcal{N}_i}\in\mathbb{R}^{n_{\scriptscriptstyle \mathcal{N}_i}}|{\Xi_{e,i}^j}(e_{\scriptscriptstyle \mathcal{N}_i})\leq 0, j=1,\dots,n_{e,i}\}\\
				\mathcal{U}_i=\{u_i\in\mathbb{R}^{m_i}|{\Xi_{u,i}^j}(u_i)\leq 0, j=1,\dots,n_{u,i}\},
			\end{array}
		\end{equation} where $\Xi_{e,i}^j(e_{\scriptscriptstyle \mathcal{N}_i}),\,\Xi_{u,i}^j(u_i)\in\mathbb{R}$ are $C^1$ functions,   $n_{e,i},\, n_{u,i}\in\mathbb{N}$ denote the overall numbers of inequalities associated with $\Xi_{e,i}^j(e_{\scriptscriptstyle \mathcal{N}_i}),\, \Xi_{u,i}^j(u_i)$, respectively.  Note that the set $\mathcal{E}_i$ is a versatile formulation representing a range of constraints, including static/dynamic collision avoidance and joint inter-robot collision avoidance, which will be discussed in Section~\ref{sec:experiment}.
		%
		%In principle, the state constraint $\mathcal{E}_i$ of robot $i$ can be formalized to represent constraints of different types as follows. For instance, (i) $\mathcal{E}_i$ with $\Xi_{x,i}^j(e_i)=E_{x,i}^je_i$ is a linear convex set,  where $E_{x,i}^j\in\mathbb{R}^{1\times n_i}$; (ii) $\mathcal{E}_i$ with $\Xi_{x,i}^j(e_i)=d^j_i-\|E^j_{x,i}e_i-c^j_i\|$ represents a nonconvex set for collision avoidance in a 2-D map, where in this case $E_i^j\in\mathbb{R}^{2\times n_i}$,  $c_i^j\in\mathbb{R}^2$ and $d_i^j\in\mathbb{R}$ are the geometric central point and radius of the obstacle, respectively.
		%
		% To derive closed-loop theoretical guarantees, we assume the overall state and control sets, i.e., $\mathcal{E}=\mathcal{E}_1\times\cdots\times\mathcal{E}_M$ and $\mathcal{U}=\mathcal{U}_1\times\cdots\times\mathcal{U}_M$, are convex and contain the origin in the interiors.  %For a candidate convexification algorithm of nonconvex sets, please refer to~\cite{liu2018convex}. 
		% A discussion on the direct application of our approach under nonconvex state constraints will be deferred in Remark~\ref{rem:nonconvex}.
		
		Collecting all the local robot systems from~\eqref{Eqn:LL-nei}, the overall dynamical model is written as 
		\begin{equation}\label{Eqn:C full model} e(k+1)=F_c(e(k))+G_c(e(k))u(k),
		\end{equation}
		and will be used later for closed-loop stability analysis, where 
		$e=\text{col}_{i\in\mathbb{N}_1^M}(e_i)\in\mathbb{R}^{n}$ is the overall state variable, $n=\sum_{i=1}^{M}n_{i}$, $u=\text{col}_{i\in\mathbb{N}_1^M}(u_i)\in\mathbb{R}^{m}$, $m=\sum_{i=1}^{M}m_{i}$, $F_c=\text{col}_{i\in\mathbb{N}_1^M}(f_i)$, the diagonal blocks of $G_c$
		are $g_{i}$, $i\in\mathbb{N}_1^M$. 

		We introduce the following standard assumption~\cite{conte2016distributed} with respect to the stabilizability of~\eqref{Eqn:C full model}.
		\begin{assumption}[Stabilizing control]\label{assum:stabilizing_control}
			There exist local feedback control policies $u_i(e_{\scriptscriptstyle\mathcal{N}_i})$  for all $i\in\mathbb{N}_1^M$, such that $u={\rm col}_{i\in\mathbb{N}_1^M}(u_i(e_{\scriptscriptstyle\mathcal{N}_i}))$ is a stabilizing control policy of~\eqref{Eqn:C full model}. 
		\end{assumption}
		
		{\color{black}The satisfaction of the above stabilizability condition does not impose restrictive requirements on the communication networks. Indeed, information exchanges among neighboring robots can be bidirectional or unidirectional, provided that a stabilizing control policy for~\eqref{Eqn:C full model} exists. Since our work focuses on designing a fast policy learning algorithm for DMPC, we also introduce a standard assumption on the communication network.
			\begin{assumption}[Communication network]\label{assum:communication}
				The communication network is time-invariant and delay-free, meaning that the connections between neighboring robots remain fixed and the information is exchanged without latency.
		\end{assumption}}		
		% \begin{comment}
			% \begin{remark} In principle, the above control problem can be solved with a decentralized control structure. In this setting, the design of local controllers is trivial since the state interactions among robots are disregarded. However, when the state interactions among robots are strong~\cite{scattolini2009architectures}, the decentralized control approach might not derive guaranteed performance and/or stability, especially for systems with the so-called {\tt{fixed mode}}, e.g., uncontrollable mode~\cite{davison1987decentralized}. Hence, we adopt a distributed control structure to solve the control problem of MRS.
				% 	 \hfill $\blacktriangle$
				% \end{remark}
			% \end{comment}
		\subsection{Distributed MPC for MRS}
		We follow a notable cooperative DMPC formulation~\cite{conte2016distributed}  for the optimal control of MRS. At each time step $k$, the following finite-horizon cooperative optimization cost is to be minimized:
		\begin{equation}\label{Eqn:optimiz}
			\begin{array}{cc}
				\min & {J(e(k))},\\
				{\bm u_i(k), \forall i\in\mathbb{N}_1^M}\\
			\end{array}
		\end{equation}
		where $\bm u_i(k)=u_i(k),\cdots,u_i(k+N-1)$, the global cost $J(e(k))=\sum_{i=1}^MJ_i(e_{\scriptscriptstyle \mathcal{N}_i}(k))$ with the local cost associated with each robot $i$ defined as
		\begin{equation}
			\begin{array}{ll}
				J_i(e_{\scriptscriptstyle \mathcal{N}_i}(k))=\vspace{1mm}\\\sum_{j=0}^{N-1}r_i(e_{\scriptscriptstyle \mathcal{N}_i}(k+j),u_i(k+j)) +\| e_i(k+N)\|_{P_i}^{2},
			\end{array}\label{Eqn:HL_cost}
		\end{equation}	
		wherein the stage cost %$r_i( e_{\scriptscriptstyle \mathcal{N}_i}(k), u_i(k))$ is of type: 
		$r_i(e_{\scriptscriptstyle \mathcal{N}_i}(k),u_i(k))=\| e_{\scriptscriptstyle \mathcal{N}_i}(k)\|_{Q_i}^{2}+\| u_i(k)\|_{R_i}^{2},$
		%and where $\delta e_i=e_i-e_{s,\scriptscriptstyle \mathcal{N}_i}$, $\delta u_i=u_i-u_{s,i}$, 
		$N\in\mathbb{N}$ is the prediction horizon, $Q_i=Q_i^{\top}\in\mathbb{R}^{n_{\scriptscriptstyle \mathcal{N}_i}\times n_{\scriptscriptstyle \mathcal{N}_i}}$, $Q_i\succ 0$, $R_i=R_i^{\top}\in\mathbb{R}^{m_i\times m_i}$, $R_i\succ 0$; $P_i=P_i^{\top}\in\mathbb{R}^{n_i\times n_i}$, $P_i\succ 0$ is the terminal penalty matrix. 
		
		At each time step $k$, the optimization problem \eqref{Eqn:optimiz} is usually solved using numerical optimization tools ~\cite{conte2016distributed,yang2019survey} and subject to model~\eqref{Eqn:LL-nei}, constraints
		$e_i(k+j)\in \mathcal{E}_i$, $u_i(k+j)\in \mathcal{U}_i$, and the terminal state constraints
		$e_i(k+N)\in \mathcal{E}_{f,i},\ \forall \, i\in\mathbb{N}_{1}^M,\, j\in\mathbb{N}_0^{N-1}$, where $\mathcal{E}_{f,i}$ can be computed as a local control invariant set of~\eqref{Eqn:LL-nei} (if exists) in the form $\mathcal{E}_{f,i}=\{e_i\in\mathbb{R}^{n_i}|e_i^{\top}S_ie_i\leq 1\}$, $S_i=S_i^{\top}\succ 0$. 
		%	
		%Since $P_i$ and $\mathcal{E}_{f,i}$, $i\in\mathbb{N}_1^M$ are decoupled among the local robots, problem \eqref{Eqn:optimiz} with~\eqref{Eqn:HL_cost} can be solved in a distributed manner via numerical optimization tools, see~\cite{conte2016distributed,yang2019survey}. As such, the local open-loop control sequence can be computed for each robot in each prediction interval. Only the first control action is applied, and problem \eqref{Eqn:optimiz} is solved repeatedly at the next instant.
		
		\begin{figure}[t!]
			\centerline{\includegraphics[scale=0.3]{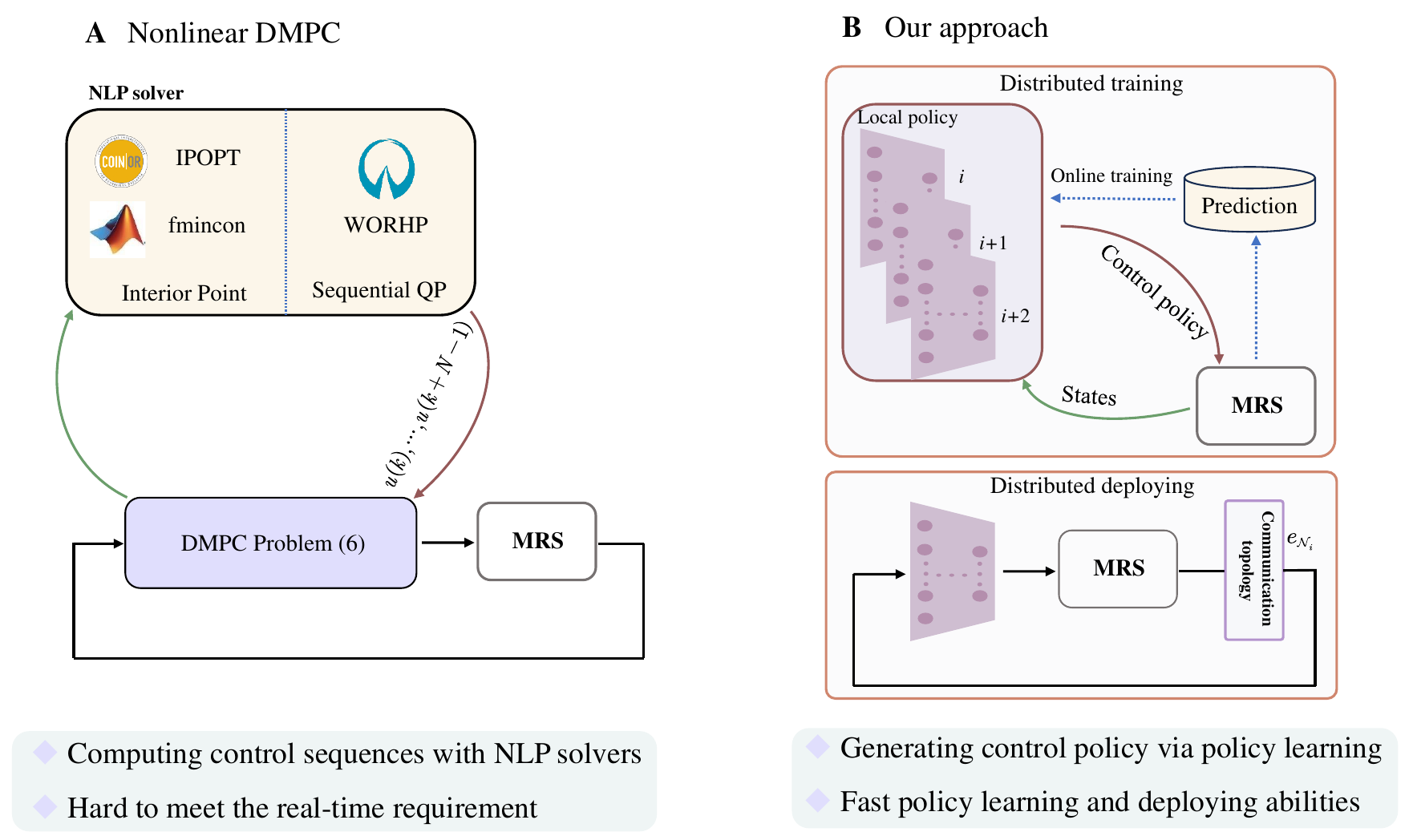}}
			\caption{
				The motivational problem. A: In nonlinear DMPC, the optimization problems are usually solved through nonlinear programming (NLP) solvers, which are computationally intensive and non-scalable, especially for nonlinear MRS with large scales. B: Our approach generates the closed-loop DMPC policies for MRS through distributed policy learning, and the learned policies are composed of parameterized functions that could be online trained and deployed with robot scales up to 10,000. }
			\label{fig:motivating_pro} 
		\end{figure}
		
		At each time instant $k$, solving the problem \eqref{Eqn:optimiz} generates an optimal control sequence. Only the first control action is applied, and \eqref{Eqn:optimiz} is solved repeatedly at the next time instant.
		%At each time instant $k$, problem \eqref{Eqn:optimiz} with~\eqref{Eqn:HL_cost} is solved numerically by distributed optimization tools~\cite{conte2016distributed,yang2019survey} to generate the overall open-loop control sequence ${\bm u}(k)=u(k),\cdots,u(k+N-1)$. Only the first control action is applied to the MRS, and problem \eqref{Eqn:optimiz} is solved repeatedly at the next instant. 
		However, it should be noted that solving~\eqref{Eqn:optimiz} using numerical solvers for nonlinear large-scale MRS is challenging and computationally intensive (see Fig.~\ref{fig:motivating_pro}). Instead of numerically calculating the control sequence ${\bm u}(k)=\text{col}_{i\in\mathbb{N}_1^M}(\bm u_i(k))$, this article aims to present a computationally fast and efficient distributed policy learning approach to generate explicit closed-loop DMPC policies, facilitating scalable policy learning and deployment in optimization-based multirobot control. %Specifically, we propose a force field-inspired distributed policy learning algorithm with fast learning capability and safety guarantees, which are significant features for real-time control of MRS. %Since $P_i$ in~\eqref{Eqn:lyap} is only valid for linear systems, we also provide a new choice of $P_i$ to guarantee the stability of nonlinear systems in our policy learning framework, deferred in~\eqref{Eqn:Lya-mod}.
		
		\section{Fast Policy Learning Framework for DMPC}
		This section presents the proposed distributed policy learning framework to solve the DMPC problem~\eqref{Eqn:optimiz}. Then, we introduce a distributed actor-critic algorithm to quickly implement the policy learning approach, generating closed-loop control policies with lightweight neural networks. 
		
		Note that this section is dedicated to elucidating the policy learning design for DMPC in an unconstrained scenario, i.e., $\mathcal{E}_i=\mathbb{R}^{n_{\scriptscriptstyle \mathcal{N}_i}}$ and $\mathcal{U}_i=\mathbb{R}^{m_i}$. The extension to safe policy learning under state and control constraints is postponed to Section~\ref{sec:safe}.
		\subsection{Policy Learning Design for DMPC}\label{sec:acrl4mpc-o}
		
		Assume now that, at a generic time instant $k$, the problem~\eqref{Eqn:optimiz} is to be solved. Our goal is to generate an analytic control policy $u(e(\tau))=\text{col}_{i\in\mathbb{N}_1^M}(u_i(e_{\scriptscriptstyle \mathcal{N}_i}(\tau))$, $\forall \tau\in[k,k+N-1]$ that optimizes~\eqref{Eqn:optimiz} with the performance index $J(e(k))$.  Unlike the numerical DMPC that seeks a numerical solution by minimizing $J(e(k))$ over the whole prediction horizon, our work decomposes the optimization problem into $N$ cooperative subproblems. Using an efficient distributed policy learning approach, it solves them stepwise and forward in time. To this end, at each time instant $ \tau\in[k,k+N-1]$, we define $ r(\tau)=\sum_{i=1}^M r_i(\tau)$, $J\big(e(\tau)\big)=\sum_{i=1}^M J_i(e_{\scriptscriptstyle \mathcal{N}_i}(\tau))$, where  $J_i(e_{\scriptscriptstyle \mathcal{N}_i}(\tau))=r_i(\tau)+  J_i(e_{\scriptscriptstyle \mathcal{N}_i}(\tau+1))$ and $J_i(e_{\scriptscriptstyle \mathcal{N}_i}(k+N))=\|e_i(k+N)\|_{P_i}^{2}$. Denoting $ J^{*}(e(\tau))$  be the optimal value function associated with the optimal control policy $u^{\ast}(e(\tau))$, we write the Hamilton-Jacobi-Bellman (HJB) equation, for $\tau\in[k,k+N-1]$, as
		\begin{equation*}\label{Eqn:u_optimal-o}
			J^{*}(e(\tau))= \min_{ u_i(e_{\scriptscriptstyle \mathcal{N}_i}(\tau)), i\in\mathbb{N}_1^M}  r(\tau)+  J^{\ast}\big(e(\tau+1)\big).
		\end{equation*}
		
		\begin{figure}[h]
			\centerline{\includegraphics[scale=0.45]{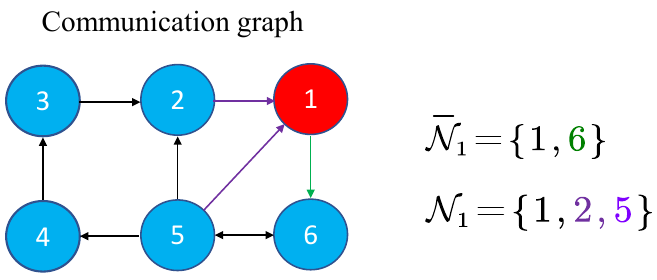}}
			\caption{ {\color{black}An exemplary scenario of communication graph with $M=6$. The arrows represent the directions of information exchange among robots. For the first robot, the set of its neighbors (including itself) is $\mathcal{N}_1=\{1,2,5\}$, while the set of robots that include robot 1 as one of the neighbors is $\bar{\mathcal{N}}_1=\{1,6\}$. The communications are instantaneously exchanged among neighboring robots at each step.}  }
			\label{fig:neighbor} 
		\end{figure}
		\begin{figure*}[h!]					\centerline{\includegraphics[scale=0.25]{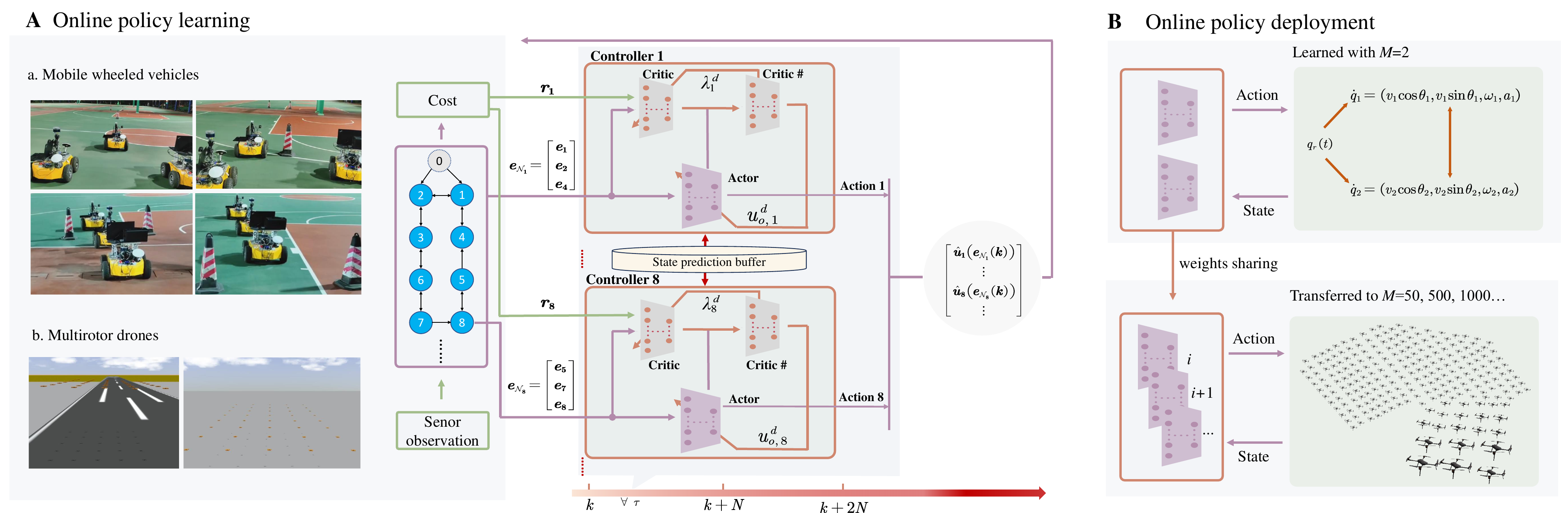}}
			\caption{ A: A sketch diagram of the distributed actor-critic learning algorithm in the prediction interval $[k,k+N-1]$, for the formation control of wheeled vehicles or multirotor drones. {\color{black}The definitions of $\lambda_i^d$ and $u_{o,i}^d$ are given in~\eqref{Eqn:lam_d-o} and~\eqref{Eqn:act-d-o}.} B: The learned control policy is of an explicit structure, and the one generated with 2 robots could be online deployed to 1,000 robots via weight sharing {\color{black}(see Section~\ref{sec:simu} for implementing details)}. } %For simplification, we use $e_{\scriptscriptstyle\mathcal{N}_i}$ and $e_{\scriptscriptstyle\mathcal{N}_i}^+$ to represent $e_{\scriptscriptstyle\mathcal{N}_i}(\tau)$ and $e_{\scriptscriptstyle\mathcal{N}_i}(\tau+1)$, respectively.}
		\label{fig:control_diagram} 
	\end{figure*}

	Note that with model~\eqref{Eqn:LL-nei}, the local control policy $u_i(e_{\scriptscriptstyle \mathcal{N}_i}(\tau))$ has only direct effects on the cost $J_j^{\ast}\big(e_{\scriptscriptstyle \mathcal{N}_j}(\tau+1)\big)$  for all $j\in \bar{ \mathcal{N}}_i$, where $\bar{ \mathcal{N}}_i$ is the collection of local robots that include the $i$-th robot as one of their neighbors (a graphical illustration of $\bar{ \mathcal{N}}_i$ is provided in Fig.~\ref{fig:neighbor} for clarity).
	Hence, the optimal control of robot $i$ at each time $\tau\in[k,\,k+N-1]$ could be calculated utilizing the related one-step ahead optimal cost $J_j^{\ast}\big(e_{\scriptscriptstyle \mathcal{N}_j}(\tau+1)\big)$ for all $j\in\bar{ \mathcal{N}}_i$, that is,
	\begin{equation}\label{Eqn:hjb-o}
		u_i^{\ast}(e_{\scriptscriptstyle \mathcal{N}_i}(\tau))=
		\argmin{u_i(e_{\scriptscriptstyle \mathcal{N}_i}(\tau))}\left\{ r_i(\tau)+ \sum_{j\in \bar{ \mathcal{N}}_i} J_j^{\ast}\big(e_{\scriptscriptstyle \mathcal{N}_j}(\tau+1)\big)\right\},
	\end{equation}
	for all $i\in\mathbb{N}_1^M$. The connection between $u_i^{\ast}(e_{\scriptscriptstyle \mathcal{N}_i}(\tau))$ and $J_j^{\ast}\big(e_{\scriptscriptstyle \mathcal{N}_j}(\tau+1)\big)$, $j\in\bar{\mathcal{N}}_i$, as depicted in~\eqref{Eqn:hjb-o}, provides insights for developing our distributed policy learning framework, which will be subsequently introduced.
	
	\emph{Distributed policy learning}:  In each prediction interval $[k,k+N-1]$, the distributed policy learning procedure %in Algorithm~\ref{alg:d-lpc-o}. 
	is started with an initial control $u^0(e(\tau))=\text{col}_{i\in\mathbb{N}_1^M}(u_i^0(e_{\scriptscriptstyle \mathcal{N}_i}(\tau))$, then each robot's value function and control policy are updated  with iteration step  $t=1,\cdots$:
	
	%			$\forall i\in\mathbb{N}_1^M$
	%			$\tau=k,\cdots,k+N-1$
	%  Compute $e_i(\tau+1)$ with $u_i^t(e_{\scriptscriptstyle \mathcal{N}_i}(\tau))$;
	\begin{enumerate}[(i)]
		\item Parallel value update, for all $i\in\mathbb{N}_1^M$:
		\begin{subequations}\label{Eqn:safempc-o}
			\begin{align}\label{Eqn:value-up-dhp-o}
				\hspace{-12mm} J^{t+1}_i(e_{\scriptscriptstyle \mathcal{N}_i}(\tau))
				= r_i(\tau)+ J_i^{t}(e_{\scriptscriptstyle \mathcal{N}_i}(\tau+1)).
			\end{align}
			\item Synchronous policy update, for all $i\in\mathbb{N}_1^M$:
			\begin{align}\label{Eqn:policy-im-DHP-o}
				\hspace{-12mm} u_i^{t+1}(e_{\scriptscriptstyle \mathcal{N}_i}(\tau))=\argmin{u_i(\tau)} \left\{r_i(\tau)+\sum_{j\in \bar {\mathcal{N}}_i} J_j^{t+1}\big(e_{\scriptscriptstyle \mathcal{N}_j}(\tau+1)\big)\right\}.
			\end{align}
		\end{subequations}
		
	\end{enumerate}
	
	Procedure~\eqref {Eqn:safempc-o} is executed stepwise and forward within each prediction interval. Consequently, at each iteration time step of each robot $i$, our approach only requires one-step ahead state predictions of its neighbors for policy updates, which can be calculated using~\eqref{Eqn:LL-nei} with the current states and actions. This approach differs from the traditional DMPC implementation, where solving~\eqref{Eqn:optimiz} over the prediction horizon for each robot $i$ usually involves all future states of neighbors within the prediction interval, which may not align with the actual states~\cite{rawlings2009model}. 

	\emph{Terminal penalty matrix}: Previous work~\cite{conte2016distributed} has addressed the terminal penalty design issue to guarantee closed-loop stability, but only for linear systems.
	In this article, we extend the design of $P_i$ to guarantee the stability of nonlinear MRS. In particular, we choose $P_i$ as the solution to the following Lyapunov equation:
	\begin{equation}\label{Eqn:Lya-mod-o}
		F_i^{\top} P_{i}F_i-\bar{P}_{i} =-\beta_i\left(Q_{i}+K_{\scriptscriptstyle\mathcal{N}_i}^{\top} R_{i} K_{\scriptscriptstyle\mathcal{N}_i}\right)+\Gamma_{\scriptscriptstyle\mathcal{N}_i},\ \ \forall i\in\mathbb{N}_1^M,
	\end{equation} where $F_i=A_{\scriptscriptstyle\mathcal{N}_i}+B_{i} K_{\scriptscriptstyle\mathcal{N}_i}$, $A_{\scriptscriptstyle\mathcal{N}_i}\in\mathbb{R}^{n_i\times n_{\mathcal{N}_i}}$ and $B_i\in\mathbb{R}^{n_i\times m_i}$ are the model parameters of the linearized model of~\eqref{Eqn:LL-nei} around the origin, i.e.,
	$e_i(k+1)=A_{\scriptscriptstyle\mathcal{N}_i} e_{\scriptscriptstyle\mathcal{N}_i}(k)+B_i u_{i}(k)+\phi_i(e_{\scriptscriptstyle\mathcal{N}_i}(k),u_i(k)),$ $\phi_i$ is the linearization error and $\lim_{e_{\scriptscriptstyle\mathcal{N}_i},u_i\rightarrow 0}{\phi_i(e_{\scriptscriptstyle\mathcal{N}_i},u_i)}/{(e_{\scriptscriptstyle\mathcal{N}_i},u_i)}\rightarrow 0;$  $K_{\scriptscriptstyle\mathcal{N}_i}\in\mathbb{R}^{m_i\times n_{\scriptscriptstyle \mathcal{N}_i}}$ are gain matrices such that $u={\rm col}_{i\in\mathbb{N}_1^M}(K_{\scriptscriptstyle\mathcal{N}_i}e_{\scriptscriptstyle\mathcal{N}_i})$ is a stabilizing control policy,  $\bar{P}_{i}:=W_{i} \Upsilon_{i}^{\top} P_{i} \Upsilon_{i} W_{i}^{\top}$ lifts $P_{i}$ to the space of neighboring states belonging in $\mathbb{R}^{\scriptscriptstyle n_{ \mathcal{N}_i}}$,  $\Upsilon_{i} \in\{0,1\}^{n_{i} \times n}$ and $W_{i} \in\{0,1\}^{n_{\scriptscriptstyle \mathcal{N}_{i}} \times n}$ are selective matrices such that $e_{i}=\Upsilon_{i} e$, $e_{\scriptscriptstyle\mathcal{N}_{i}}=W_{i} e$; and
	$\sum_{i=1}^{M} W_{i}^{\top} \Gamma_{\mathcal{N}_{i}} W_{i} \leq 0.$
	
	Note that differently from~\cite{conte2016distributed}, the tuning parameter $\beta_i$ is introduced in~\eqref{Eqn:Lya-mod-o} to account for the nonlinearity $\phi_i(e_{\scriptscriptstyle\mathcal{N}_i}(k),u_i(k))$, which plays a crucial role in deriving the closed-loop stability result (deferred to Appendix~\ref{sec:theoretical}).

	\subsection{Distributed Online Actor-Critic Learning Implementation}\label{sec:aclearning-o} 
	%In large-scale applications, executing distributed policy learning procedure~\eqref{Eqn:safempc-o} is computationally intensive, particularly in online learning scenarios. 
	%Distributed policy learning procedure~\eqref{Eqn:safempc-o} is yet not ready for fast policy generation due to the policy optimization step~\eqref{Eqn:policy-im-DHP-o}. 
	The distributed policy learning procedure~\eqref{Eqn:safempc-o} is not ready for fast policy generation, primarily due to the complexities associated with~\eqref{Eqn:policy-im-DHP-o}.
	We now discuss a distributed actor-critic learning algorithm to efficiently implement~\eqref{Eqn:safempc-o} through lightweight neural network approximations of the control policy and value function. In detail, the implementation consists of $M$ actor-critic network pairs, each designed with linear combinations of basis functions for the local robot, to learn the associated control policy and value function. Each robot's local actor and critic networks are trained in a fully distributed manner, and the parameters therein are updated incrementally within each prediction interval, enabling fast online policy learning for large-scale MRS. Furthermore, the actor and critic models acquired during each prediction interval are successively refined in subsequent intervals, thus enhancing learning efficiency and guaranteeing closed-loop stability. %Compared with the policy procedure~\eqref{Eqn:safempc-o}, this implementation is more efficient and allows for online incremental learning. Moreover, the actor networks generate neural control policies that can be directly deployed to the robots.  %Importantly, the local robot does not need to possess information about the global communication topology. 
	
	\begin{algorithm}[h]
		\caption{ Online fast policy learning implementation for DMPC.}
		\label{alg:d-lpc-AC-o}
		\begin{algorithmic}[1]
			\REQUIRE  {
				\STATE Initialize $W_{c,i}$ and $W_{a,i}$ with uniformly distributed random matrices,   $i\in\mathbb{N}_1^M$;
				\STATE Set $ {\epsilon}>0$, ${\rm Err}\geq \epsilon$, $t=0$,  $t_{\rm max}$;}
			%\textbf{for} {$k=1,2,\cdots$} \textbf{do loop}
			\FOR{$k=1,2,\cdots$}
			\STATE Set $W_{c,i}^0=W_{c,i}$ and $W_{a,i}^0=W_{a,i}$, $\forall\, i\in\mathbb{N}_1^M$;
			\WHILE{${\rm Err}\geq  {\epsilon}\,\vee\, t\leq t_{\rm max}$}
			%		\STATE \textbf{1)} Policy $\pi^t=u^t(k),\cdots,u^t(k+N-1)$;
			\FOR{$\tau=k,\cdots,k+N-1$}
			\STATE  Compute $e_i(\tau+1)$ with $\hat u_i(e_{\scriptscriptstyle \mathcal{N}_i}(\tau))$ using~\eqref{Eqn:LL-nei}, $\forall\, i\in\mathbb{N}_1^M$; 
			\STATE  Derive $\hat{\lambda}_i(\tau)$ using $e_{\scriptscriptstyle\mathcal{N}_i}(\tau)$ and $\hat{\lambda}_i(\tau+1)$ using the one-step-ahead prediction $e_{\scriptscriptstyle\mathcal{N}_i}(\tau+1)$ with \eqref{eqn:critic-o}, $\forall\, i\in\mathbb{N}_1^M$;
			\STATE   Calculate $\lambda_i^{d}(\tau)$ with~\eqref{Eqn:lam_d-o} and $u_{o,i}^d(\tau)$ with~\eqref{Eqn:act-d-o}, $\forall\, i\in\mathbb{N}_1^M$;
			\STATE Update $W_{c,i}$  with \eqref{Eqn:wc-o} and $W_{a,i}$ with~\eqref{Eqn:wa-o}, $\forall\, i\in\mathbb{N}_1^M$;
			%Generate $\hat{\lambda}_i(\tau)$ using $W_{c,i}$  with \eqref{eqn:critic-o}, and set $\hat{\lambda}_i^{t+1}(\tau)=\hat{\lambda}_i(\tau), \,\forall  i\in\mathbb{N}_1^M$;
			%\STATE %Generate $\hat{u}_i(\tau)$ using $W_{a,i}$  with \eqref{Eqn:actor-o}, and set $\hat{u}_i^{t+1}(\tau)=\hat{u}_i(\tau), \,\forall  i\in\mathbb{N}_1^M$;
			\ENDFOR
			\STATE Set $W_{c,i}^{t+1}=W_{c,i}$ and $W_{a,i}^{t+1}=W_{a,i}$, $\forall\, i\in\mathbb{N}_1^M$; 
			\STATE Compute $${\rm Err}=
			\sum_{i=1}^M\|W_{c,i}^{t+1}-W_{c,i}^{t}\|+ \|W_{a,i}^{t+1}-W_{a,i}^{t}\|;$$
			\STATE $t\leftarrow t+1$;
			\ENDWHILE
			\STATE Calculate cost $J(e(k))$ with~\eqref{Eqn:HL_cost};
			\IF{Condition deferred in~\eqref{Eqn:sta-con-o} is violated}
			\STATE Re-initialize $W_{c,i}$ and $W_{a,i}$,  and repeat steps 3-15;
			\ENDIF
			\STATE Update $e_i(k+1)$, $i\in\mathbb{N}_1^M$, by applying $\hat u_i(e_{\scriptscriptstyle \mathcal{N}_i}(k))$ to~\eqref{Eqn:LL-nei}.
			\ENDFOR
		\end{algorithmic}
	\end{algorithm}
	
	\emph{Critic learning}: In principle, the local critic network for robot $i$ could be designed to approximate ${ J}_i(e_{\scriptscriptstyle\mathcal{N}_i}(\tau))$ or the so-called costate $\lambda_i(e_{\scriptscriptstyle\mathcal{N}_i}(\tau))=\partial {J}_i(e_{\scriptscriptstyle\mathcal{N}_i}(\tau))/\partial e_{\scriptscriptstyle\mathcal{N}_i}(\tau)$~\cite{venayagamoorthy2002comparison}. In the latter case, more model information is used to accelerate convergence in online learning. %For space limitations, we only describe how to design distributed DHP for implementing DLPC.  
	Hence, the critic network is constructed to represent the costate, i.e.,
	\begin{equation}\label{eqn:critic-o}
		\hspace{-2mm}\begin{array}{ll}
			\hat{\lambda}_i(e_{\scriptscriptstyle\mathcal{N}_i}(\tau))=W_{c,i}^{\top}\sigma_{c,i}(e_{\scriptscriptstyle\mathcal{N}_i}(\tau),\tau), \ \ i\in\mathbb{N}_1^M, 
		\end{array}
	\end{equation}
	for all $\tau\in [k,k+N-1]$, where
	$W_{c,i}\in\mathbb{R}^{n_{c,i}\times n_{\scriptscriptstyle\mathcal{N}_i} }$ is the weighting matrix,  $\sigma_{c,i}\in\mathbb{R}^{n_{c,i}}$ is a vector composed of basis functions, {\color{black} including polynomials, radial basis functions, sigmoid functions, hyperbolic tangent functions, and others.} 
	
	The goal of training the critic network is to minimize the deviation between $\hat{\lambda}_i(e_{\scriptscriptstyle\mathcal{N}_i}(\tau))$ and ${\lambda}_i^{\ast}(e_{\scriptscriptstyle\mathcal{N}_i}(\tau)):=\partial   J_i^{\ast}(e_{\scriptscriptstyle\mathcal{N}_i}(\tau))/\partial e_{\scriptscriptstyle\mathcal{N}_i}(\tau)$.  Since ${\lambda}_i^{\ast}(e_{\scriptscriptstyle\mathcal{N}_i}(\tau))$ is unknown,  we define the desired value of $\hat{\lambda}_i(e_{\scriptscriptstyle\mathcal{N}_i}(\tau))$ by taking the partial derivative of $e_{\scriptscriptstyle\mathcal{N}_i}(\tau)$ on~\eqref{Eqn:value-up-dhp-o}, i.e., 
	\begin{equation}\label{Eqn:lam_d-o}
		\begin{array}{l}
			\hspace{-2mm}{\lambda}_i^{d}(\tau)=
			2Q_{i}e_{\scriptscriptstyle\mathcal{N}_i}(\tau)+ \sum_{j\in \bar{\mathcal{N}}_i}\left({\frac{\partial f_{j}(e_{\scriptscriptstyle\mathcal{N}_j}(\tau))}{\partial e_{\scriptscriptstyle\mathcal{N}_i}(\tau)}}\right)^{\top}\hat{\lambda}^{[j]}_i(\tau+1),
		\end{array}
	\end{equation}
	for $\tau\in[k,k+N-1]$, where $\hat{\lambda}^{[j]}_i(e_{\scriptscriptstyle\mathcal{N}_i})\in\mathbb{R}^{n_j}$ is the associated entries of $\hat{\lambda}_i(e_{\scriptscriptstyle\mathcal{N}_i})$ corresponding to robot $j$.
	
	Let $ \epsilon_{c,i}(\tau)= {\lambda}_i^d(e_{\scriptscriptstyle\mathcal{N}_i}(\tau))-\hat\lambda_i(e_{\scriptscriptstyle\mathcal{N}_i}(\tau))$,  $\forall i\in\mathbb{N}_1^M$ be the local approximation error. Minimizing the quadratic cost $\delta_{c,i}(\tau)=\|\epsilon_{c,i}(\tau)\|^2$ leads to the update rule of $W_{c,i}$ as
	\begin{equation}\label{Eqn:wc-o}
		%	\begin{align}
			W_{c,i}(\tau+1)={W_{c,i}(\tau)} - {\gamma_{c,i}}\frac{\partial\delta_{c,i}(\tau)}{\partial W_{c,i}(\tau)}, 
		\end{equation}
		where $\gamma_{c,i}\in\mathbb{R}^+$ is the local learning rate.
		
		\emph{Actor learning}:  Likewise, for each robot $i$, we construct the actor network as
		\begin{equation}\label{Eqn:actor-o}
			\hat u_i(e_{\scriptscriptstyle \mathcal{N}_i}(\tau))=W_{a,i}^{\top}\sigma_{a,i}(e_{\scriptscriptstyle \mathcal{N}_i}(\tau),\tau),
		\end{equation}
		where  $W_{a,i}\in\mathbb{R}^{n_{u,i}\times m_i}$  is the weighting matrix, $\sigma_{a,i}\in\mathbb{R}^{\scriptscriptstyle n_{u,i}}$ is a vector composed of basis functions like in~\eqref{Eqn:actor-o}. 
		In view of the first-order optimality condition of~\eqref{Eqn:policy-im-DHP-o}, letting $u_{o,i}=2R_i\hat u_i$, we define a desired target of $u_{o,i}$ as
		\begin{equation}\label{Eqn:act-d-o}
			u_{o,i}^d(\tau):=-\sum_{j\in\bar{\mathcal{N}}_i}g_{i}^{\top}(e_i(\tau)) \hat{\lambda}^{[i]}_j(\tau+1),
		\end{equation}
		$\tau\in[k,k+N-1]$.
		Letting $\epsilon_{a,i}(\tau)= u_{o,i}^d(\tau)- u_{o,i}(\tau)$, at each time instant $\tau\in[k,k+N-1]$, each robot $i$ minimizes the quadratic cost
		$\delta_{a,i}(\tau)=\|\epsilon_{a,i}(\tau)\|^2$,
		leading to the update rule of $W_{a,i}$  as
		\begin{equation}\label{Eqn:wa-o}
			W_{a,i}(\tau+1)=W_{a,i}(\tau) - {\gamma_{a,i}}\frac{\partial\delta_{a,i}(\tau)}{\partial W_{a,i}(\tau)},\\
		\end{equation}
		where  $\gamma_{a,i}\in\mathbb{R}^+$ is the local learning rate. %The main implementing steps of Algorithm~\ref{alg:d-lpc-AC} is listed in Algorithm~\ref{alg:d-lpc-AC}.
		
		The learning steps of the distributed actor-critic implementation are summarized in Algorithm~\ref{alg:d-lpc-AC-o} (see also Fig.~\ref{fig:control_diagram}).
		After completion of the learning process in the prediction interval $[k,k+N-1]$, the first control action $u_i(e_{\scriptscriptstyle \mathcal{N}_i}(k))$ calculated with~\eqref{Eqn:actor-o} is applied to~\eqref{Eqn:LL-nei}. Then, the above learning process is repeated in the subsequent prediction interval $[k+1,k+N]$. In such a manner, the control policies generated from the prediction interval $[k,k+N-1]$ are successively refined in subsequent intervals, enabling fast and efficient policy learning with the closed-loop stability guarantee. In addition to online policy learning, the convergent control policy in the form~\eqref{Eqn:actor-o} could be directly deployed to MRS.
		\begin{remark}\label{rem:computation}
			{\color{black} Numerical DMPC methods can be roughly classified into non-iterative and iterative
				approaches, each with different computational demands~\cite{conte2012computational}. In non-iterative linear/linearized DMPC where neighbors communicate once per time step, the computational complexity is roughly $O(\sum_{i=1}^MN(n_{\scriptscriptstyle \mathcal{N}_i}+m_i)n_{\scriptscriptstyle \mathcal{N}_i}^2)$ if the local MPC is implemented with an efficient sparse solver~\cite{wang2009fast,diehl2009efficient}. In iterative DMPC methods where neighbors communicate several times per step, distributed optimization algorithms such as the alternating direction method of multipliers (ADMM) could also be used for negotiations between robots, further increasing the computational load~\cite{conte2012computational}.}  The main computational complexity of our approach to policy learning is due to~\eqref{Eqn:wc-o},~\eqref{Eqn:wa-o}, and the forward prediction with~\eqref{Eqn:LL-nei}, which is approximately $O(\sum_{i=1}^MN(n_{c,i}+n_{u,i}+n_{\scriptscriptstyle \mathcal{N}_i})n_{\scriptscriptstyle \mathcal{N}_i})$. When directly deploying the learned control policy, the overall online computational complexity is reduced to $O(\sum_{i=1}^Mn_{u,i}n_{\scriptscriptstyle \mathcal{N}_i})$ even for nonlinear MRS.  
		\end{remark}
		\begin{remark}
			Compared with traditional numerical DMPC, our approach has the following significant characteristics. i) Our approach learns the closed-loop control policy rather than calculating open-loop control sequences. ii) The policy is generated by a distributed online actor-critic implementation, and no numerical solver is required; iii) In each prediction interval, our policy learning procedure is executed forward in time and stepwise rather than DMPC numerically optimizing the performance index over the prediction horizon. {\color{black}As shown later in Table~\ref{tab:Tab_com1} of Section~\ref{sec:experiment}, our approach significantly improves computational efficiency compared to two numerical DMPC approaches with different numerical solvers~\cite{7546918} and~\cite{conte2016distributed}.} iv) The policy learning process is executed successively between adjacent prediction intervals to improve learning efficiency. This means that the control policies generated from each prediction interval are iteratively refined in subsequent intervals. This approach fundamentally contrasts the common independent problem-solving paradigm of numerical DMPC in different prediction intervals. v) Our control policy has an explicit structure and could be learned offline and deployed online to MRS with different scales (see Panel B in Fig.~\ref{fig:control_diagram}). Hence, our approach facilitates scalability and rapid adaptability through rapid policy learning and deployment.
		\end{remark}

		\subsection{Practical Stability Verification Condition}
		The learned control policy using Algorithm~\ref{alg:d-lpc-AC-o} may approximate the optimum  $\bm u^{\ast}(k)$ with non-negligible errors.  In this scenario, the overall cost value $J(k)$ might not be monotonically decreasing (see Panel A in Fig.~\ref{fig:barrier_function}) under the actor-critic implementation, and the stability argument commonly used in MPC is not applicable.  Therefore, we introduce a novel and practical condition to ensure closed-loop stability in our framework. To this end, we recall from Assumption~\ref{assum:stabilizing_control} that there exists a baseline stabilizing control policy sequence $\bm u^{b}=\break{\rm col}_{i\in\mathbb{N}_1^M}\bm u^{b}_i$,
		$\forall\, k\in\mathbb{N}$,  such that $J^b(e^b(k))$  is a Lyapunov candidate function satisfying
		\begin{equation}\label{Eqn:baseline_cost-o} J^b(e^b(k+1))- J^b(e^b(k))<-s(e^b(k),u^b(k)),
		\end{equation}
		where $s(\cdot,\cdot)$ is a class $\mathcal{K}$ function, $ J^b(e^b)=\sum_{i=1}^{M} J_i^b(e_{\scriptscriptstyle \mathcal{N}_i}^b)$, $J^b(e_{\scriptscriptstyle \mathcal{N}_i}^b)$ and $e_{\scriptscriptstyle \mathcal{N}_i}^b$ are the associated performance index in~\eqref{Eqn:HL_cost} and the evolution of the state under $u_i^{b}$, respectively.
		Hence, we introduce the following condition to verify closed-loop stability:
		\begin{equation}\label{Eqn:sta-con-o}
			J(e(k))\leq J^b(e^b(k)), \ k\in\mathbb{N}.
		\end{equation}
		
		This condition represents a practical and easily verifiable solution to address the challenge of ensuring closed-loop stability arising from the possible nonmonotonic decrease of the overall cost $J(e(k))$ during learning. In a novel insight, the key is to draw a monotonic decreasing function $J^b(e^b(k))$ and verify its consistent role as an upper bound for $J(e(k))$ (see Panel A in Fig.~\ref{fig:barrier_function}).
		\begin{remark}
			Note that verifying condition~\eqref{Eqn:sta-con-o} requires collecting the cost $J_i(e_{\scriptscriptstyle\mathcal{N}_i}(k))$ calculated within each local robot $i\in\mathbb{N}_1^M$ through communication networks. Alternatively, to mitigate the communication load, one can verify $J_i(e_{\scriptscriptstyle\mathcal{N}_i}(k))\leq J_i^b(e^b_{\scriptscriptstyle\mathcal{N}_i}(k))+\eta_i$, where $\eta_i$ with $i\in\mathbb{N}_1^M$ satisfies $\sum_{i=1}^M\eta_i\leq 0$, for ensuring condition~\eqref{Eqn:sta-con-o}.
		\end{remark}
		\begin{remark}\label{rem:baseline} The design of $J^b(e^b(k))$ with the baseline stabilizing policy, $\bm u^b(k)$, is not unique. Two candidate choices are described below.
			\begin{enumerate}[(i)] 
				\item  Calculate the optimal control sequence $\bm u^{\ast}(0)=\break{\rm col}_{i\in\mathbb{N}_1^M}\bm u^{\ast}_i(0)$ by solving~\eqref{Eqn:optimiz}. For each $i\in\mathbb{N}_1^M$, set $\bm u^{b}_i(0)=\bm u^{\ast}_i(0)$ at time $k=0$ and update $\bm u^{b}_i(k)=u_i^{b}(k|k-1),\cdots,u_i^{b}(k+N-1|k-1),K_{\scriptscriptstyle \mathcal{N}_i}e_{\scriptscriptstyle \mathcal{N}_i}(k+N|k)$ iteratively at each time $k\in\mathbb{N}$. This choice ensures the satisfaction of condition~\eqref{Eqn:sta-con-o}, as proven in Theorem~\ref{theo:stability-o} of Appendix~\ref{sec:32}.
				\item  Design $ J^b(k)$ as a monotonically decreasing function, with $J^b(\infty)=0$. This approach eliminates the need for prior knowledge of the associated baseline control policy $\bm u^{b}(k)$. Adopting this design makes the condition~\eqref{Eqn:sta-con-o} for stability verification less restrictive and more practical. 
			\end{enumerate} 
		\end{remark}

		We have rigorously established the convergence condition of the policy learning algorithm and the closed-loop properties of our approach. Please refer to Appendix~\ref{sec:32} for detailed theoretical proofs and analysis. 
		
		\begin{figure*}[h!]
			\centerline{\includegraphics[scale=0.40]{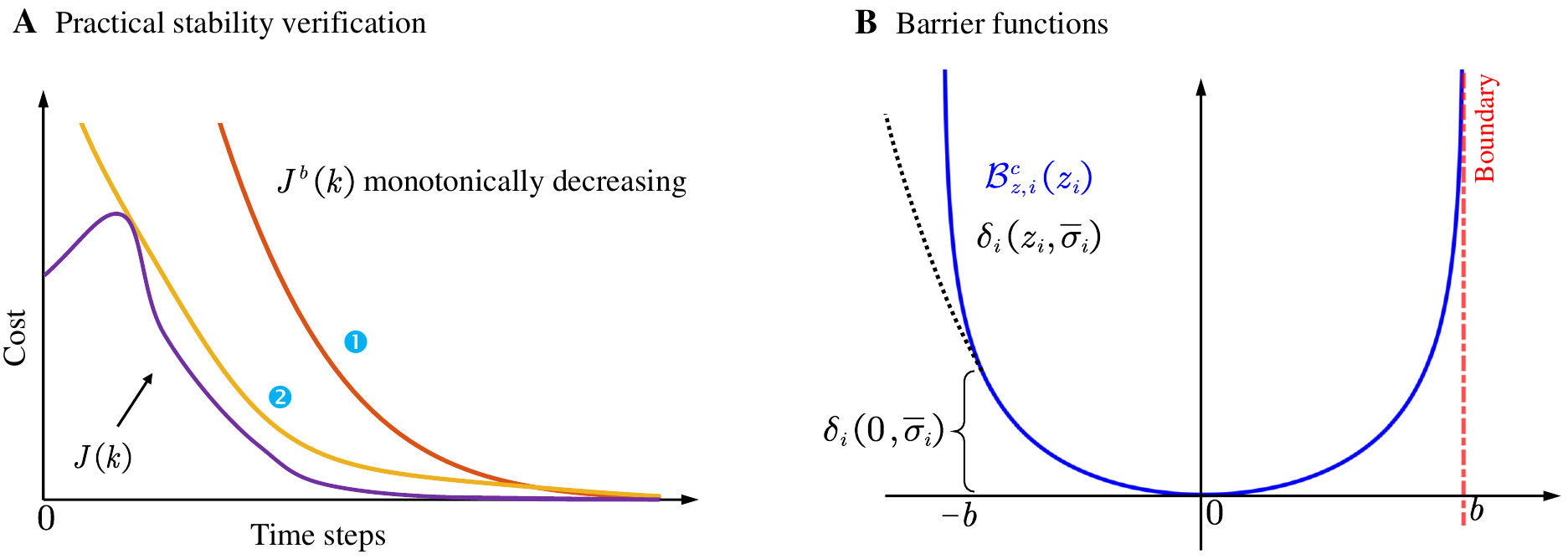}}
			\caption{ A: An example of the practical stability verification condition. {\color{black}The purple line represents the cost value using the distributed actor-critic learning algorithm, which may not be monotonically decreasing but is bounded by two monotonically decreasing cost values $J^b(k)$ under two baseline stabilizing control policies.} B: An example of the relaxed barrier function for $\mathcal{B}^o_{z,i}(z_i)=-\text{log}(b-z_i)-\text{log}(b+z_i)$, where the black dotted line represents $	\delta_i(z_i,\bar\sigma_i)$ in~\eqref{Eqn:relaxed_B}, while the blue line represents the recentered transformation ${\mathcal{B}}_{z,i}^c(z_i)=\mathcal{B}^o_{z,i}(z_i)+2\text{log}b$ centered at $z_{c,i}=0$.}
			\label{fig:barrier_function} 
		\end{figure*}	
		\section{Safe Policy Learning}\label{sec:safe}
		
		This section extends our approach to safe policy learning for DMPC under state and control constraints in~\eqref{eqn:constraints}. 
		We begin by introducing a novel force field-inspired safe policy learning design. This design ensures learning safety through a unique force field-inspired policy structure, offering clear physical interpretations.
		Subsequently, we provide a fast, distributed, safe actor-critic implementation, ensuring safety and efficiency in real-world applications.
		
		Note that barrier functions are commonly used in interior point optimization~\cite{boyd2004convex} to solve constrained optimization problems. 
		First, we provide some definitions of barrier functions, which will be used to form the force field-inspired policy structure, drawing inspiration from~\cite{boyd2004convex}. Please see Fig.~\ref{fig:barrier_function} for a visual description of the barrier functions defined below.
		%Having gained insight from~\cite{boyd2004convex}, we design the force field-inspired policy structure using barrier functions that are defined as follows (see also Fig.~\ref{fig:barrier_function} for a visual description of these functions).  %, resulting in near-optimal solutions with rigorous constraints satisfaction.
		
		\begin{definition}[Barrier functions~\cite{wills2004barrier}]\label{defini-log}
			For a set $\mathcal{Z}_i=\mathcal{E}_i$ or $\mathcal{U}_i$, define
			${\mathcal{B}}^o_{z,i}(z_i)=
			-\sum_{j=1}^{q_{z,i}}{\rm{log}}\big(-\Xi_{z,i}^j(z_i)\big),\ \ z_i\in {\rm {Int}}(\mathcal{Z}_i),$ and ${\mathcal{B}}^o_{z,i}(z_i)=+\infty$, $ \rm{otherwise}$.
			A recentered transformation of $ {\mathcal{B}}_{z,i}^o(z_i)$ centered at $z_{c,i}$ is defined as
			%\begin{equation}\label{Eqn:recenter}
			${\mathcal{B}}_{z,i}^c(z_i)= {\mathcal{B}}^o_{z,i}(z_i)- {\mathcal{B}}_{z,i}^o(z_{c,i})-\triangledown{\mathcal{B}}^o_{z,i}(z_{c,i})^{\top}z_i,$ with ${\mathcal{B}}_{z,i}^c(z_{c,i})=0$.
			A relaxed barrier function of ${\mathcal{B}}_{z,i}^c(z_i)$ is defined as
			\begin{equation}\label{Eqn:relaxed_B}
				{\mathcal{B}}_{z,i}^r(z_i)=\left\{\begin{array}{ll}
					{\mathcal{B}}_{z,i}^c(z_i)&\bar\sigma_i\geq \kappa_{i}\\
					\delta_i(z_i,\bar\sigma_i)&\bar\sigma_i <\kappa_{i},
				\end{array}\right.
			\end{equation}
			where $\kappa_{i}\in\mathbb{R}^+$ is a relaxing factor, 
			$\bar\sigma_i={\rm min}_{j\in\mathbb{N}_{1}^{q_{z,i}}}-\Xi^{j}_{z,i}(z_i)$, the function $\delta_i(z_i,\bar\sigma_i)$ is strictly monotone and differentiable on $(-\infty,\kappa_i)$, and $\triangledown^2\delta_{i}(z_i,\bar\sigma_i)\leq \triangledown^2{\mathcal{B}}_{z,i}^r(z_i)|_{\bar\sigma_i= \kappa_i}$. 		
		\end{definition}%
		
		%Note that other alternative functions, such as impulse potential fields and exponential functions~\cite{lu2021distributed,luo2019multi}, might also be used to construct barrier functions in line with Definition~\ref{defini-log}.  
		%
		
		\subsection{Force field-inspired Policy Learning Design}\label{sec:force-rl}
		
		\emph{Barrier-based cost shaping:} In line with~\cite{wills2004barrier}, %we formulate the optimization problem by introducing barrier functions of constraints in the cost function $J(e(k))$ (see~\eqref{Eqn:HL_cost}) for policy learning. Now, 
		we reconstruct the cost function with barrier functions as 
		$\bar J(e(k))=\sum_{i=1}^{M}\bar J_i(e_{\scriptscriptstyle \mathcal{N}_i}(k))$, and	
		\begin{equation}
			\begin{array}{ll}
				\bar J_i(e_{\scriptscriptstyle \mathcal{N}_i}(k))=\vspace{1mm}\\\sum_{j=0}^{N-1}\bar r_i(e_{\scriptscriptstyle \mathcal{N}_i}(k+j),u_i(k+j))+\bar J_i(e_i(k+N)),
			\end{array}\label{Eqn:re-cost}
		\end{equation}	
		where $\bar r_i(e_{\scriptscriptstyle \mathcal{N}_i}(\tau),u_i(\tau))=r_i(e_{\scriptscriptstyle \mathcal{N}_i}(\tau),u_i(\tau))+\hspace{5mm}\mu (\mathcal{B}_{e,i}(e_{\scriptscriptstyle \mathcal{N}_i}(\tau))+\mathcal{B}_{u,i}(u_i(\tau)))$, $\mathcal{B}_{z,i}(z_i(\tau))$ for $z_i=e_{\scriptscriptstyle \mathcal{N}_i},u_i$ are the relaxed barrier functions in Definition~\ref{defini-log}, $\tau\in[k,k+N-1]$, $\bar J_i(e_i(k+N))=\|e_i(k+N)\|_{P_i}^{2}+\mu \mathcal{B}^{[{\rm T}]}_{e,i}(e_i(k+N))$, $\mathcal{B}^{[{\rm T}]}_{e,i}(e_i(k+N))$ is constructed with the recentered barrier function of the terminal constraint (see again Definition~\ref{defini-log}), the tuning parameter $\mu>0$ adjusts the influence of barrier functions on $J(e(k))$. 
		
		\emph{Force field-inspired policy structure}:  It's worth noting that optimizing $\bar J(e(k))$ in~\eqref{Eqn:re-cost} does not guarantee safe learning within the actor-critic framework~\cite{Paternain9718160,xinglongzhang2022_robust}. As discussed in the interior point optimization~\cite{boyd2004convex}, minimizing $\bar J(e(k))$ results in an optimal solution influenced by two acting forces. One is the constraint force associated with the barrier functions in $\bar J(e(k))$, while the other originates from the objective function $J(e(k))$. Balancing these two acting forces within an actor-critic structure presents a considerable challenge~\cite{Paternain9718160}. Consequently, we devise a force field-inspired policy structure representing the joint action of the objective and constraint forces, ensuring safety during policy optimization with clear physical interpretations. 
		
		%Our control policy structure is inspired by the previously described force field interpretation of the optimal solution. 
		Specifically, for each robot $i$, our proposed control policy comprises a nominal control policy that generates the objective force, along with two gradient terms of barrier functions that generate constraint forces associated with the control and state constraints, i.e., 
		\begin{equation}\label{Eqn:control}
			\begin{array}{ll}
				\bar u_i(e_{\scriptscriptstyle \mathcal{N}_i})&=\nu_i(e_{\scriptscriptstyle \mathcal{N}_i})+L_{e,i}\triangledown\mathcal{B}_{e,i}(e_{\scriptscriptstyle \mathcal{N}_i})+L_{\nu,i}\triangledown\mathcal{B}_{\nu,i}(\nu_i(e_{\scriptscriptstyle \mathcal{N}_i})),\\
				&=\nu_i(e_{\scriptscriptstyle \mathcal{N}_i})+L_i\cdot(\triangledown\mathcal{B}_{e,i}(e_{\scriptscriptstyle \mathcal{N}_i}),\triangledown\mathcal{B}_{\nu,i}(\nu_i(e_{\scriptscriptstyle \mathcal{N}_i}))),
			\end{array} %\sum_{j\in{\mathcal{N}}_i}L_{ij}\triangledown_{e_j}\mathcal{B}_j(e_{j})
		\end{equation}
		%
		\begin{comment}
			also,
			\begin{equation}\label{Eqn:V_bzk}
				V_{b,\tau}\big(\hat {s}(\tau)\big)=\left\{\begin{array}{ll}r_b(\tau)+\bar V_{b,\tau+1}\big(\hat{s}(\tau+1)\big)+\mu B(e_{{s}}(\tau)),& \mbox{for}\ \tau=k\\
					r_b(\tau)+\bar V_{b,\tau+1}\big(\hat{s}(\tau+1)\big),&\mbox{for}\ \tau\in[k+1,k+N-1]
				\end{array}\right.
			\end{equation}
			\begin{remark}
				In~\eqref{Eqn:extend-cost},  the overall cost function consists of the classical quadratic-type costs and the barrier functions on the state and control. The optimization of such a cost can be regarded as a trade-off between quadratic optimality and control safety. The barrier functions become sufficiently large if the control and state are close to the boundaries of the safety constraints. In this case,  the safety cost instead of the quadratic one is a priority that needs to be minimized. \hfill $\blacktriangle$
			\end{remark}
			To balance the quadratic cost and safety cost in~\eqref{Eqn:extend-cost}, we propose the safe control policy of a special type as follows:
		\end{comment}
		%	\begin{comment}
			%	
			where  $\nu_i(e_{\scriptscriptstyle \mathcal{N}_i})\in\mathbb{R}^{m_i}$ is a parameterized control policy to generate the objective force, the remaining gradient-based terms are to generate the constraint forces, $L_i=[L_{e,i}\ L_{\nu,i}]$,  $L_{e,i}\in\mathbb{R}^{m_i\times n_{\scriptscriptstyle \mathcal{N}_i}}$ and $L_{\nu,i}\in\mathbb{R}^{m_i\times m_i}$. The parameters of $\nu_i(e_{\scriptscriptstyle \mathcal{N}_i})$ and $L_i$ are decision variables to be further optimized by minimizing~\eqref{Eqn:re-cost}. 
			%\begin{remark}
			%	The control policy \eqref{Eqn:control} of robot $i$ is influenced by its neighboring state $e_{\scriptscriptstyle \mathcal{N}_i}$. The second and third terms generate local repulsive constraint forces associated with the $i$-th robot, which grow rapidly as the local state $e_i$ or control $u_i$ moves toward the boundary of the associated set. %In doing so, $e_i$ and $u_i$ can be restricted in their interiors, i.e., the gradients of the barrier functions can guarantee state and control constraints satisfaction in~\eqref{Eqn:control}.  
			% \hfill $\blacktriangle$
			% \end{remark}
		
		\emph{Terminal penalty matrix}: Since barrier functions are employed in cost reconstruction~\eqref{Eqn:re-cost},  the penalty matrix $P_i$ determined from \eqref{Eqn:Lya-mod-o} is rendered inapplicable for ensuring stability guarantees. We recall from~\cite{wills2004barrier}, there exists a positive-definite matrix $H_{z,i}$ satisfying 
		\begin{equation}\label{eqn:barrier}
			H_{z,i}\geq\triangledown^2{\mathcal{B}}_{z,i}^r(z_i)|_{\bar\sigma_i= \kappa_i},
		\end{equation}
		%\end{equation} 
		such that $\|\triangledown {\mathcal{B}}_{z,i}^r(z_i)\|\leq {\mathcal{B}}_{z_i,m}$, for $z_i=e_{\scriptscriptstyle\mathcal{N}_i}$ or $u_i$,  where ${\mathcal{B}}_{z_i,m}=\max_{z_i\in\mathcal{Z}_i}\|2H_{z,i}(z_i-z_{c,i})\|$. Hence,  matrix $P_i$ is now calculated as the solution to the following Lyapunov equation: 
		\begin{equation}\label{Eqn:Lya-mod}
			\begin{array}{ll}
				F_i^{\top} P_{i}F_i-\bar{P}_{i} =-\beta_i\left(\mu (H_{e,i}+ K_{\scriptscriptstyle\mathcal{N}_i}^{\top}H_{u,i}K_{\scriptscriptstyle\mathcal{N}_i})+\right.\\
				\hspace{45mm}\left.Q_{i}+K_{\scriptscriptstyle\mathcal{N}_i}^{\top} R_{i} K_{\scriptscriptstyle\mathcal{N}_i}\right)+\Gamma_{\scriptscriptstyle\mathcal{N}_i},
			\end{array}
		\end{equation}$\forall i\in\mathbb{N}_1^M$. %matrix $H_{\scriptscriptstyle\mathcal{N}_i}$ is such that $H_{\scriptscriptstyle\mathcal{N}_i}=T_{\scriptscriptstyle\mathcal{N}_i}^{\top}H_{e,i}T_{\scriptscriptstyle\mathcal{N}_i}$, $T_{\scriptscriptstyle\mathcal{N}_i} \in\{0,1\}^{ n_i\times n_{\scriptscriptstyle \mathcal{N}_{i}}}$ is a selective matrix satisfying $e_{i}=T_{\scriptscriptstyle\mathcal{N}_i} e_{\scriptscriptstyle\mathcal{N}_{i}}$.
		Unlike~\eqref{Eqn:Lya-mod-o}, $H_{e,i}$ and $H_{u,i}$ are derived satisfying~\eqref{eqn:barrier} to account for the barrier functions in~\eqref{Eqn:re-cost}. 
		\subsection{Distributed Safe Actor-Critic Learning Implementation}\label{sec:aclearning} 
		We design the distributed safe actor-critic learning algorithm following the line in Section~\ref{sec:aclearning-o}. In this scenario, the actor and critic are constructed with barrier forces consistent with the force field-inspired policy and barrier-based cost function. This design has clear physical force field interpretations to ensure safety and convergence during policy learning.
		%Differently from the actor-critic implementation in Section~\ref{sec:aclearning-o}, we present a distributed safe actor-critic structure in which the actor and critic are designed with barrier functions and consistent with the proposed force field-inspired policy and the barrier-based cost function. Hence, our distributed safe actor-critic learning algorithm has clear physical force field interpretations to ensure safety and convergence during learning. 
		
		\emph{Barrier-based critic learning}: 
		For any robot $i\in\mathbb{N}_1^M$, the critic network is constructed with barrier gradients, i.e., 
		\begin{equation}\label{eqn:critic}
			\hspace{-2mm}\begin{array}{ll}
				\hat{\bar{\lambda}}_i(e_{\scriptscriptstyle\mathcal{N}_i}(\tau))&\hspace{-3mm}=(\bar W_{c,i}^{[1]})^{\top}\sigma_{c,i}(e_{\scriptscriptstyle\mathcal{N}_i}(\tau),\tau)+(\bar W_{c,i}^{[2]})^{\top}\triangledown\mathcal{B}_{e,i}(e_{\scriptscriptstyle\mathcal{N}_i}(\tau)),\vspace{1mm}\\
				\hspace{-2mm}&	\hspace{-3mm}=(\bar W_{c,i})^{\top}h_{c,i}(e_{\scriptscriptstyle\mathcal{N}_i}(\tau),\tau),
			\end{array}
		\end{equation}
		for all $\tau\in [k,k+N-1]$, where
		$\bar W_{c,i}^{[1]}\in\mathbb{R}^{n_{c,i}\times n_{\scriptscriptstyle\mathcal{N}_i} }$ and $\bar W_{c,i}^{[2]}\in\mathbb{R}^{ n_i\times n_{\scriptscriptstyle\mathcal{N}_i}}$ are the weighting matrices,  $\sigma_{c,i}\in\mathbb{R}^{n_{c,i}}$ is a vector composed of basis functions like in~\eqref{Eqn:actor-o}, $\bar W_{c,i}=[(\bar W_{c,i}^{[1]})^{\top}\ (\bar W_{c,i}^{[2]})^{\top}]^{\top}$, $h_{c,i}(e_{\scriptscriptstyle\mathcal{N}_i}(\tau),\tau)=(\sigma_{c,i}(e_{\scriptscriptstyle\mathcal{N}_i}(\tau),\tau),\triangledown\mathcal{B}_{e,i}(e_{\scriptscriptstyle\mathcal{N}_i}(\tau)))$. 
		
		%The goal is to minimize the deviation between $\hat{\bar {\lambda}}_i(e_{\scriptscriptstyle\mathcal{N}_i})$ and $\bar{\lambda}_i^{\ast}(e_{\scriptscriptstyle\mathcal{N}_i}):=\partial \bar J_i^{\ast}(e_{\scriptscriptstyle\mathcal{N}_i})/\partial e_{\scriptscriptstyle\mathcal{N}_i}$. To this end, w
		In line with~\eqref{Eqn:lam_d-o}, define the desired value of $\hat{\bar{\lambda}}_i(e_{\scriptscriptstyle\mathcal{N}_i}(\tau))$ as 
		\begin{equation}\label{Eqn:lam_d}
			\begin{array}{ll}
				\hspace{-2mm}{\bar{\lambda}}_i^{d}(e_{\scriptscriptstyle\mathcal{N}_i}(\tau))=
				2Q_{i}e_{\scriptscriptstyle\mathcal{N}_i}(\tau)+\mu\frac{\partial \mathcal{B}_{e,i}(e_{\scriptscriptstyle\mathcal{N}_i}(\tau))}{\partial e_{\scriptscriptstyle\mathcal{N}_i}(\tau)}+\vspace{2mm}\\
				\hspace{20mm} \sum_{j\in \bar{\mathcal{N}}_i}\left({\frac{\partial f_{j}(e_{\scriptscriptstyle\mathcal{N}_j}(\tau))}{\partial e_{\scriptscriptstyle\mathcal{N}_i}(\tau)}}\right)^{\top}\hat{\bar{\lambda}}^{[j]}_i(\tau+1),
			\end{array}
		\end{equation}
		for $\tau\in[k,k+N-1]$, where $\hat{\bar{\lambda}}^{[j]}_i(\tau)\in\mathbb{R}^{n_j}$ is the associated entries of $\hat{\bar{\lambda}}_i(e_{\scriptscriptstyle\mathcal{N}_i}(\tau))$ corresponding to robot $j$.
		%,
		%	 $\hat{\bar{\lambda}}_i(e_{\scriptscriptstyle\mathcal{N}_i}(k+N))=\partial \hat {\bar J}_i(e_{\scriptscriptstyle\mathcal{N}_i}(k+N))/\partial e_{\scriptscriptstyle\mathcal{N}_i}(k+N)$, $\hat {\bar J}_i(e_{\scriptscriptstyle\mathcal{N}_i}(k+N))=\|e_i(k+N)\|_{P_i}^2+\mu\mathcal{B}_{e,i}(e_i(k+N)).$
		
		Let $\bar{\epsilon}_{c,i}(\tau)= {\bar{\lambda}}_i^d(e_{\scriptscriptstyle\mathcal{N}_i}(\tau))-\hat{\bar{\lambda}}_i(e_{\scriptscriptstyle\mathcal{N}_i}(\tau))$,  $\forall i\in\mathbb{N}_1^M$. Minimizing $\bar{\delta}_{c,i}(\tau)=\|\bar{\epsilon}_{c,i}(\tau)\|^2$ leads to the update rule:
		\begin{equation}\label{Eqn:wc}
			%	\begin{align*}
				\bar W_{c,i}^{[j]}(\tau+1)={\bar W_{c,i}^{[j]}(\tau)} - {\gamma_{c,i}^{[j]}}\frac{\partial\bar{\delta}_{c,i}(\tau)}{\partial \bar W_{c,i}^{[j]}(\tau)}, \ \forall j=1,2, 
			\end{equation}
			where $\gamma_{c,i}^{[j]}\in\mathbb{R}^+$, $j=1,2$, are the local learning rates.
			
			\begin{figure*}[h!]
				
				\centerline{\includegraphics[scale=0.3]{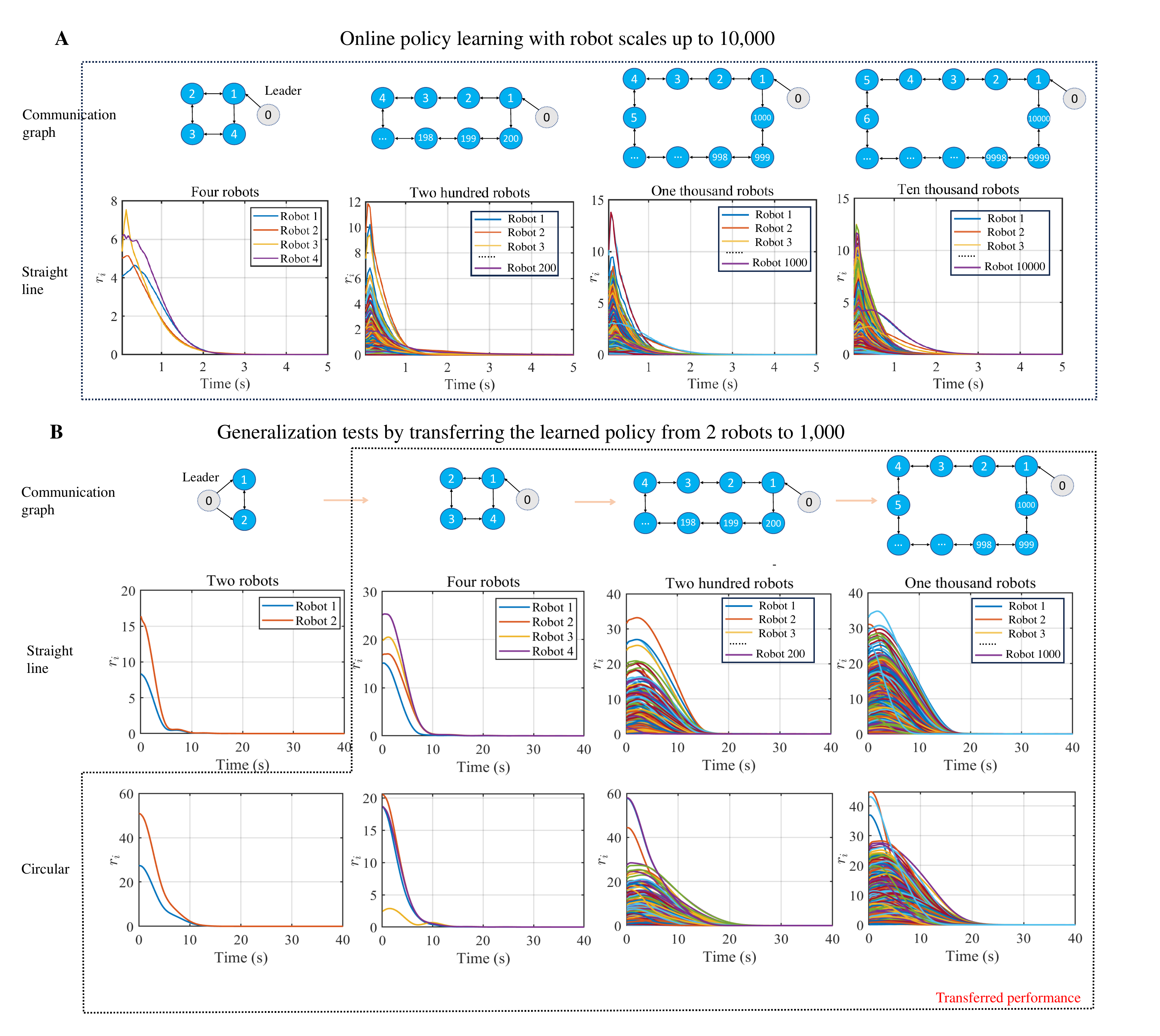}}
				\caption{ A: Online policy learning with robot scales up to 10,000, where $r_i(k)=\| e_{\scriptscriptstyle \mathcal{N}_i}(k)\|_{Q_i}^{2}+\| u_i(k)\|_{R_i}^{2}.$ B: Transferred performance of straight-line formation of 2 robots to the circular formation of 2 robots and different formation scenarios of 4, 200, and 1,000 robots. {\color{black}Note that, ``two robots" actually pertains to two follower robots and a leader. The leader adopted in this work is a virtual entity, which is not counted in the total number of robots.}}
				\label{fig:trajectory-gen} 
			\end{figure*}
			
			\emph{Force field-inspired actor learning}:  Likewise, for each robot $i$, we construct the actor network to generate the force field-inspired policy, i.e.,
			\begin{equation}\label{Eqn:actor}
				\begin{array}{ll}
					\hat {\bar u}_i(e_{\scriptscriptstyle \mathcal{N}_i}(\tau))&\hspace{-2mm}=(\bar W_{a,i}^{[1]})^{\top}\sigma_{a,i}(e_{\scriptscriptstyle \mathcal{N}_i}(\tau),\tau)+\\
					&\hspace{2mm} (\bar W_{a,i}^{[2]})^{\top}\triangledown\mathcal{B}_{e,i}(e_{\scriptscriptstyle\mathcal{N}_i}(\tau))+(\bar W_{a,i}^{[3]})^{\top}\triangledown\mathcal{B}_{\nu,i}(\hat{\nu}_i(\tau)), \vspace{2mm}\\
					&\hspace{-2mm}=\bar W_{a,i}^{\top}h_{a,i}(e_{\scriptscriptstyle \mathcal{N}_i}(\tau),\tau),
				\end{array}
			\end{equation}
			where $\hat{\nu}_i(\tau)=(\bar W_{a,i}^{[1]})^{\top}\sigma_{a,i}(e_{\scriptscriptstyle \mathcal{N}_i}(\tau),\tau)$, $\bar W_{a,i}^{[1]}\in\mathbb{R}^{n_{u,i}\times m_i}$  is the weighting matrix, 
			$[(\bar W_{a,i}^{[2]})^{\top}\ (\bar W_{a,i}^{[3]})^{\top}]\in\mathbb{R}^{m_i\times (n_{\scriptscriptstyle\mathcal{N}_i}+m_i)}$ is the approximation of $L_i$,  $\sigma_{a,i}\in\mathbb{R}^{\scriptscriptstyle n_{u,i}}$ is a vector composed of basis functions like in~\eqref{Eqn:actor-o}, $W_{a,i}=[(\bar W_{a,i}^{[1]})^{\top}\ (\bar W_{a,i}^{[2]})^{\top}\ (\bar W_{a,i}^{[3]})^{\top}]^{\top}$, $h_{a,i}(e_{\scriptscriptstyle\mathcal{N}_i}(\tau),\tau)=(\sigma_{a,i}(e_{\scriptscriptstyle\mathcal{N}_i}(\tau),\tau),\triangledown\mathcal{B}_{e,i}(e_{\scriptscriptstyle\mathcal{N}_i}(\tau)),\triangledown\mathcal{B}_{\nu,i}(\hat{\nu}_i(\tau)))$. 
			Letting $\bar u_{o,i}=2R_i\hat {\bar{u}}_i+\mu\triangledown \mathcal{B}_{u,i}(\hat {\bar{u}}_i)$, we define a desired target of $\bar u_{o,i}$ as
			\begin{equation}\label{Eqn:act-d}
				\bar u_{o,i}^d(\tau):=-\sum_{j\in\bar{\mathcal{N}}_i}g_{i}^{\top}(e_i(\tau)) \hat{\bar{\lambda}}^{[i]}_j(\tau+1),
			\end{equation}
			for $\tau\in[k,k+N-1]$. Let $\bar{\epsilon}_{a,i}(\tau)= \bar u_{o,i}^d(\tau)- \bar u_{o,i}(\tau)$.
			At each time instant $\tau\in[k,k+N-1]$, each robot $i$ minimizes
			$\bar{\delta}_{a,i}(\tau)=\|\bar{\epsilon}_{a,i}(\tau)\|^2$,
			leading to the update rule:
			\begin{equation}\label{Eqn:wa}
				\bar W_{a,i}^{[j]}(\tau+1)=\bar W_{a,i}^{[j]}(\tau) - {\gamma_{a,i}^{[j]}}\frac{\partial\bar{\delta}_{a,i}(\tau)}{\partial \bar W_{a,i}^{[j]}(\tau)},\ \forall j=1,2,3,\\
			\end{equation}
			where  $\gamma_{a,i}^{[j]}\in\mathbb{R}^+$, $j=1,2,3$, are the local learning rates. 
			
			Due to space limitations, we have omitted the summarized implementation steps of the safe policy learning algorithm and the theoretical results. Please refer to the attached materials (see ``auxiliary-results.pdf" in the uploaded package ``auxiliary-material.zip") for comprehensive implementation steps and details of the theoretical analysis.
			
			%
			%
			%		\begin{figure*}[h!]
				%			% \setlength{\abovecaptionskip}{0.cm}
				%			\centerline{\includegraphics[scale=0.55]{figure/Communication_diagram}}
				%			\caption{ A: Communication graph of $M=16$ mobile robots, where the arrows indicate the direction of information transmission, and robot 0 is the leading one. B: The path of robots in formation control and collision avoidance, where the black circular areas (0.4m in diameter) represent the obstacles, and the coloured lines represent the robots' paths. Meanwhile, the robots in the same column are marked with the same coloured dots.}
				%			\label{fig:Communication diagram1} 
				%		\end{figure*}
			\section{Simulation and Experimental Results}\label{sec:experiment}
			This section evaluates our methodology for formation control, which involves simulated and real-world experiments on mobile wheeled vehicles and multirotor drones. Through simulated and real-world experiments, we aim to show that: a) our approach could online learn near-optimal control policies efficiently for very large-scale MRS and is more scalable than nonlinear numerical DMPC; b) the control policy learned using nominal kinematic models could be directly transferred to real-world mobile wheeled vehicles and multirotor drones; c) our approach shows strong transferability by deploying the learned policy to robots with different scales.  
			
			We have shown on different computing platforms, i.e., a laptop and a Raspberry PI 5, that our approach could efficiently learn the DMPC policies for MRS with scales up to 10,000, and the computational load grows linearly with robot scales in both platforms. As far as we know, no optimization-based control approach has realized distributed control for MRS on such a large scale. Our learned control policies, trained with 2 robots, could be deployed directly to MRS with scales up to 1,000. Furthermore, our learned control policies could be deployed to real-world wheeled MRS with different scales.
			
			%In this section, we evaluate our approach on formation control of mobile wheeled vehicles and multirotor drones. Notably, our learned policy with small scales can be directly deployed to mobile wheeled vehicles with scales up to 1,000 and to multirotor drones in Gazebo with scales up to 54. Also, we have validated our approach through real-world experiments on wheeled MRS (see Fig.~\ref{fig:policy_vehicle} for detailed experimental setup).  Through simulated and real-world experiments, We aim to show that: a) our approach is more scalable than DMPC for distributed control of MRS; b) the control policy learned using nominal kinematic models can be directly transferred to real-world mobile wheeled vehicles and multirotor drones; c) our approach shows strong transferability to robots with different scales. 
			
			%In Appendix~\ref{sec:singlerobot},  we also utilize our approach to online solve a single-robot path-following control problem in a distributed way, demonstrating that our approach is near-optimal and comparable to the centralized solution in terms of control performance. 
			%
			%
			%
			\begin{table}
				\begin{center}
					\caption{Hyperparameters of DLPC.}
					\begin{threeparttable}
						\scalebox{0.9}{
							\renewcommand{\arraystretch}{1.5}
							\begin{tabular}{cccc}
								\toprule  %????????
								\textbf{Hyperparameter\tnote{(1)}} & \textbf{Value}& \textbf{Hyperparameter} & \textbf{Value}\\
								\midrule  %???????
								$\Delta t$ & 0.05 s & 	$Q_i$ & $I_{\mathcal{N}_i}$ \\\hline
								$R_i$ & $0.5I_2$ & $\kappa_i$ & 0.1 \\\hline
								$n_{u,i}$ & 4 & $n_{c,i}$&4 \\\hline
								$\gamma_{c,i}$& 0.4& $\gamma_{a,i}$& $0.2$\\
								\hline
								$\gamma_{c,i}^{[1]}$& 0.4& $\gamma_{c,i}^{[2]}$& e--5\\
								\hline
								$\gamma_{a,i}^{[1]}$ &0.2& $\gamma_{a,i}^{[2]}$&$0.1$\\
								\hline
								$\gamma_{a,i}^{[3]}$&$0.1$& $t_{\rm max}$&30\\ \hline
								$\beta_i$& 1.1& $\mu$& 0.02\\ \bottomrule
								\label{tab:training_hyperparameters}
						\end{tabular}}
						\begin{tablenotes}
							\scriptsize
							\item[(1)] {\color{black}\scalebox{0.9}{The parameters in the table are dynamically tuned in the simulation studies.}}
						\end{tablenotes}
					\end{threeparttable}
				\end{center}	
			\end{table}
			\begin{table}[h!tb]
				\centering \caption{Numerical error measure for formation control of 16 robots. MAE=Maximum absolute error, RMSE=Root mean square error (corresponds to Fig.~\ref{fig:Communication diagram2}-B).}
				\label{tab:Tab_com0}
				%\vskip 0.2cm
				%
				\scalebox{0.9}{
					\renewcommand{\arraystretch}{1.2}
					\begin{tabular}{ccccc}
						%			\toprule 
						\toprule
						Error &$e_{x,i}$(m) &$e_{y,i}$(m)& $e_{\theta,i}$(rad)&$e_{v,i}$(m/s) \\ \midrule
						MAE in coll. avoid.	& $0.38$ & $0.51$&0.56&0.15\\\midrule
						RMSE in coll. avoid.	& $0.04$ & $0.11$&0.08&0.04\\\midrule
						Steady-state error 	 & $2$e--7 &$2$e--7&$2$e--7&$7$e--8\\
						\bottomrule
					\end{tabular}
				}	 
			\end{table}	
			\subsection{Simulated Experiments on MRS}\label{sec:simu}
			\begin{figure*}[h!]
				\centerline{\includegraphics[scale=0.38]{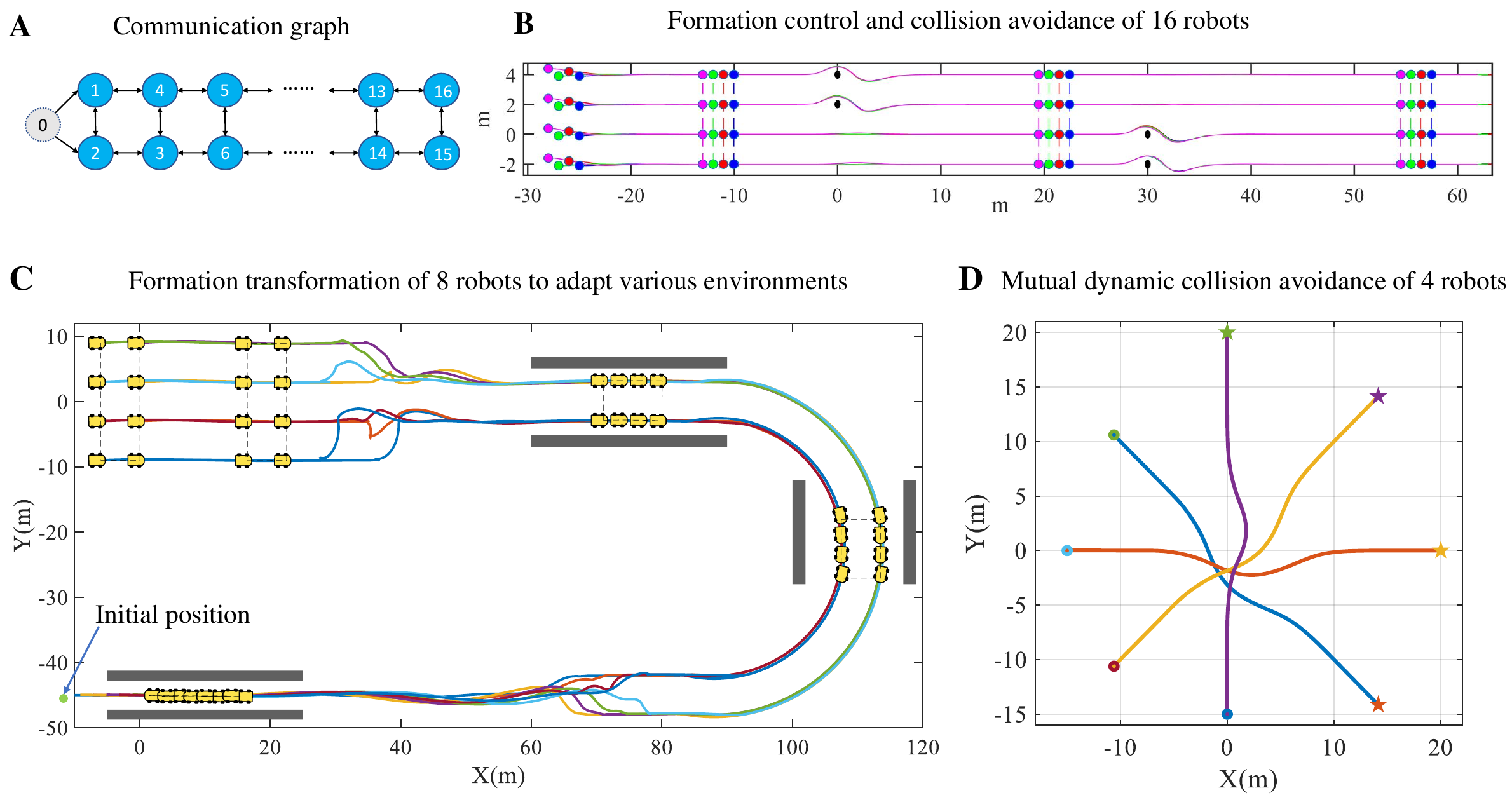}}
				\caption{A: Communication graph of $M=16$ mobile robots, where the arrows indicate the direction of information transmission, and robot 0 is the leading one. B: The path of robots in formation control and collision avoidance under the communication graph, where the black circular areas (0.4m in diameter) represent the obstacles, and the colored lines represent the robots' paths. Meanwhile, the robots in the same column are marked with the same colored dots. C: Transformation of 8 robots to adapt to various environments, where the black rectangles are the obstacles. D: Verification of inter-robot collision avoidance of 4 robots.}
				\label{fig:Communication diagram2} 
			\end{figure*}
			\begin{figure*}[h!]
				\centerline{\includegraphics[scale=0.45]{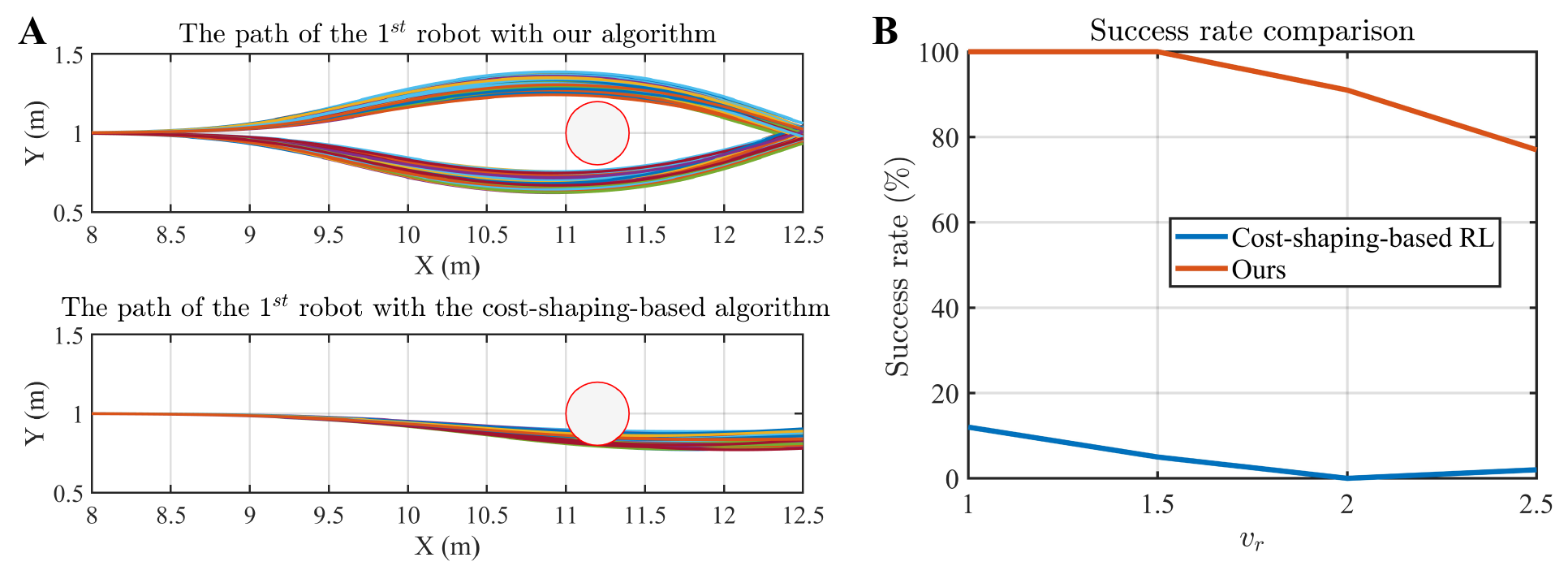}}
				\caption{ Comparison with the cost-shaping-based RL approach for formation control with collision avoidance in 100 repetitive tests. A: The path of the first robot under $v_r=1$ m/s. B: Comparison of the success rate under $v_r=1,\,1.5,\,2,\,2.5$ m/s. 
				}
				\label{fig:collision_compari}  
			\end{figure*}
			%	\begin{figure*}[h!]
				%	% \setlength{\abovecaptionskip}{0.cm}
				%	\centerline{\includegraphics[scale=0.45]{figure/comp_time1}}
				%	\caption{ Visual display of average computational time in the no obstacle scenario, ``I" stands for ``Iteration". 
					%	}
				%	\label{fig:com_time}  
				%\end{figure*}
				%
				\textbf{\emph{Simulation setup and parameters tuning:}} In DLPC, the prediction horizon was set as $N=20$. 
				The critic in~\eqref{eqn:critic-o} and actor in~\eqref{Eqn:actor-o} were chosen as single-hidden-layer neural networks with hyperbolic tangent activation functions. 
				The values of other hyperparameters are listed in Table~\ref{tab:training_hyperparameters}.
				In the training process, the weighting matrices of the actor and critic networks were set as uniformly distributed random values. The simulation tests were performed within a MATLAB environment on a Laptop with Intel Core i9@2.30 GHz.
				%
				%%%%%%%%%%%%%%%%%%%%%%%%%%%%%
				% \begin{figure*}[h!]
					% 	\centerline{\includegraphics[scale=0.5]{figure/experiment_diagram}}
					% 	\caption{The experimental validation scheme. The control policy was trained offline with two robots' kinematic models. Then the local control policy was generalized to two and more wheeled MRS and multirotor drones.}
					% 	\label{fig:experimental-setup} 
					% \end{figure*}
				%	\begin{figure*}[h!]
					%	% \setlength{\abovecaptionskip}{0.cm}
					%	\centerline{\includegraphics[scale=0.4]{figure/drone_experi}}
					%	\caption{State errors of the multirotor drones in Gazebo. We directly deploy the learned control policy to formation control of multirotor drones.  Stage A: Leader in Hover mode; Stage B: Leader speeds up by keyboard control; Stage C: Leader at a constant speed of 2 m/s; Stage D: Change to a new formation shape.}
					%	\label{fig:drone_experi} 
					%\end{figure*}
					\begin{table*}[h!tb]
						\centering \caption{Comparison with DMPC in terms of performance and online computational efficiency.}
						\label{tab:Tab_com1}
						%\vskip 0.2cm
						%
						\begin{threeparttable}
							\scalebox{0.85}{
								% The {|c|c|c|c|c|} define the number of columns.
								% c means centered
								% | defines a vertical line between two columns
								\renewcommand{\arraystretch}{1.2}
								\begin{tabular}{ccccccccccc}
									\toprule
									\multirow{2}{*}{Scenario}& \multirow{2}{*}{Item} &\multirow{2}{*}{Algorithm}&\multirow{2}{*}{Solver}&\multirow{2}{*}{Max. Iteration}&\multicolumn{6}{c}{$M$ (Robot scales)}\\\cline{6-11}
									&&&&& $2$ &$4$&$6$&$16$&$1,000$&10,000\\ \midrule
									\multirow{9}{*}{No obstacle} 
									&\multirow{4}{*}{Cost $V$\tnote{(1)}}&{Our}&  Our &30& $\bm{0.73}$ & $\bm{0.53}$ &  $0.34$&4.18&--&--\\\cline{3-11}%\multirow{2}{*}
									%&&& Our (deploying) &--& $-$ & $-$ &  $-$&&\\\cline{3-10}
									%&&&	& & 30&$0.05$ & $0.29$ &  $6.7$\\\cline{3-10}
									%   &&Our &30& $\bm {0.73}$ & $\bm{0.53}$ &  $\bm {0.34}$\\\cline{3-10}
									%&&\cite{conte2016distributed}& IPOPT &30 & $1.46$ & $1.26$ &  $-$\\\cline{3-10}
									&&\cite{conte2016distributed}\tnote{(2)}& IPOPT &30 & $1.46$ & $1.26$ &  $-$&--&--&--\\\cline{3-11}
									&&\cite{7546918}&fmincon&30 & $0.88$ & $11.9$ &  $-$&--&--&--\\\cline{2-11}%$1159$\\\cline{2-8}
									&\multirow{6}{*}{Ave. comp. time (s)} &\multirow{2}{*}{Our}&  Our  (learning)&30& $\bm{0.02}\tnote{(3)}$ & $\bm{0.052}$ &  $\bm{0.084}$&0.12&1.48&14.57\\\cline{4-11}
									&&& Our (deploying) &30& $\bm {2}$\textbf{e--5} & $\bm {5.2}$\textbf{e--4} &  $\bm{9.6}$\textbf{e--4}&$0.018$&0.05&0.28\\\cline{3-11}
									&&\multirow{2}{*}{\cite{conte2016distributed}}&\multirow{2}{*}{IPOPT}&5 & $0.06$ & $0.76$ &  $4.74$&--&--&--\\\cline{5-11}
									&&	& & 30&$0.1$ & $1.16$ &  $40.2$&--&--&--\\\cline{3-11}
									&&\multirow{2}{*}{\cite{7546918}}&\multirow{2}{*}{fmincon}&5 & $0.08$ & $1.0$ &  $2.76$&--&--&--\\\cline{5-11}
									&&&&30 & $0.4$ & $2.28$ &  $7.2$&--&--&--\\
									\midrule
									\multirow{8}{*}{Collision avoidance} 
									&\multirow{3}{*}{Cost $V$} & Our& Our & 30&${13.5}$ & $\bm {18.8}$ &  ${8.4}$&12.1&--&--\\\cline{3-11}
									&&\cite{conte2016distributed}&IPOPT &30& $\bm {5.0}$ & Inf\tnote{(4)} &  $-$&--&--&--\\\cline{3-11}
									&&\cite{7546918}&fmincon &30& $25.4$ & 61.8&  $-$&--&--&--\\\cline{2-11}
									&\multirow{5}{*}{Ave. comp. time (s)}&	\multirow{2}{*}{Our}& Our (learning)& 30&$\bm {0.02}$ & $\bm {0.056}$ &  $\bm{0.114}$&0.16&1.85&18.9\\\cline{4-11}
									&&& Our (deploying) &30& $\bm {2} $\textbf{e--5} & $\bm {8 }$\textbf{e--4} &  $\bm{1.2 }$\textbf{e--3}&0.025&0.08&0.35\\\cline{3-11}
									&&	\multirow{2}{*}{\cite{conte2016distributed}}& \multirow{2}{*}{IPOPT}  & 5&$0.03$ & $0.17$ &  $0.79$&--&--&--\\\cline{5-11}
									&&	 & & 30&$0.05$ & $0.28$ &  $6.7$&--&--&--\\\cline{3-11}
									&&	\multirow{2}{*}{\cite{7546918}}&\multirow{2}{*}{fmincon}  & 5&$0.35$ & $0.74$ &  $0.46$&--&--&--\\\cline{5-11}
									&&  && 30&$0.68$ & $1.99$ &  $1.2$&--&--&--\\
									\bottomrule
								\end{tabular}
							}	 
							\begin{tablenotes}
								\scriptsize
								\item[(1)] {\color{black}\scalebox{0.75}{$V=1/M\sum_{j=1}^{N_{\rm sim}}r_i(e_{\scriptscriptstyle \mathcal{N}_i}(j),u_i(j))$, $N_{\rm sim}=180$ is the simulation length.}}
								\item[(2)] {\color{black}\scalebox{0.75}{The DMPC algorithm in~\cite{conte2016distributed} was modified with the terminal penalty in~\eqref{Eqn:Lya-mod-o} to adapt to the nonlinear MRS problem.}}
								\item[(3)] {\color{black}\scalebox{0.75}{The overall computational time for solving all the subproblems is collected  at each time instant. The simulations involving 2 to 10,000 robots were conducted with a centralized computation setup on a laptop.}}
								\item[(4)] {\color{black}\scalebox{0.75}{Inf stands for infeasibility issue.}}
							\end{tablenotes}
						\end{threeparttable}
					\end{table*}
					\begin{figure}[h!]
						\centerline{\includegraphics[scale=0.28]{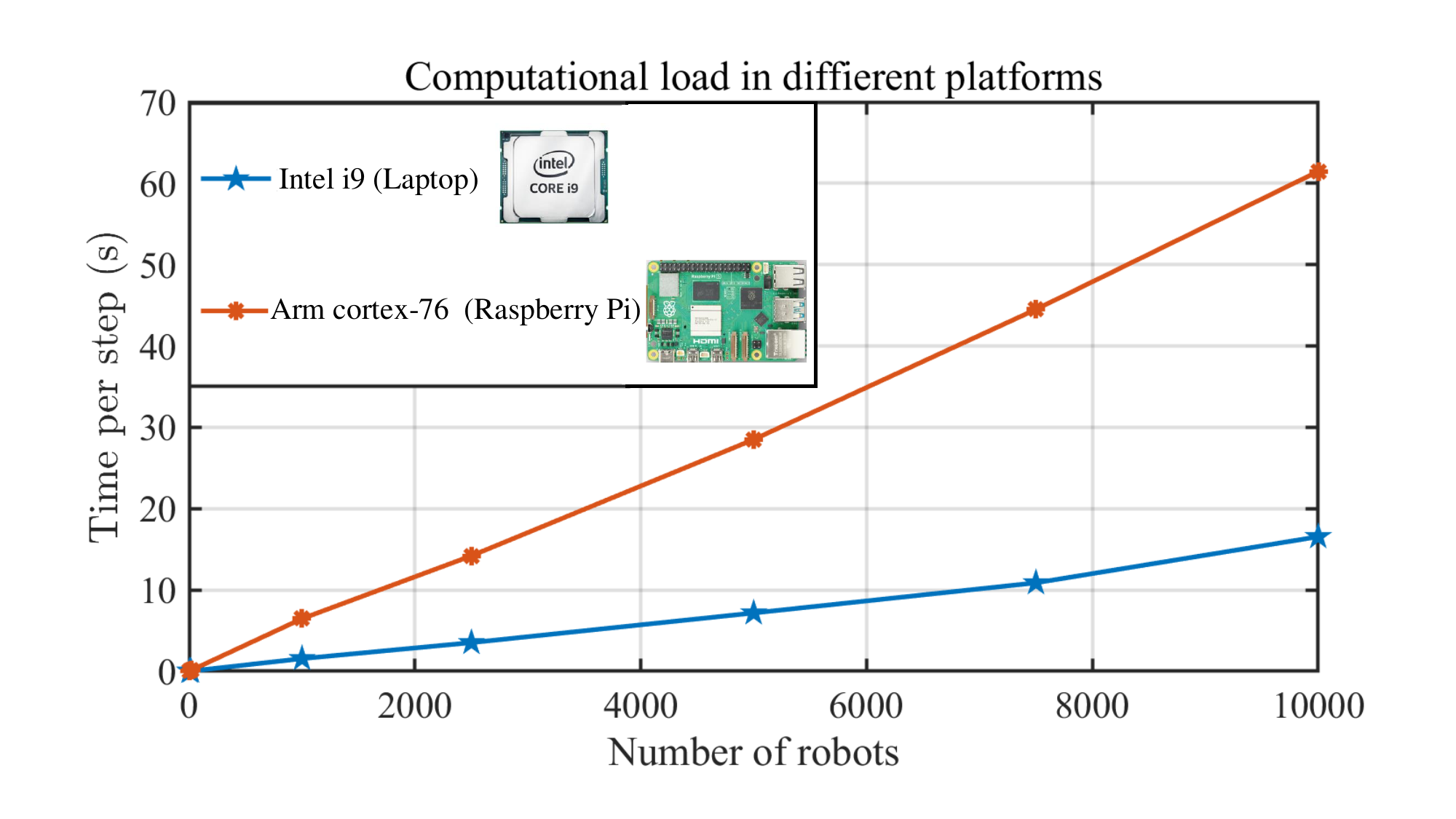}}
						\caption{
							Computational load during online learning within Python at different computing platforms. The average computational time per step within an 8-core Arm processor on a small-scale Raspberry Pi 5 module is about 4 times larger than that with an Intel i9 processor on a Laptop. Also, the computational time in both platforms grows linearly with robot scales. 
						}
						\label{fig:jetson} 
					\end{figure}

					\textbf{\emph{Online policy training with robot scales up to 10,000:}} Firstly, we have verified our approach's learning convergence and closed-loop stability, which is omitted here. Please refer to Appendix~\ref{sec:verification} for implementing details and results. We next show that our approach could efficiently train robots for formation control with robot scales varying from 4 to 10,000 (see Fig.~\ref{fig:trajectory-gen}-A), verifying the scalability in policy training. {\color{black} We adopted a sparse communication topology in all the scenarios from 4 robots to 10,000, where each local robot only received the information from three neighbors at most (including itself). All the weights in the actor and critic networks were initialized with uniformly distributed random values within the range $[0,\,0.1]$ in the first prediction interval. They were successively updated in the subsequent prediction intervals.} As shown in Fig.~\ref{fig:trajectory-gen}-A, our approach could successfully train the DMPC policies to drive the robots to achieve the predefined formation shape from a disordered initialization. Notably, the transient periods in formation generation are only about 3 s even in the scenario with robot scales $M=10,000$.  Moreover, the average computational time for solving the overall optimization problem at each time step grows linearly with robot scales (see Table~\ref{tab:Tab_com1}), which is 0.02 s for $M=2$ and 14.57 s for $M=10,000$. The results demonstrate the effectiveness and scalability of our approach in online policy generation.  %a rectangular formation shape of 4 rows and 4 columns, in the meanwhile avoiding the obstacles on the path (see Fig.~\ref{fig:Communication diagram1}),  where the desired distances between neighboring robots in the same row and column were 2m and 1m respectively. The communication graph between the local robots is shown in Panel A of Fig.~\ref{fig:Communication diagram1}.  All the obstacles to be avoided were circular objects with a diameter of 0.5m (see again Fig.~\ref{fig:Communication diagram1}). The centers of the four obstacles were $(0,4),\,(0,2),\,(30,-0),\,(30,-2)$. The  constraint for collision avoidance was of type $\mathcal{E}_i=\{(p_{x,i},p_{y,i})|\|(p_{x,i},p_{y,i})-c_i\|\geq d_i\}$, where $d_i=0.25$, $c_i$ is the center of the obstacle. In the collision avoidance, we set $P_i=Q_i$ and $\mathcal{E}_{f,i}=\mathcal{E}_i$. 
					
					\textbf{\emph{Transferability from 2 robots to 1,000:}}
					We verified generalizability by transferring our offline learned policy with 2 robots directly to multiple robots with scales up to 1,000 (see Fig.~\ref{fig:trajectory-gen}-B). {\color{black}The training was performed for 2 robots' formation control in a straight-line formation scenario. The learned weighting matrix of the actor for the first robot was $W_{a,1}=[w_{1} \quad w_{2}]^{\top}$, where $$w_{1}\hspace{-1.5mm}=\hspace{-1.5mm}\begin{bmatrix}
							0.65&\hspace{-2.5mm}0.4&\hspace{-2.5mm}0.3&\hspace{-2.5mm}1.58\\
							-0.26&\hspace{-2.5mm}0.47&\hspace{-2.5mm}1.2&\hspace{-2.5mm}-0.16
						\end{bmatrix}, w_{2}\hspace{-1.2mm}=\hspace{-1.5mm}\begin{bmatrix}
							-0.01&\hspace{-2.5mm}-0.2&\hspace{-2.5mm}0.05&\hspace{-2.5mm}-0.23\\
							0.05&\hspace{-2.5mm}0.15&\hspace{-2.5mm}0.12&\hspace{-2.5mm}-0.19
						\end{bmatrix};
						$$ here $w_1$ and $w_2$ associated with to the error states of the first robot and its connected neighbor in $e_{\scriptscriptstyle \mathcal{N}_1}$, respectively. The weighting matrix was then directly used to construct control policies for formation control with 4, 200, and 1,000 robots in both straight-line and circular formation scenarios.
						Note that in policy deployment for 4, 200, and 1,000 robots, the first robot has two neighbors while others have three neighbors (see Panel B in Fig.~\ref{fig:trajectory-gen}), unlike the 2-robot scenario where each robot has two neighbors. Therefore, the weighting matrix used for policy deployment was constructed as $W_{a,1}=[w_{1}\quad w_{2}]^{\top}$ for the first robot and $W_{a,i}=[w_{1}\quad w_{2}\quad w_{2}]^{\top}$ for the other robots, which means that the weighting matrix $w_2$, corresponding to the connected neighbor in the 2-robot scenario, was repeatedly used in the 4, 200, and 1,000 robot scenarios.} 
					As displayed in Panel B of Fig.~\ref{fig:trajectory-gen}, the transferred policy stabilizes the formation control system under various robot scales. The results represent a substantial reduction in computational load by only training a limited number of robots with a similar control goal. 
					
					\textbf{\emph{Policy training and deployment under collision avoidance constraints:}} We first show that our approach could efficiently train 16 robots to form a rectangular shape of 4 rows and 4 columns while avoiding the obstacles on the path (see Panels A and B in Fig.~\ref{fig:Communication diagram2}),  where the desired distances between neighboring robots in the same row and column were 1 m and 2 m respectively. The communication graph between the local robots is shown in Panel A of Fig.~\ref{fig:Communication diagram2}.  All the obstacles to be avoided were circular objects with a diameter of 0.4 m. The centers of the four obstacles were $(0,4),\,(0,2),\,(30,-0),\,(30,-2)$. The constraint for collision avoidance was of type $\mathcal{E}_i=\{(p_{x,i},p_{y,i})|\|(p_{x,i},p_{y,i})-c_i\|\geq d_i\}$, where $d_i=0.2$, $c_i$ is the center of the obstacle. The constructed recentered barrier functions for $\mathcal{E}_i$ were centered at a circle with $\|(p_{x_c,i},p_{y_c,i})-c_i\|\rightarrow +\infty$. In the collision avoidance, we set $P_i=Q_i$ and $\mathcal{E}_{f,i}=\mathcal{E}_i$.

					\begin{figure*}[h!]
						\centerline{\includegraphics[scale=0.48]{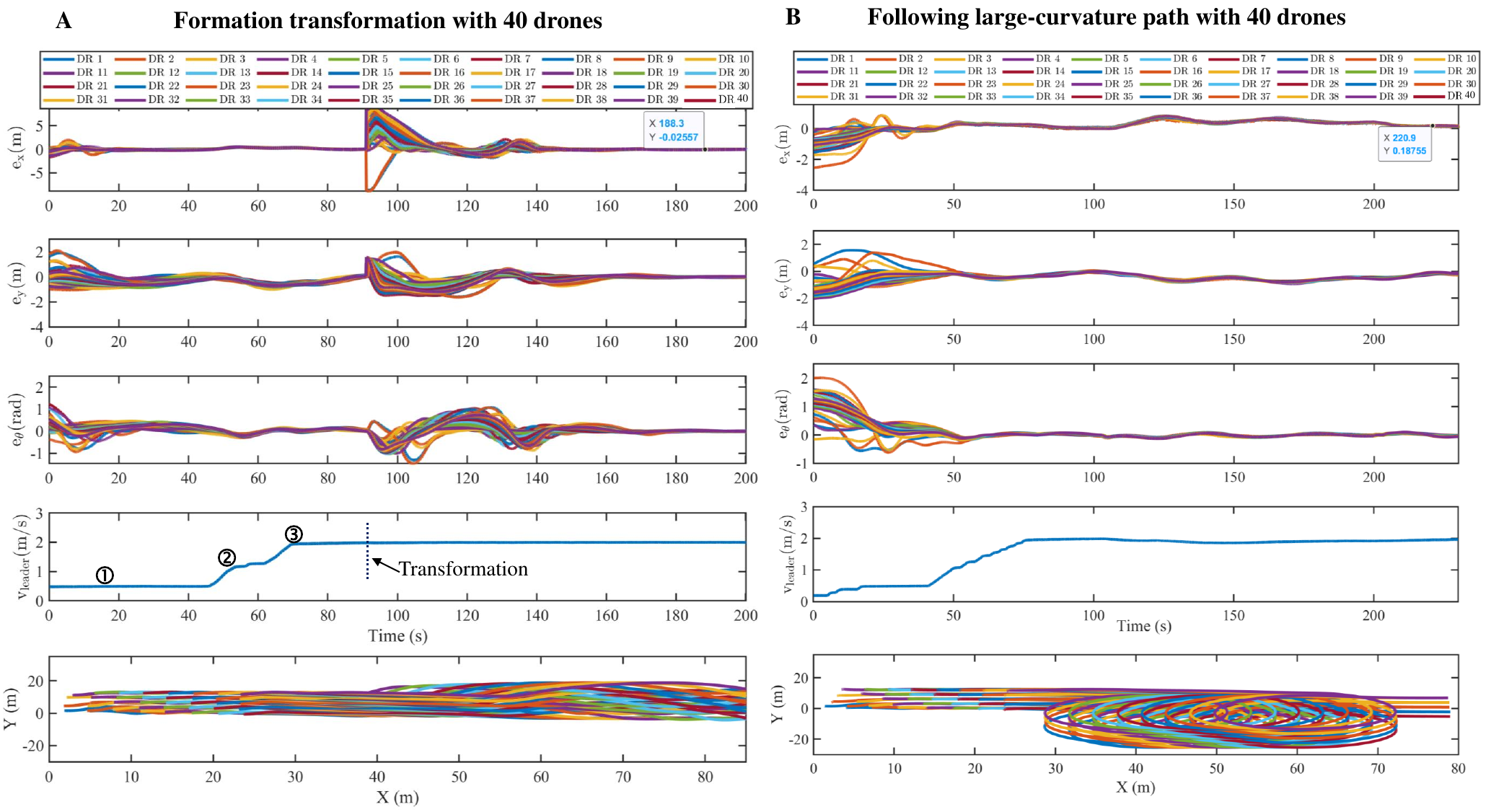}}
						\caption{State errors and paths of the multirotor drones in Gazebo. We directly deployed the learned control policy to the formation control of multirotor drones with $M=40$.  Stage \textcircled{1}: Leader in hover mode; Stage \textcircled{2}: Leader speeds up with keyboard control; Stage \textcircled{3}: Leader at a constant speed of 2 m/s. %; Stage D: Change to a new formation shape.
						}
						\label{fig:drone_experi} 
					\end{figure*}
					\begin{figure}[h!]
						\centerline{\includegraphics[scale=0.52]{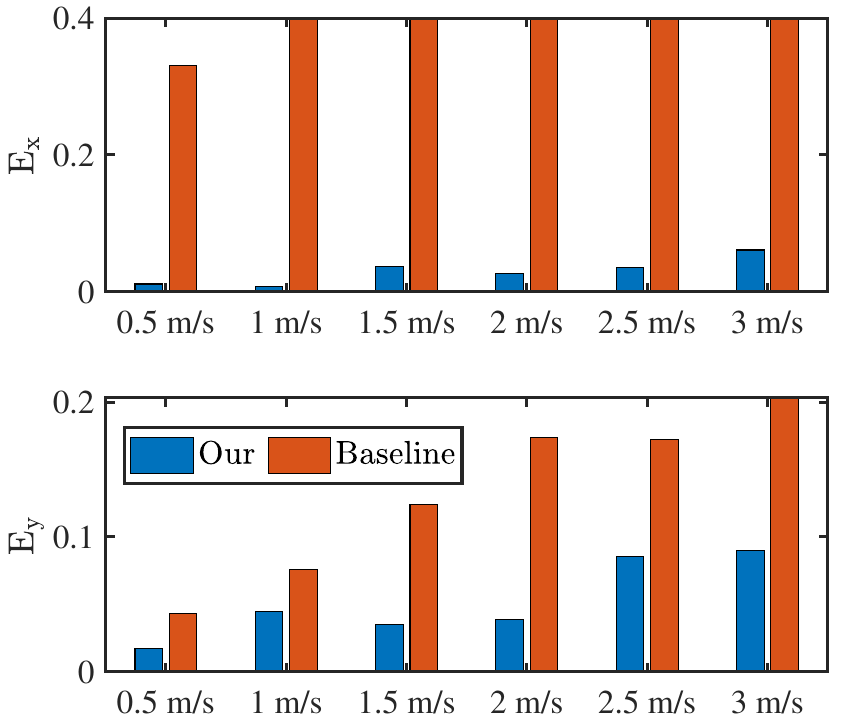}}
						\caption{Formation control of six multirotor drones: Comparison in state errors $e_x$ and $e_y$ with the baseline controller under leader's speed at $0.5$ m/s, $1$ m/s, $1.5$ m/s, $2$ m/s, $2.5$ m/s, $3$ m/s, where $E_{\star}=1/(N_{\rm sim}M)\sum_{i=1}^M|\sum_{j=1}^{N_{\rm sim}}e_{\star}|$, $\star=x,\,y$, $N_{\rm sim}$ is the length of the simulation. %; Stage D: Change to a new formation shape.
						}
						\label{fig:comparison_multirotor} 
					\end{figure} 
					\begin{figure}[h!]
						\centerline{\includegraphics[scale=0.37]{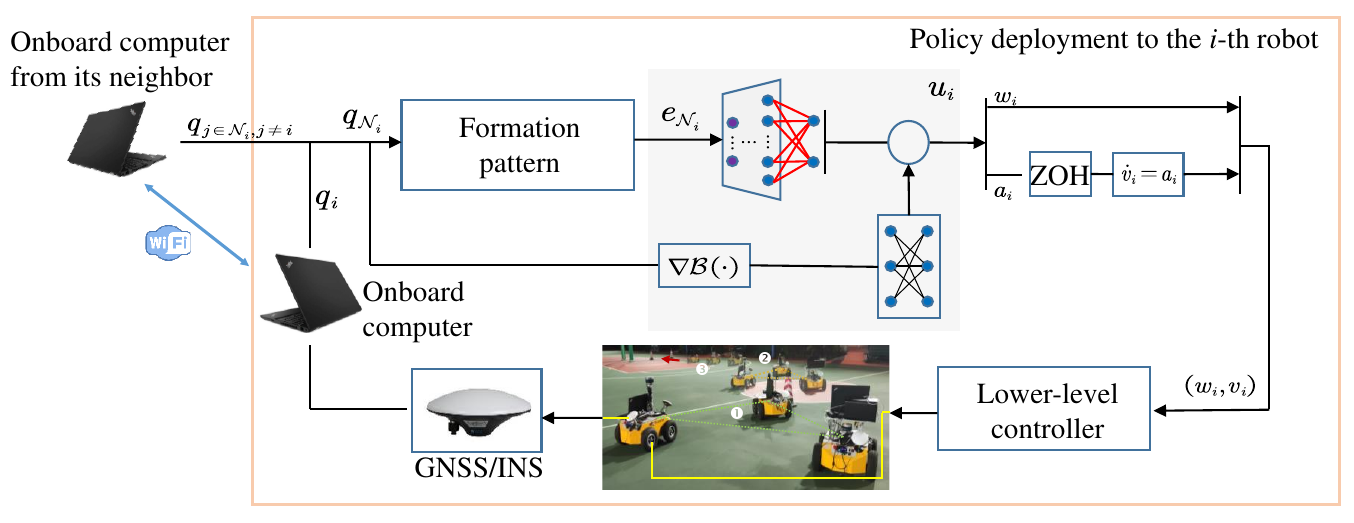}}
						\caption{ Block diagram for deploying the learned control policy to the $i$-th local robot with neighbor-to-neighbor communication, ZOH=zero-order holder. The gray block encloses the learned control policy. The experimental scenario presented in the picture involves the coordination of wheeled robots for formation control and collision avoidance.  In stage \textcircled{1}, the robots actively avoid collisions with obstacles. In stage \textcircled{2}, the robots recover their formation after successfully avoiding the collision. In stage \textcircled{3}, the formation is strategically transformed to navigate a narrow corridor.}
						\label{fig:policy_vehicle} 
					\end{figure}
					\begin{figure*}[h!]\centerline{\includegraphics[scale=0.42]{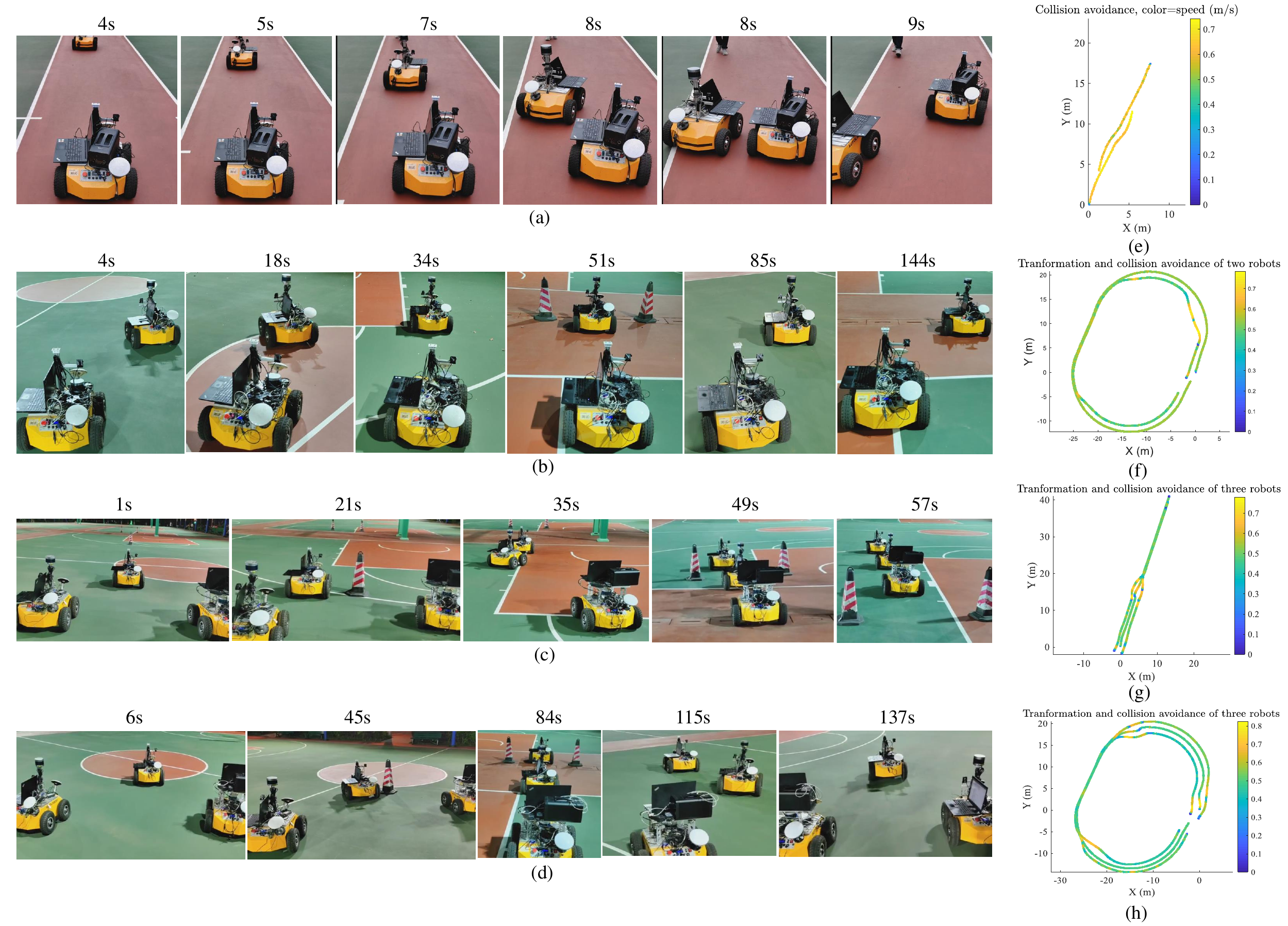}}
						\caption{Snapshots and trajectories of MRS' formation control and collision avoidance: (a) Inter-robot collision avoidance; (b) Formation control and transformation for two robots; (c) and (d) Formation transformation  and collision avoidance for three robots in different scenarios; (e), (f), (g), and (h) show the trajectory of MRS associated with (a), (b), (c), and (d).}
						\label{fig:path-profile} 
					\end{figure*}
					
					The simulation results in Fig.~\ref{fig:Communication diagram2} and Table~\ref{tab:Tab_com0} show that the mobile robots could achieve a predefined formation shape from a disordered initialization.  In addition, they effectively avoid obstacles encountered along the path and restore the shape of the formation after collision avoidance. Eventually, the formation error of each local robot converges to the origin. 
					
					To assess the adaptability of our approach in different constrained environments, we directly deployed the learned policy for formation transformation with eight robots. The simulation results are illustrated in Panel C of Fig.~\ref{fig:Communication diagram2}, which verifies our approach's ability in the rapid formation transformation. We also verified our approach on inter-robot collision avoidance tests of 4 mobile robots, where a joint constraint was formed for each robot $i$   of type $\mathcal{E}_i=\{(p_{x,i},p_{y,i})|\|(p_{x,i},p_{y,i})-(p_{x,j},p_{y,j})\|\geq d_i,\,\forall\,j\in\mathbb{N}_1^4\}$. The simulation results are displayed in Panel D of Fig.~\ref{fig:Communication diagram2}, verifying our approach's effectiveness in coping with the type of joint inter-robot collision avoidance constraints.

					\textbf{\emph{Comparison with cost/reward-shaping-based RL approaches:}}  The cost function in the cost-shaping-based RL was shaped according to~\cite{Zhang9812083} with the same barrier functions used in our approach. Also, the parameters used in cost-shaping-based RL were fine-tuned for a fair comparison. We performed 100 repetitive online training tests for our approach and cost-shaping-based RL on the formation control of 2 robots with collision avoidance. The simulation results in Fig.~\ref{fig:collision_compari} show that our method outperforms the cost-shaping-based RL approach regarding control safety due to the unique force field-inspired control policy design in~\eqref{Eqn:control}. {\color{black}Note that when the velocity exceeds 1.5 m/s, the success rate of our approach gradually decreases as the velocity grows. This is due to the violation of condition~\eqref{learning rate-inequa} in Appendix~\ref{sec:saferl} (see also Theorem 8 of Appendix B in the attached ``auxiliary-results.pdf") during the learning process. The underlying cause is the rapid growth of the barrier function's gradient as the velocity increases. Deriving a more relaxed safety guarantee condition for fast and safe control will be a focus of future research.} %; the adopted incremental learning mechanism might not be able to refine the weights of actor and critic timely to account for the change of value functions affected by the gradient change. This issue will be addressed in future investigations by designing a new policy learning mechanism that reduces the influence of gradient changes on policy learning.
					%}
				%
				
				\textbf{\emph{Comparison with DMPC using numerical solvers:}} We compared our approach with DMPC on various robot scales. The DMPC approaches in~\cite{conte2016distributed} and~\cite{7546918} were adopted for comparison and designed to adapt to the nonlinear MRS problem. The parameters $Q_i$ and $R_i$ in the comparative DMPC approaches were chosen similarly to ours. %The terminal penalty $P_i$ and terminal set $\mathcal{E}_{f,i}$ for each robot $i$ are computed according to~\eqref{Eqn:Lya-mod}. 
				In the comparison, the prediction horizon was chosen as $N=10$ to reduce the computational load, especially for DMPC. As~\cite{conte2016distributed} was initially developed for linear interconnected systems, it was modified with the terminal penalty matrix in~\eqref{Eqn:Lya-mod-o} to guarantee stability under the nonlinear model constraint~\eqref{Eqn:LL-nei}. The DMPC algorithm was implemented with the ALADIN-$\alpha$ toolbox~\cite{engelmann2020aladin} and the CasADi toolbox~\cite{andersson2019casadi} and using the IPOPT solver, while the nonlinear DMPC algorithm~\cite{7546918} was implemented in Matlab using the \emph{fmincon} solver. In contrast to~\cite{conte2016distributed} and~\cite{7546918}, our approach is library-free and does not require nonlinear optimization solvers.  
				
				As shown in Table~\ref{tab:Tab_com1}, our approach shows a significant advantage in computational efficiency. Moreover, our approach results in lower cumulative cost values than DMPC in the no-obstacle scenario. This result is counterintuitive but reasonable. As acknowledged within the MPC community, optimality is only achievable in the prediction interval, while stability, rather than optimality, can be established in a closed-loop manner~\cite{rawlings2009model}. The above results are further validated by the cumulative cost function values $J_c=\sum_{k=1}^{N_{\rm sim}}J(k)$ collected with $N_{\rm sim}=180$ for all the prediction intervals. In a scenario with $M=2$, this value is 8.9 with DMPC, which is lower than that achieved with our approach, which is 16.9. 
				These findings suggest that although DMPC results in lower cost function values in each prediction interval, our approach achieves superior closed-loop control performance. This can be attributed to the analytic policy structure and the successive policy learning mechanism within different prediction intervals.

				\textbf{\emph{Computational load in different platforms:}}
				We also tested our approach within a Python environment on different computing platforms (see Fig.~\ref{fig:jetson}), showing that our approach could be efficiently deployed to small-scale modules such as Raspberry Pi 5 with an Arm Cortex-A76 processor. The computational load is only about 4 times higher than that using a powerful Intel i9 processor. Also, in both the i9 and Arm processors, {\color{black}the computational time for solving all the subproblems grows linearly with the robot scales.}
				\begin{figure}[h!]
					\centerline{\includegraphics[scale=0.4]{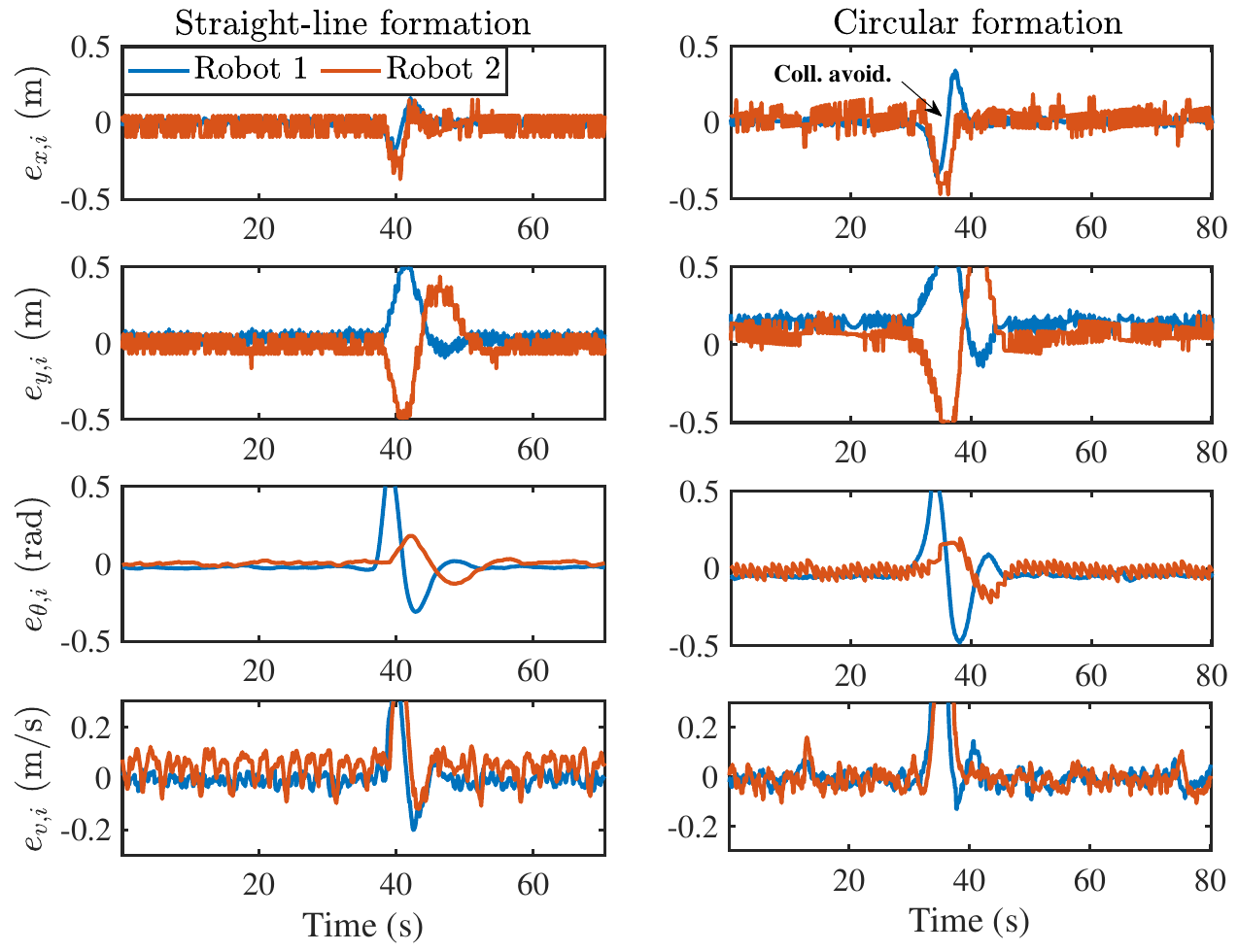}}
					\caption{ Experimental results: State errors of the mobile robots in the straight-line formation and circular formation scenarios.}
					\label{fig:state-error} 
				\end{figure}
				\subsection{Policy Deployment to Multirotor Drones in Gazebo}
				
				To further verify our algorithm's transferability and robustness, we directly deploy the learned distributed policy to formation control of multirotor drones in Gazebo. We show the scalability of our approach by performing tests on different scales of drones (in particular, $M=6,\,18,\,40$) and demonstrate the effectiveness by comparing it with a baseline formation controller.
				
				The simulation was implemented in Python using the XTdrone platform~\cite{xiao2022implementation}. The platform utilizes PX4 as flight control software and Gazebo as the simulation environment. {\color{black}Our learned policy only accounts for the formation control in the horizontal direction, while height control is based on a baseline controller~\cite{xiao2022implementation}. Variations in rolling and pitch angles were treated as external disturbances to assess the robustness of our approach.}   In the experiment, our deployed control policy can realize stabilizing formation control and transformation in the formation control scenario of 6, 18, and 40 drones. Please see Fig.~\ref{fig:drone_experi} for the detailed variation of state errors of drones under $M=40$, while other experimental results are given in Appendix~\ref{sec:verification}.  In the formation transformation scenario (Scenario A in Fig.~\ref{fig:drone_experi}), the state errors approach the origin together with the speed-up process of the leader and then recover promptly from a short transient formation transformation. The state errors remain close to the origin in the subsequent scenario with a large-curvature path (Panel B in Fig.~\ref{fig:drone_experi}).
				
				We also compare the performance of our approach with our previously developed baseline feedback formation controller~\cite{xiao2022implementation} in a formation scenario of six multirotor drones. The results demonstrate the superior formation performance of our approach. Detailed numerical comparisons of the state errors $e_x$ and $e_y$ are shown in Fig.~\ref{fig:comparison_multirotor}. {\color{black}These findings highlight the sim-to-real transferability of our learned control policy and the effectiveness of our proposed approach for the formation control of multirotor drones.} 
				
				\subsection{Real-World Experiments on Wheeled MRS}
				
				We tested our proposed algorithm on several real-world wheeled mobile robots for formation control with collision avoidance. 
				The control policies were learned offline with two robots' kinematics and deployed to real-world robots across different scales. Through the experiment, we want to show that the off-line learned policy: a) could be generalized to control two real-world robots; b) could be further transferred to control more robots.  
				Each robot in the experiment was equipped with a laptop running the Ubuntu operating system. Considering the computational lightness of the implementation, the laptops utilized in the experiment are interchangeable with other computing platforms, such as the Raspberry Pi 5. The sampling interval was set to $\Delta t = 0.1$ seconds. At each sampling instant, the integrated satellite and inertial guidance positioning module onboard measured the local state $q_i$ for each robot $i$. A wireless network transmitted the measured $q_i$ and the corresponding reference $q_r$, among the laptops of the neighbors (see Fig.~\ref{fig:policy_vehicle}).  
				In each laptop, the control input was generated from the actor network in real-time using the measured state information, which was regarded as the reference to be followed by the lower-level control (see again Fig.~\ref{fig:policy_vehicle}). 
				
				\begin{table}[h!tb]
					\centering \caption{Maximum absolute error in the experiments (corresponds to Fig.~\ref{fig:state-error}).}
					\label{tab:Tab_com3}
					%\vskip 0.2cm
					%
					\scalebox{0.8}{
						% The {|c|c|c|c|c|} define the number of columns.
						% c means centered
						% | defines a vertical line between two columns
						\renewcommand{\arraystretch}{1.2}
						\begin{tabular}{ccccccc}
							%			\toprule 
							\toprule
							Scenario & Robot& Case&$e_{x,i}$(m) &$e_{y,i}$(m)& $e_{\theta,i}$(rad)&$e_{v,i}$(m/s)\\ \midrule
							\multirow{4}{*}{Linear} &
							\multirow{2}{*}{1st robot} &Coll. avoid. & 0.16&   0.57&   0.61&    0.40\\
							&&Form. keep. & 0.04  &  0.10  & 0.05   & 0.09\\\cline{2-7}%
							&\multirow{2}{*}{2nd robot} &Coll. avoid. & 0.15 &  0.44&  0.18& 0.49\\
							&&Form. keep. & 0.04  &  0.06 &   0.01  &  0.17\\
							\midrule
							\multirow{4}{*}{Circular} 
							&\multirow{2}{*}{1st robot} &Coll. avoid. &    0.35& 0.81&0.58&0.63
							\\
							&&Form. keep. &   0.05 &   0.21&  0.11&    0.12\\\cline{2-7}%\vspace{1mm}
							&\multirow{2}{*}{2nd robot} &Coll. avoid. & 0.47&0.66& 0.22& 0.72\\
							&&Form. keep. &0.14&  0.17&  0.08 &   0.12\\
							\bottomrule
						\end{tabular}
					}	 
				\end{table}	
				In the experiment, we first tested the mutual collision avoidance capability between two robots using the learned control policy. Please see Panels (a) and (e) of Fig.~\ref{fig:path-profile} for the snapshots and the trajectory of the two robots.  We then directly deployed the control policy for the realizing formation control and collision avoidance of two- and three robots (see Panels (b), (c), (d), and (f), (g), (h) of Fig.~\ref{fig:path-profile} for the snapshots and the associated trajectories). In the case of three robots, as shown in Panels (d) and (h) of Fig.~\ref{fig:path-profile},  the three robots initially formed a triangle formation, then avoided collision on the path, and transformed into a straight-line formation to pass the narrow passage and into a non-equilateral triangle formation after that.  
				Furthermore, we quantitatively assessed the control policy's performance by measuring the robot error states in standard straight-line and circular scenarios with collision avoidance. The numerical measures of the error states are depicted in Fig.~\ref{fig:state-error} and summarized in Table~\ref{tab:Tab_com3}.  These results demonstrated that wheeled mobile robots could flexibly maintain formations, successfully avoid dynamic obstacles, and accurately follow desired trajectories. The average computational time per step for the online policy deployment of each robot is about 0.02ms.
				
				In general, our experimental evaluation has verified two significant features of our approach. First, the control policies learned from simulation exhibit strong sim-to-real transferability. Second, the learned policies could also be directly deployed to real-world MRS across different scales, enabling scalability for optimization-based control of large-scale MRS. %The results demonstrate our approach's capability for achieving formation control and collision avoidance in MRS and its potential for scalability in distributed control of nonlinear systems.
				\subsection{Discussions and Limitations}  \emph{Discussions}: Our results suggest that our DLPC approach is suitable for optimal cooperative control of large-scale MRS. Our method centers around a computationally fast and efficient policy learning algorithm to generate explicit DMPC policies in a closed-loop manner. This algorithm is executed with a distributed incremental actor-critic learning implementation, enabling online policy learning with robot scales up to 10,000 and rapid policy deployment with scales up to 1,000, offering theoretical insights and practical value for optimization-based multirobot control with strong scalability. Furthermore, our approach is also extended to address the challenge of safe policy learning under state and control constraints, employing a force field-inspired policy structure informed by interior point optimization techniques~\cite{boyd2004convex}.
				
				The comparative analysis indicates that although our policy learning algorithm may yield suboptimal control policies within each prediction interval, it results in superior closed-loop control performance compared with numerical DMPC methods (see Table~\ref{tab:Tab_com1}). This improvement can be attributed to our analytical policy structure and the successive policy learning mechanism applied across different prediction intervals. This observation is consistent with MPC theory, which acknowledges that optimality is achievable within the prediction horizon, while stability, rather than optimality, can be established in a closed-loop manner \cite{rawlings2009model}.
				
				\emph{Limitations}: This article focuses on nonlinear cooperative control problems with quadratic cost function formulations to ensure stability guarantees. It does not address, but is not limited to, decision-making problems with non-quadratic cost forms, which we leave for future investigation.  Although we have demonstrated that our approach is robust to bounded disturbances, including modeling uncertainties~\cite{zhang2022robust11}, our approach relies on dynamical models for policy learning. Extensions to model-free designs will be considered in the future.
				{\color{black}The theoretical and experimental results of our approach were obtained in a static communication network, which could change when robots occasionally leave or join the network~\cite{farina2018hierarchical}. A future direction is to extend our approach to support plug-and-play operations in time-varying communication networks while also ensuring robustness against possible communication delays by resorting to the Lyapunov-Krasovskii function~\cite{seuret2015complete}.}

				%\footnote[1]{For video results, please see \href{https://youtube.com/playlist?list=PLPE5-2sIdTliwcptTGRRN6UoF-J6mlT8N}{https://youtube.com/playlist?list=PLPE5-2sIdTliwcptTGRRN6UoF-J6mlT8N.}}
				
				%\subsection{Implementing Issues and Discussions}
				%Firstly, the local model must be constructed to form the model type~\eqref{Eqn:LL-nei} (only direct interconnections among the states are allowed), like the obtained model~\eqref{Eqn:discrete-time formation tracking error model}, which is not restrictive since most MRS models can be transformed to satisfy~\eqref{Eqn:LL-nei}. Such a model structure allows us to design a sparse control structure associated with cost function~\eqref{Eqn:re-cost}; also, the mutual influences among the neighboring control policies can be limited in the learning process. %, when compared with~\cite{abouheaf2014multi}.    
				%Secondly, it is suggested that the initialized weights $W_{c,i}$ and $W_{a,i}$  are made small to obtain satisfactory control performance since the tuning parameter $\mu$  is usually selected smaller than the entries of $Q$ and $R$.
				%Thirdly, the learning rates must be chosen properly in order to ensure condition~\eqref{learning rate-inequa}. Such a choice can avoid online learning divergence caused by large barrier function values in collision avoidance. Note that these hyperparameters are dynamically tuned in different simulation experiments stated above. %As witnessed in Fig.~\ref{fig:collision_compari}, our approach can significantly improve control safety compared to cost-shaping-based RL approaches due to the adopted safety-related term in~\eqref{Eqn:control}. 
				%
				\section{Conclusions}
				This article has proposed a distributed learning predictive control framework for real-time optimal control of large-scale MRS. Our approach generates DMPC's closed-loop control policies through a computationally fast and efficient distributed policy learning approach.  By implementing the policy learning algorithm in a fully distributed manner, we enable fast online learning of control policies for MRS, with scales up to 10,000 robots.  In the knowledge-sharing aspect, control policies learned with 2 robots exhibit performance guarantees when applied to MRS with scales up to 1,000.  Furthermore, the policies learned from the simulation work very well on mobile wheeled vehicles across different scales in the real world.
				Theoretical guarantees have been provided for the convergence and safety of policy learning and the stability and robustness of the closed-loop system. 
				In summary, our work represents a significant advancement toward achieving fast and scalable nonlinear optimal control of large-scale MRS by a distributed policy learning approach and paves the way for applying distributed RL to the safety-critical control of MRS.
				Future directions of our approach include, but are not limited to, model-free policy learning extension, {\color{black}policy learning and deploying under time-varying communication networks}, and multiagent decision-making with more general forms of cost functions.
				
				%\bibliographystyle{plain}
				%\bibliography{lpc,ref}
				\section*{Multimedia Material}
				Source codes for implementing our method are available at \href{https://github.com/xinglongzhangnudt/policy-learning-for-distributed-mpc}{https://github.com/xinglongzhangnudt/policy-learning-for-distributed-mpc}. 
				Additional qualitative results and videos are available at \href{https://sites.google.com/view/pl-dpc/}{https://sites.google.com/view/pl-dpc/}.
				
				%\bibliographystyle{plain}
				%\bibliography{ref}
				\bibliographystyle{Bibliography/IEEEtranTIE}
				\bibliography{Bibliography/IEEEabrv,ref}

% Generated by IEEEtran.bst, version: 1.12 (2007/01/11)
\begin{thebibliography}{10}
\providecommand{\url}[1]{#1}
\csname url@samestyle\endcsname
\providecommand{\newblock}{\relax}
\providecommand{\bibinfo}[2]{#2}
\providecommand{\BIBentrySTDinterwordspacing}{\spaceskip=0pt\relax}
\providecommand{\BIBentryALTinterwordstretchfactor}{4}
\providecommand{\BIBentryALTinterwordspacing}{\spaceskip=\fontdimen2\font plus
\BIBentryALTinterwordstretchfactor\fontdimen3\font minus
  \fontdimen4\font\relax}
\providecommand{\BIBforeignlanguage}[2]{{%
\expandafter\ifx\csname l@#1\endcsname\relax
\typeout{** WARNING: IEEEtran.bst: No hyphenation pattern has been}%
\typeout{** loaded for the language `#1'. Using the pattern for}%
\typeout{** the default language instead.}%
\else
\language=\csname l@#1\endcsname
\fi
#2}}
\providecommand{\BIBdecl}{\relax}
\BIBdecl

\bibitem{quan2023robust}
L.~Quan, L.~Yin, T.~Zhang, M.~Wang, R.~Wang, S.~Zhong, X.~Zhou, Y.~Cao, C.~Xu,
  and F.~Gao, ``Robust and efficient trajectory planning for formation flight
  in dense environments,'' \emph{IEEE Transactions on Robotics}, 2023.

\bibitem{Xuan2023}
X.~Wang, S.~Mou, and B.~D.~O. Anderson, ``Consensus-based distributed
  optimization enhanced by integral feedback,'' \emph{IEEE Transactions on
  Automatic Control}, vol.~68,
  \href{http://dx.doi.org/10.1109/TAC.2022.3169179}{DOI
  10.1109/TAC.2022.3169179}, no.~3, pp. 1894--1901, 2023.

\bibitem{farina2018hierarchical}
M.~Farina, X.~Zhang, and R.~Scattolini, ``A hierarchical multi-rate {MPC}
  scheme for interconnected systems,'' \emph{Automatica}, vol.~90, pp. 38--46,
  2018.

\bibitem{hujinwen2020}
J.~Hu, H.~Zhang, L.~Liu, X.~Zhu, C.~Zhao, and Q.~Pan, ``Convergent multiagent
  formation control with collision avoidance,'' \emph{IEEE Transactions on
  Robotics}, vol.~36, \href{http://dx.doi.org/10.1109/TRO.2020.2998766}{DOI
  10.1109/TRO.2020.2998766}, no.~6, pp. 1805--1818, 2020.

\bibitem{Fathian2021}
K.~Fathian, S.~Safaoui, T.~H. Summers, and N.~R. Gans, ``Robust distributed
  planar formation control for higher order holonomic and nonholonomic
  agents,'' \emph{IEEE Transactions on Robotics}, vol.~37,
  \href{http://dx.doi.org/10.1109/TRO.2020.3014022}{DOI
  10.1109/TRO.2020.3014022}, no.~1, pp. 185--205, 2021.

\bibitem{liu2018}
Y.~Liu, J.~M. Montenbruck, D.~Zelazo, M.~Odelga, S.~Rajappa, H.~H. Bülthoff,
  F.~Allgöwer, and A.~Zell, ``A distributed control approach to formation
  balancing and maneuvering of multiple multirotor {UAV}s,'' \emph{IEEE
  Transactions on Robotics}, vol.~34,
  \href{http://dx.doi.org/10.1109/TRO.2018.2853606}{DOI
  10.1109/TRO.2018.2853606}, no.~4, pp. 870--882, 2018.

\bibitem{ren2020}
Y.~Ren, S.~Sosnowski, and S.~Hirche, ``Fully distributed cooperation for
  networked uncertain mobile manipulators,'' \emph{IEEE Transactions on
  Robotics}, vol.~36, \href{http://dx.doi.org/10.1109/TRO.2020.2971416}{DOI
  10.1109/TRO.2020.2971416}, no.~4, pp. 984--1003, 2020.

\bibitem{Santilli2022}
M.~Santilli, M.~Franceschelli, and A.~Gasparri, ``Dynamic resilient containment
  control in multirobot systems,'' \emph{IEEE Transactions on Robotics},
  vol.~38, \href{http://dx.doi.org/10.1109/TRO.2021.3057220}{DOI
  10.1109/TRO.2021.3057220}, no.~1, pp. 57--70, 2022.

\bibitem{Qingbiao}
Q.~Li, F.~Gama, A.~Ribeiro, and A.~Prorok, ``Graph neural networks for
  decentralized multi-robot path planning,'' in \emph{2020 IEEE/RSJ
  International Conference on Intelligent Robots and Systems (IROS)},
  \href{http://dx.doi.org/10.1109/IROS45743.2020.9341668}{DOI
  10.1109/IROS45743.2020.9341668}, pp. 11\,785--11\,792, 2020.

\bibitem{Liangming}
L.~Chen, H.~G. de~Marina, and M.~Cao, ``Maneuvering formations of mobile agents
  using designed mismatched angles,'' \emph{IEEE Transactions on Automatic
  Control}, vol.~67, \href{http://dx.doi.org/10.1109/TAC.2021.3066388}{DOI
  10.1109/TAC.2021.3066388}, no.~4, pp. 1655--1668, 2022.

\bibitem{hou2021distributed}
B.~Hou, S.~Li, and Y.~Zheng, ``Distributed model predictive control for
  reconfigurable systems with network connection,'' \emph{IEEE Transactions on
  Automation Science and Engineering}, vol.~19, no.~2, pp. 907--918, 2022.

\bibitem{todorovic2020distributed}
U.~Todorovi{\'c}, J.~R.~D. Frejo, and B.~De~Schutter, ``Distributed {MPC} for
  large freeway networks using alternating optimization,'' \emph{IEEE
  Transactions on Intelligent Transportation Systems}, vol.~23, no.~3, pp.
  1875--1884, 2022.

\bibitem{shen2020distributed}
C.~Shen and Y.~Shi, ``Distributed implementation of nonlinear model predictive
  control for {AUV} trajectory tracking,'' \emph{Automatica}, vol. 115, p.
  108863, 2020.

\bibitem{7546918}
Y.~Zheng, S.~E. Li, K.~Li, F.~Borrelli, and J.~K. Hedrick, ``Distributed model
  predictive control for heterogeneous vehicle platoons under unidirectional
  topologies,'' \emph{IEEE Transactions on Control Systems Technology},
  vol.~25, \href{http://dx.doi.org/10.1109/TCST.2016.2594588}{DOI
  10.1109/TCST.2016.2594588}, no.~3, pp. 899--910, 2017.

\bibitem{navsalkar2023data}
A.~Navsalkar and A.~R. Hota, ``Data-driven risk-sensitive model predictive
  control for safe navigation in multi-robot systems,'' in \emph{2023 IEEE
  International Conference on Robotics and Automation (ICRA)}, pp.
  1442--1448.\hskip 1em plus 0.5em minus 0.4em\relax IEEE, 2023.

\bibitem{xinglongzhang2022_robust}
X.~Zhang, J.~Liu, X.~Xu, S.~Yu, and H.~Chen, ``Robust learning-based predictive
  control for discrete-time nonlinear systems with unknown dynamics and state
  constraints,'' \emph{IEEE Transactions on Systems, Man, and Cybernetics:
  Systems}, vol.~52, \href{http://dx.doi.org/10.1109/TSMC.2022.3146284}{DOI
  10.1109/TSMC.2022.3146284}, no.~12, pp. 7314--7327, 2022.

\bibitem{Ferranti}
L.~Ferranti, L.~Lyons, R.~R. Negenborn, T.~Keviczky, and J.~Alonso-Mora,
  ``Distributed nonlinear trajectory optimization for multi-robot motion
  planning,'' \emph{IEEE Transactions on Control Systems Technology}, vol.~31,
  \href{http://dx.doi.org/10.1109/TCST.2022.3211130}{DOI
  10.1109/TCST.2022.3211130}, no.~2, pp. 809--824, 2023.

\bibitem{Wei2021}
H.~Wei, C.~Shen, and Y.~Shi, ``Distributed {L}yapunov-based model predictive
  formation tracking control for autonomous underwater vehicles subject to
  disturbances,'' \emph{IEEE Transactions on Systems, Man, and Cybernetics:
  Systems}, vol.~51, \href{http://dx.doi.org/10.1109/TSMC.2019.2946127}{DOI
  10.1109/TSMC.2019.2946127}, no.~8, pp. 5198--5208, 2021.

\bibitem{el2015distributed}
S.~El-Ferik, B.~A. Siddiqui, and F.~L. Lewis, ``Distributed nonlinear {MPC} of
  multi-agent systems with data compression and random delays,'' \emph{IEEE
  Transactions on Automatic Control}, vol.~61, no.~3, pp. 817--822, 2015.

\bibitem{Dieter2022}
D.~Büchler, S.~Guist, R.~Calandra, V.~Berenz, B.~Schölkopf, and J.~Peters,
  ``Learning to play table tennis from scratch using muscular robots,''
  \emph{IEEE Transactions on Robotics}, vol.~38,
  \href{http://dx.doi.org/10.1109/TRO.2022.3176207}{DOI
  10.1109/TRO.2022.3176207}, no.~6, pp. 3850--3860, 2022.

\bibitem{wei2020continuous}
Q.~Wei, H.~Li, X.~Yang, and H.~He, ``Continuous-time distributed policy
  iteration for multicontroller nonlinear systems,'' \emph{IEEE Transactions on
  Cybernetics}, vol.~51, no.~5, pp. 2372--2383, 2020.

\bibitem{odekunle2020reinforcement}
A.~Odekunle, W.~Gao, M.~Davari, and Z.-P. Jiang, ``Reinforcement learning and
  non-zero-sum game output regulation for multi-player linear uncertain
  systems,'' \emph{Automatica}, vol. 112, p. 108672, 2020.

\bibitem{Han2022RL}
R.~Han, S.~Chen, S.~Wang, Z.~Zhang, R.~Gao, Q.~Hao, and J.~Pan, ``Reinforcement
  learned distributed multi-robot navigation with reciprocal velocity obstacle
  shaped rewards,'' \emph{IEEE Robotics and Automation Letters}, vol.~7,
  \href{http://dx.doi.org/10.1109/LRA.2022.3161699}{DOI
  10.1109/LRA.2022.3161699}, no.~3, pp. 5896--5903, 2022.

\bibitem{Sartoretti2019}
G.~Sartoretti, W.~Paivine, Y.~Shi, Y.~Wu, and H.~Choset, ``Distributed learning
  of decentralized control policies for articulated mobile robots,'' \emph{IEEE
  Transactions on Robotics}, vol.~35,
  \href{http://dx.doi.org/10.1109/TRO.2019.2922493}{DOI
  10.1109/TRO.2019.2922493}, no.~5, pp. 1109--1122, 2019.

\bibitem{wang2018model}
W.~Wang, X.~Chen, H.~Fu, and M.~Wu, ``Model-free distributed consensus control
  based on actor--critic framework for discrete-time nonlinear multiagent
  systems,'' \emph{IEEE Transactions on Systems, Man, and Cybernetics:
  Systems}, vol.~50, no.~11, pp. 4123--4134, 2018.

\bibitem{Yang2022databased}
X.~Yang, H.~Zhang, and Z.~Wang, ``Data-based optimal consensus control for
  multiagent systems with policy gradient reinforcement learning,'' \emph{IEEE
  Transactions on Neural Networks and Learning Systems}, vol.~33,
  \href{http://dx.doi.org/10.1109/TNNLS.2021.3054685}{DOI
  10.1109/TNNLS.2021.3054685}, no.~8, pp. 3872--3883, 2022.

\bibitem{emam2021safe}
Y.~Emam, G.~Notomista, P.~Glotfelter, Z.~Kira, and M.~Egerstedt, ``Safe
  reinforcement learning using robust control barrier functions,'' \emph{IEEE
  Robotics and Automation Letters},
  \href{http://dx.doi.org/10.1109/LRA.2022.3216996}{DOI
  10.1109/LRA.2022.3216996}, pp. 1--8, 2022.

\bibitem{cheng2019end}
R.~Cheng, G.~Orosz, R.~M. Murray, and J.~W. Burdick, ``End-to-end safe
  reinforcement learning through barrier functions for safety-critical
  continuous control tasks,'' in \emph{Proceedings of the AAAI Conference on
  Artificial Intelligence}, vol.~33, no.~01, pp. 3387--3395, 2019.

\bibitem{annurev-control-042920-020211}
L.~Brunke, M.~Greeff, A.~W. Hall, Z.~Yuan, S.~Zhou, J.~Panerati, and A.~P.
  Schoellig, ``Safe learning in robotics: From learning-based control to safe
  reinforcement learning,'' \emph{Annual Review of Control, Robotics, and
  Autonomous Systems}, vol.~5,
  \href{http://dx.doi.org/10.1146/annurev-control-042920-020211}{DOI
  10.1146/annurev-control-042920-020211}, no.~1, pp. 411--444, 2022.

\bibitem{Motoya8746143}
M.~Ohnishi, L.~Wang, G.~Notomista, and M.~Egerstedt, ``Barrier-certified
  adaptive reinforcement learning with applications to brushbot navigation,''
  \emph{IEEE Transactions on Robotics}, vol.~35,
  \href{http://dx.doi.org/10.1109/TRO.2019.2920206}{DOI
  10.1109/TRO.2019.2920206}, no.~5, pp. 1186--1205, 2019.

\bibitem{lu2021distributed}
Y.~Lu, Y.~Guo, G.~Zhao, and M.~Zhu, ``Distributed safe reinforcement learning
  for multi-robot motion planning,'' in \emph{2021 29th Mediterranean
  Conference on Control and Automation (MED)}, pp. 1209--1214.\hskip 1em plus
  0.5em minus 0.4em\relax IEEE, 2021.

\bibitem{Zhang9812083}
Z.~Zhang, X.~Wang, Q.~Zhang, and T.~Hu, ``Multi-robot cooperative pursuit via
  potential field-enhanced reinforcement learning,'' in \emph{2022
  International Conference on Robotics and Automation (ICRA)},
  \href{http://dx.doi.org/10.1109/ICRA46639.2022.9812083}{DOI
  10.1109/ICRA46639.2022.9812083}, pp. 8808--8814, 2022.

\bibitem{Paternain9718160}
S.~Paternain, M.~Calvo-Fullana, L.~F.~O. Chamon, and A.~Ribeiro, ``Safe
  policies for reinforcement learning via primal-dual methods,'' \emph{IEEE
  Transactions on Automatic Control}, vol.~68,
  \href{http://dx.doi.org/10.1109/TAC.2022.3152724}{DOI
  10.1109/TAC.2022.3152724}, no.~3, pp. 1321--1336, 2023.

\bibitem{boyd2004convex}
S.~Boyd and L.~Vandenberghe, \emph{Convex optimization}.\hskip 1em plus 0.5em
  minus 0.4em\relax Cambridge university press, 2004.

\bibitem{9811604}
X.~Zhang, Y.~Peng, W.~Pan, X.~Xu, and H.~Xie, ``Barrier function-based safe
  reinforcement learning for formation control of mobile robots,'' in
  \emph{2022 International Conference on Robotics and Automation (ICRA)},
  \href{http://dx.doi.org/10.1109/ICRA46639.2022.9811604}{DOI
  10.1109/ICRA46639.2022.9811604}, pp. 5532--5538, 2022.

\bibitem{BEMPORAD20023}
A.~Bemporad, M.~Morari, V.~Dua, and E.~N. Pistikopoulos, ``The explicit linear
  quadratic regulator for constrained systems,'' \emph{Automatica}, vol.~38,
  no.~1, pp. 3--20, 2002.

\bibitem{alonso2020explicit}
C.~A. Alonso, N.~Matni, and J.~Anderson, ``Explicit distributed and localized
  model predictive control via system level synthesis,'' in \emph{2020 59th
  IEEE Conference on Decision and Control (CDC)}, pp. 5606--5613.\hskip 1em
  plus 0.5em minus 0.4em\relax IEEE, 2020.

\bibitem{Song2022}
Y.~Song and D.~Scaramuzza, ``Policy search for model predictive control with
  application to agile drone flight,'' \emph{IEEE Transactions on Robotics},
  vol.~38, \href{http://dx.doi.org/10.1109/TRO.2022.3141602}{DOI
  10.1109/TRO.2022.3141602}, no.~4, pp. 2114--2130, 2022.

\bibitem{Li2021learning}
T.~Li, B.~Sun, Y.~Chen, Z.~Ye, S.~H. Low, and A.~Wierman, ``Learning-based
  predictive control via real-time aggregate flexibility,'' \emph{IEEE
  Transactions on Smart Grid}, vol.~12,
  \href{http://dx.doi.org/10.1109/TSG.2021.3094719}{DOI
  10.1109/TSG.2021.3094719}, no.~6, pp. 4897--4913, 2021.

\bibitem{xu2018learning}
X.~Xu, H.~Chen, C.~Lian, and D.~Li, ``Learning-based predictive control for
  discrete-time nonlinear systems with stochastic disturbances,'' \emph{IEEE
  Transactions on Neural Networks and Learning Systems}, no.~99, pp. 1--12,
  2018.

\bibitem{jiang2020cooperative}
Y.~Jiang, J.~Fan, W.~Gao, T.~Chai, and F.~L. Lewis, ``Cooperative adaptive
  optimal output regulation of nonlinear discrete-time multi-agent systems,''
  \emph{Automatica}, vol. 121, p. 109149, 2020.

\bibitem{Zhang2021}
Z.~Zhang, Y.-S. Ong, D.~Wang, and B.~Xue, ``A collaborative multiagent
  reinforcement learning method based on policy gradient potential,''
  \emph{IEEE Transactions on Cybernetics}, vol.~51,
  \href{http://dx.doi.org/10.1109/TCYB.2019.2932203}{DOI
  10.1109/TCYB.2019.2932203}, no.~2, pp. 1015--1027, 2021.

\bibitem{bhattacharya2023multiagent}
S.~Bhattacharya, S.~Kailas, S.~Badyal, S.~Gil, and D.~Bertsekas, ``Multiagent
  reinforcement learning: Rollout and policy iteration for pomdp with
  application to multi-robot problems,'' \emph{IEEE Transactions on Robotics},
  2023.

\bibitem{fan2020distributed}
T.~Fan, P.~Long, W.~Liu, and J.~Pan, ``Distributed multi-robot collision
  avoidance via deep reinforcement learning for navigation in complex
  scenarios,'' \emph{The International Journal of Robotics Research}, vol.~39,
  no.~7, pp. 856--892, 2020.

\bibitem{conte2016distributed}
C.~Conte, C.~N. Jones, M.~Morari, and M.~N. Zeilinger, ``Distributed synthesis
  and stability of cooperative distributed model predictive control for linear
  systems,'' \emph{Automatica}, vol.~69, pp. 117--125, 2016.

\bibitem{yang2019survey}
T.~Yang, X.~Yi, J.~Wu, Y.~Yuan, D.~Wu, Z.~Meng, Y.~Hong, H.~Wang, Z.~Lin, and
  K.~H. Johansson, ``A survey of distributed optimization,'' \emph{Annual
  Reviews in Control}, vol.~47, pp. 278--305, 2019.

\bibitem{rawlings2009model}
J.~B. Rawlings and D.~Q. Mayne, \emph{Model predictive control: Theory and
  design}.\hskip 1em plus 0.5em minus 0.4em\relax Nob Hill Pub., 2009.

\bibitem{venayagamoorthy2002comparison}
G.~K. Venayagamoorthy, R.~G. Harley, and D.~C. Wunsch, ``Comparison of
  heuristic dynamic programming and dual heuristic programming adaptive critics
  for neurocontrol of a turbogenerator,'' \emph{IEEE Transactions on Neural
  Networks}, vol.~13, no.~3, pp. 764--773, 2002.

\bibitem{conte2012computational}
C.~Conte, T.~Summers, M.~N. Zeilinger, M.~Morari, and C.~N. Jones,
  ``Computational aspects of distributed optimization in model predictive
  control,'' in \emph{2012 IEEE 51st IEEE Conference on Decision and Control
  (CDC)}, pp. 6819--6824.\hskip 1em plus 0.5em minus 0.4em\relax IEEE, 2012.

\bibitem{wang2009fast}
Y.~Wang and S.~Boyd, ``Fast model predictive control using online
  optimization,'' \emph{IEEE Transactions on Control Systems Technology},
  vol.~18, no.~2, pp. 267--278, 2009.

\bibitem{diehl2009efficient}
M.~Diehl, H.~J. Ferreau, and N.~Haverbeke, ``Efficient numerical methods for
  nonlinear mpc and moving horizon estimation,'' \emph{Nonlinear model
  predictive control: towards new challenging applications}, pp. 391--417,
  2009.

\bibitem{wills2004barrier}
A.~G. Wills and W.~P. Heath, ``Barrier function based model predictive
  control,'' \emph{Automatica}, vol.~40, no.~8, pp. 1415--1422, 2004.

\bibitem{engelmann2020aladin}
A.~Engelmann, Y.~Jiang, H.~Benner, R.~Ou, B.~Houska, and T.~Faulwasser,
  ``Aladin-—an open-source matlab toolbox for distributed non-convex
  optimization,'' \emph{Optimal Control Applications and Methods}, vol.~43,
  no.~1, pp. 4--22, 2022.

\bibitem{andersson2019casadi}
J.~A. Andersson, J.~Gillis, G.~Horn, J.~B. Rawlings, and M.~Diehl, ``{CasADi}:
  a software framework for nonlinear optimization and optimal control,''
  \emph{Mathematical Programming Computation}, vol.~11, no.~1, pp. 1--36, 2019.

\bibitem{xiao2022implementation}
K.~Xiao, L.~Ma, S.~Tan, Y.~Cong, and X.~Wang, ``Implementation of {UAV}
  coordination based on a hierarchical multi-{UAV} simulation platform,'' in
  \emph{Advances in Guidance, Navigation and Control: Proceedings of 2020
  International Conference on Guidance, Navigation and Control, ICGNC 2020,
  Tianjin, China, October 23--25, 2020}, pp. 5131--5143.\hskip 1em plus 0.5em
  minus 0.4em\relax Springer, 2022.

\bibitem{zhang2022robust11}
X.~Zhang, W.~Pan, R.~Scattolini, S.~Yu, and X.~Xu, ``Robust tube-based model
  predictive control with koopman operators,'' \emph{Automatica}, vol. 137, p.
  110114, 2022.

\bibitem{seuret2015complete}
A.~Seuret, F.~Gouaisbaut, and Y.~Ariba, ``Complete quadratic lyapunov
  functionals for distributed delay systems,'' \emph{Automatica}, vol.~62, pp.
  168--176, 2015.

\bibitem{gronauer2022multi}
S.~Gronauer and K.~Diepold, ``Multi-agent deep reinforcement learning: a
  survey,'' \emph{Artificial Intelligence Review}, vol.~55, no.~2, pp.
  895--943, 2022.

\bibitem{zhang2021model}
X.~Zhang, Y.~Peng, B.~Luo, W.~Pan, X.~Xu, and H.~Xie, ``Model-based safe
  reinforcement learning with time-varying constraints: Applications to
  intelligent vehicles,'' \emph{IEEE Transactions on Industrial Electronics},
  \href{http://dx.doi.org/10.1109/TIE.2023.3317853}{DOI
  10.1109/TIE.2023.3317853}, pp. 1--10, 2024.

\bibitem{magni2001stabilizing}
L.~Magni, G.~De~Nicolao, L.~Magnani, and R.~Scattolini, ``A stabilizing
  model-based predictive control algorithm for nonlinear systems,''
  \emph{Automatica}, vol.~37, no.~9, pp. 1351--1362, 2001.

\bibitem{marruedo2002input}
D.~L. Marruedo, T.~Alamo, and E.~Camacho, ``Input-to-state stable {MPC} for
  constrained discrete-time nonlinear systems with bounded additive
  uncertainties,'' in \emph{Proceedings of the 41st IEEE Conference on Decision
  and Control, 2002.}, vol.~4, pp. 4619--4624.\hskip 1em plus 0.5em minus
  0.4em\relax IEEE, 2002.

\end{thebibliography}
				\appendices
				\section{}\label{sec:32}
				\subsection{Model Derivation of MRS}\label{sec:model}
				By discretizing~\eqref{Eqn:kinematic model} under~\eqref{Eqn:formation error model}, we write the discrete-time local formation error model for the $i$-th robot as follows:
				\begin{equation}\label{Eqn:discrete-time formation tracking error model}
					\left\{ \begin{array}{l}
						e_{x,i}( k+1 ) =e_{x,i}( k ) +\varDelta t\big(\omega _i( k ) e_{y,i}( k ) -(d_i+s_i ) v_i( k)\big.\\
						\hspace{5mm}\big. +\sum_{j\in\mathcal{N}_i,j\neq i}{c_{ij}v_j( k ) \cos \theta _{ji}( k )}+s_iv_r( k ) \cos \theta _{ri}( k )\big)\\
						e_{y,i}(k+1) =e_{y,i}(k)+\varDelta t\big(-\omega _i(k) e_{x,i}(k) +\big.\\
						\hspace{10mm}\big.\sum_{j\in\mathcal{N}_i,j\neq i}c_{ij}v_j(k) \sin \theta _{ji}(k) +s_iv_r(k) \sin \theta _{ri}(k) \big)\\
						e_{\theta,i}\left( k+1 \right) =e_{\theta,i}\left( k \right) +\varDelta t\left( \omega _r\left( k \right) -\omega _i\left( k \right) \right) \,\,\\
						e_{v,i}\left( k+1 \right) =e_{v,i}\left( k \right) +\varDelta t\left( a_r\left( k \right) -a_i\left( k \right) \right),\\
					\end{array} \right.  
				\end{equation}
				where $e_{x,i},\,e_{y,i},\,e_{\theta,i},\,e_{v,i}$ are the corresponding entries of $e_i$,  $v_r$ and $\omega_r$ are the leader's linear velocity and angular velocity respectively, $\varDelta t$ is the adopted sampling interval, the parameter $d_i=\sum_{j\in\mathcal{N}_i,j\neq i}{c_{ij}}$,  $\theta _{ji} =\theta _j -\theta _i$, $\theta _{ri} =\theta _r -\theta _i$.
				Hence, one can straightforwardly rewrite~\eqref{Eqn:discrete-time formation tracking error model} in a concise input-affine form like~\eqref{Eqn:LL-nei}. 
				
				\begin{remark}
					In model~\eqref{Eqn:LL-nei}, the local control input directly affects the behavior of individual robots, while interactions among neighboring robots are conveyed through their respective states. This structural characteristic allows for the design of our distributed policy learning algorithms with synchronous updating mechanisms, mitigating the non-stationary issue commonly encountered in multiagent RL~\cite{gronauer2022multi}. Deriving such an input-affine local model from general MRS is not overly restrictive. One approach is to introduce an auxiliary control input with an integral action, as demonstrated in \eqref{Eqn:kinematic model} with $\dot v=a$.  
				\end{remark}
				
				\subsection{Theoretical Results of Policy Learning for DMPC}\label{sec:theoretical}
				In the following, we first prove the convergence and closed-loop stability under the distributed policy learning procedure~\eqref{Eqn:safempc-o}. Then, practical conditions for convergence and stability under the distributed actor-critic implementation, i.e., Algorithm~\ref{alg:d-lpc-AC-o}, are established. Finally, the closed-loop robustness of online deployment is proven under bounded disturbances.
				
				\textbf{\emph{1) Convergence and stability guarantees under procedure~\eqref{Eqn:safempc-o}}:}
				In what follows, we show that the control policy and the value function eventually converge to the optimal values respectively, i.e.,  $u^t(e(\tau))={\rm col}_{i\in\mathbb{N}_1^M}u_i^t(e_{\scriptscriptstyle \mathcal{N}_i}(\tau))\rightarrow u^{\ast}(e(\tau))$ and  $ J^t(\tau)=\sum_{i=1}^M J_i^t(\tau)\rightarrow  J^{\ast}(\tau)$ as $t\rightarrow+\infty$. %, $\forall i\in\mathbb{N}_1^N$.   
				\begin{theorem}[Convergence]\label{theo:safety_convergence-o}
					Let $\bm u^0(k)$ be an initial policy and the initial value function $ J^0(e(\tau))\geq r(e(\tau),u^0(\tau))+ J^{0}(e(\tau+1))$, $\tau\in[k,k+N-1]$; then under iteration~\eqref{Eqn:safempc-o}, it holds that
					\begin{enumerate}[(i)]
						\item $ J^{t+1}(e(\tau))\leq  J^{t}(e(\tau))$;
						\item $ J^t(e(\tau))\rightarrow  J^{\ast}(e(\tau))$ and $u^t(\tau)\rightarrow u^{\ast}(\tau)$ for all $\tau\in[k,k+N]$, as $t\rightarrow +\infty$. \hfill $\blacksquare$
					\end{enumerate}
				\end{theorem}
				%\textbf{Proof}. %Please refer to Appendix~\ref{app-a}.
				%\hfill $\square$
				%\subsection{Proof of Theorem~\ref{theo:safety_convergence}}\label{app-a}
				\textbf{Proof}. 1): First, collecting the iterative step~\eqref{Eqn:value-up-dhp-o} for all $i\in\mathbb{N}_1^M$ results in the following centralized form 
				\begin{subequations}\label{Eqn:hdp-o}
					\begin{align}\label{Eqn:value-up-hdp-o}
						J^{t+1}(e(\tau))
						=& r(\tau)+ J^{t}(e(\tau+1)).
					\end{align}
					Moreover, since $u_i$ is only related to $e_{\scriptscriptstyle \mathcal{N}_i}$,~\eqref{Eqn:policy-im-DHP-o} is equivalent to
					\begin{equation}%\label{Eqn:policy-hdp}
						u_i^{t+1}(e_{\scriptscriptstyle \mathcal{N}_i}(\tau))=
						\argmin{u_i(e_{\scriptscriptstyle \mathcal{N}_i}(\tau))}\{ r_i(\tau)+  J^{t+1}\big(e(\tau+1)\big)\},
						%	u^{t+1}(\tau)=&\arg\min_{u}\{\bar r(\tau)+ \bar J^{t+1}\big(e(\tau+1)\big)\}.
					\end{equation} 
					and equivalent to the centralized form of policy update under $u={\rm col}_{i\in\mathbb{N}_1^M}(u_i(e_{\scriptscriptstyle\mathcal{N}_i}))$, i.e.,  
					\begin{align}\label{Eqn:policy-hdp-centra-o}
						u^{t+1}(e(\tau))=
						\argmin{u_i(e_{\scriptscriptstyle \mathcal{N}_i}(\tau)),\,i\in\mathbb{N}_1^M}\{ r(\tau)+  J^{t+1}\big(e(\tau+1)\big)\}.
					\end{align} 
				\end{subequations}
				Then, one can apply the proof arguments in~\cite{zhang2021model} to the centralized system, which proves that $ J^{t+1}(e(\tau))\leq  J^t(e(\tau))$, for all $\tau\in[k,k+N-1]$.
				Moreover, according to~\cite{zhang2021model}, the second point can be proven. 
				\hfill $\square$
				
				To state the following theorem in a compact form, let $\bar \phi_i(e_{\scriptscriptstyle \mathcal{N}_i})=\phi_i(e_{\scriptscriptstyle \mathcal{N}_i},K_{\scriptscriptstyle \mathcal{N}_i}e_{\scriptscriptstyle \mathcal{N}_i})$ and $L_{\phi,i}=\sup\|\bar \phi_i(e_{\scriptscriptstyle\mathcal{N}_i})\|/\|e_{\scriptscriptstyle\mathcal{N}_i}\|\rightarrow 0$ in a small neighbor of the origin. Let $\mathcal{E}_{f,i}$ (i.e., $S_i$)  be selected as a subset  of the control invariant set of~\eqref{Eqn:LL-nei} under~\eqref{Eqn:Lya-mod-o} in the neighbor of the origin, and let $\beta_i$ and $P_i$ be such that for all $e_{\scriptscriptstyle\mathcal{N}_i}\in\mathcal{E}_{f,i}$ (see~\cite{magni2001stabilizing})
				\begin{equation}\label{Eqn:q_in-o}
					\|P_i\|L_{\phi,i}^2+2\|P_iF_i\|L_{\phi,i}<(\beta_i-1)\lambda_{\rm min}(\bar Q_i),
				\end{equation} where $\bar Q_i=Q_{i}+K_{\scriptscriptstyle\mathcal{N}_i}^{\top} R_{i} K_{\scriptscriptstyle\mathcal{N}_i}$. In a collective form, define the terminal constraint in centralized form as $\mathcal{E}_{f}=\mathcal{E}_{f,1}\times\cdots\times\mathcal{E}_{f,M}$. 
				\begin{theorem} [Closed-loop stability]\label{theo:stability-o}
					Suppose the prediction horizon $N$ has been selected such that the optimal control $\bm u_i^{\ast}(0)\in\mathcal{U}_i^N$ at time $k=0$, $\forall i\in\mathbb{N}_1^M$ satisfies $e_i(N)\in\mathcal{E}_{f,i}$. Under Assumption~\ref{assum:stabilizing_control}, if, for any $e\in\mathcal{E}_f$,  the next local state evolution,
					denoted as $e^+_{i}$, under control $u_i(e_{\scriptscriptstyle \mathcal{N}_i})$ is such that $e_i^+\in\mathcal{E}_{f,i}$,  $\forall i\in\mathbb{N}_1^M,$ 
					%		\begin{equation}\label{Eqn:condition-o}
						%		(e^+_{i})^{\top}P_ie_i^+\leq e_{\scriptscriptstyle \mathcal{N}_i}^{\top}F_i^{\top}P_iF_ie_{\scriptscriptstyle \mathcal{N}_i}+(\beta_i-1)\lambda_{\rm min}(\bar Q_i)\|e_{\scriptscriptstyle\mathcal{N}_i}\|^2,
						%		\end{equation} $\forall i\in\mathbb{N}_1^M,$ 
					then the global state and control, i.e., $e$ and $u$, converge to the origin asymptotically. \hfill $\blacksquare$
				\end{theorem}
				%\textbf{Proof}. Please refer to Appendix~\ref{app-b}.
				%\hfill$\square$
				%\subsection{Proof of Theorem~\ref{theo:stability}}\label{app-b}
				\textbf{Proof}. First note that, at the initial time instant $0$, $\bm u_i^{\ast}(0)$, $\forall i\in\mathbb{N}_1^M$ are optimal policies. Let at the subsequent time $k=1$, $\bm u_i^{f}(1)=u_i^{\ast}(e_{\scriptscriptstyle \mathcal{N}_i}(1)),\cdots,u_i^{\ast}(e_{\scriptscriptstyle \mathcal{N}_i}(N-1)),u_i(e_{\scriptscriptstyle \mathcal{N}_i}(N))$ such that $e_i(N+1)\in\mathcal{E}_{f,i}$. Denoting $ J^{f}(e(1))$ as the cost associated with $\bm u_i^{f}(1)$, $\forall i\in\mathbb{N}_1^M$,  one has \begin{equation*}%\label{Eqn:V-OPT-o}
					\begin{array}{ll}
						J^f(e(1))-  J^{\ast}(e(0))=-D(e(0),u^{\ast}(0))+ \chi(e(N)),
					\end{array}
				\end{equation*}
				where $D(e(0),u(0))=\sum_{i=1}^M(\|e_{\scriptscriptstyle\mathcal{N}_i}(0)\|_{Q_i}^2+\|u_i^{\ast}(0)\|_{R_i}^2$, $\chi(e(N))=\sum_{i=1}^M\|e_{\scriptscriptstyle\mathcal{N}_i}(N)\|_{Q_i}^2+ \|u_i(N)\|_{R_i}^2+\|e_i(N+1)\|_{P_i}^2-\|e_i(N)\|_{P_i}^2$.
				%Specifically, letting $u_i^{\ast}(e_{\scriptscriptstyle \mathcal{N}_i}(N))$ be  $K_{\scriptscriptstyle \mathcal{N}_i}e_{\scriptscriptstyle \mathcal{N}_i}(N)$, which of course satisfies~ in $\bm u_i^{\ast}(1)$ results in feasible control policies $\bm u_i^{f}(1)$, $i\in\mathbb{N}_1^M$. 
				Then, given the definition of $\phi_i$, one has 
				\begin{equation}\label{Eqn:Vf-0}
					\begin{array}{lll}
						\|e_i(N+1)\|_{P_i}^2&=&\|e_{\scriptscriptstyle\mathcal{N}_i}(N)\|^2_{F_i^{\top} P_iF_i}+\|\bar\phi_i(e_{\scriptscriptstyle\mathcal{N}_i}(N)\|^2_{P_i}+\\
						&&2\bar\phi_i(e_{\scriptscriptstyle\mathcal{N}_i}(N))^{\top}P_ie_{\scriptscriptstyle\mathcal{N}_i}(N)\\
						&\leq&\|e_{\scriptscriptstyle\mathcal{N}_i}(N)\|^2_{F_i^{\top} P_iF_i}+(\|P_i\|L_{\phi,i}^2+\\
						&&2\|P_iF_i\|L_{\phi,i})\|e_{\scriptscriptstyle\mathcal{N}_i}(N)\|^2\\
						&\leq& \|e_{\scriptscriptstyle \mathcal{N}_i}(N)\|^2_{F_i^{\top}P_iF_i+(\beta_i-1)\bar Q_i}
					\end{array}
				\end{equation}
				where the last inequality is due to~\eqref{Eqn:q_in-o}. 
				Hence, in view of~\eqref{Eqn:Lya-mod-o},~\eqref{Eqn:q_in-o}, we have $\chi(e(N))\leq 0$.
				In view of~\eqref{Eqn:Lya-mod-o}, one has
				\begin{equation*}%\label{Eqn:V-MONO-1}
					\begin{array}{ll}
						J^{f}(e(1))-  J^{\ast}(e(0))\leq-D(e(0),u^{\ast}(0)), 
					\end{array}
				\end{equation*}
				which by induction leads to $ J^{f}(k+1)-  J^{f}(k)\rightarrow0$ as $k\rightarrow+\infty$. Hence, $e$ and $u$ converge to the origin asymptotically. 
				\hfill$\square$
				
				\textbf{\emph{2) Convergence and stability under Algorithm~\ref{alg:d-lpc-AC-o}}:} We first prove the convergence of Algoritihm~\ref{alg:d-lpc-AC-o}.
				To this end, we write the local optimal costate and control policy  for all $\tau\in[k,k+N-1]$ and $i\in\mathbb{N}_1^M$ as
				$${\lambda}_i^{\ast}(e_{\scriptscriptstyle\mathcal{N}_i}(\tau))=({W_{c,i}^{\ast})}^{\top}h_{c,i}(e_{\scriptscriptstyle\mathcal{N}_i}(\tau),\tau)+\kappa_{c,i}(\tau)$$
				$$u_i^{\ast}(e_{\scriptscriptstyle\mathcal{N}_i}(\tau))=({W_{a,i}^{\ast})}^{\top}h_{a,i}(e_{\scriptscriptstyle\mathcal{N}_i}(\tau),\tau)+\kappa_{a,i}(\tau),$$ where $W_{c,i}^{\ast}$ and $W_{a,i}^{\ast}$ are the optimal weights of $W_{c,i}$ and $W_{a,i}$, $\kappa_{c,i}$ and $\kappa_{a,i}$ are the associated reconstruction errors. Given the universal capability of one-hidden-layer-based neural networks, the following standard assumption on the actor and critic network is introduced.
				\begin{assumption}[Weights and reconstruction errors]\label{assum:network-o} For all $i\in\mathbb{N}_1^M$,  it holds that
					\begin{enumerate}[(i)]
						\item $\|W_{c,i}^{\ast}\|\leq {W}_{c,i}^{\scriptscriptstyle[m]}$, $\|\sigma_{c,i}\|\leq {\sigma}_{c,i}^{\scriptscriptstyle[m]}$,  $\|\kappa_{c,i}\|\leq  {\kappa}_{c,i}^{\scriptscriptstyle[m]}$;
						\item $\|W_{a,i}^{\ast}\|\leq  W_{a,i}^{\scriptscriptstyle[m]}$, $\|\sigma_{a,i}\|\leq {\sigma}_{a,i}^{\scriptscriptstyle[m]}$,   $\|{\kappa}_{a,i}\|\leq  {\kappa}_{a,i}^{\scriptscriptstyle[m]}$. 
					\end{enumerate} 
				\end{assumption}
				\begin{assumption}[Model]\label{ASSUM: INPUT MAPPING-o}
					There exist finite scalars $\bar f_{\bar{\mathcal{N}}_i}$ and $\bar g_{i}$ such that, for any $e_i\in\mathcal{E}_i$ and $i\in\mathbb{N}^M_1$,
					\begin{equation}\label{equ:assum:model}
						\|\vec f_{\bar{\mathcal{N}}_i}\|\leq \bar f_{\bar{\mathcal{N}}_i} \ \text{and}\	\|g_{i}(e_i)\|\leq \bar g_{i}, 
					\end{equation}
					where  $\vec f_{\bar{\mathcal{N}}_i}$ is the row collection of matrices $(\partial {f_{j}(e_{\scriptscriptstyle\mathcal{N}_j}(\tau))/\partial e_{\scriptscriptstyle\mathcal{N}_i}(\tau)})^{\top}$ for all $j\in \bar{\mathcal{N}}_i$. 
				\end{assumption}
				
				To state the following theorem in a compact form,  for a general variable, we use $q$ and $q^{+}$ to denote $q(k)$ and $q(k+1)$ respectively, unless otherwise specified.
				Let  $$\Gamma_{c,i}(z)= \|z\|^2-2\|z^{\top}z^+\|\| \vec f_{\bar{\mathcal{N}}_i}\|+\|z^+\|^2\| \vec f_{\bar{\mathcal{N}}_i}\|^2,$$ for $z={\sigma}_{c,i}$. Define $\tilde W_{\star,i}= W^{\ast}_{\star,i}- W_{\star,i}$, $\star=a,c$ in turns. 
				\begin{theorem}[Convergence of Algorithm~\ref{alg:d-lpc-AC-o}]\label{THEOM:convergence-o}
					Under Assumptions~\ref{assum:network-o}-\ref{ASSUM: INPUT MAPPING-o},  if  the learning rates are designed such that
					\begin{subequations}\label{learning rate-inequa-o}
						\begin{align}
							&\begin{array}{l}
								C_{c,i}:=\gamma_{c,i}\Gamma_{c,i}({\sigma}_{c,i})<1,
							\end{array}\\
							&\begin{array}{l}
								C_{a,i}:=2\lambda_{\rm max}(R_i)\cdot\gamma_{a,i}  \|{\sigma}_{a,i}\|^2<1, 
							\end{array}
						\end{align}
						where $\lambda_{\rm max}(\cdot)$ denotes the maximal eigenvalue;
					\end{subequations}
					then the terms $$
					\begin{array}{ll}
						\xi_{a,i}(\tau)=\tilde W_{a,i}^{\top}(\tau)h_{a,i}(\tau),\vspace{1mm}\\ \xi_{c,i}(\tau)=-\vec f_{\bar{\mathcal{N}}_i}\tilde{W}_{c,i}^{\top}(\tau) h_{c,i}^{+}(\tau)+\tilde{W}_{c,i}^{\top}(\tau) h_{c,i}(\tau),
					\end{array}$$ are uniformly ultimate bounded,  as the iteration step $t\rightarrow +\infty$ (see Algorithm~\ref{alg:d-lpc-AC-o}). Moreover, if $\kappa_{a,i},\,\kappa_{c,i}\rightarrow 0$,  
					\begin{equation*}
						\xi_{a,i}\rightarrow 0\ \text{and}\ \xi_{c,i}\rightarrow  0,
					\end{equation*}
					as $t\rightarrow +\infty$.  \hfill $\blacksquare$
				\end{theorem}
				%%	\textbf{Proof}. Please refer to Appendix~\ref{app-d}.
				%  \hfill $\square$
				
				\textbf{Proof}.
				Define a collective Lyapunov function as
				\begin{equation}\label{Eqn:v_{k}-o}
					\begin{array}{ll}
						V(\tau)=\sum_{i=1}^MV_{c,i}(\tau)+V_{a,i}(\tau),
					\end{array}
				\end{equation}
				where
				\begin{equation*}
					\begin{array}{ll}
						V_{c,i}=\text{tr}\left( 1/\gamma_{c,i}(\tilde{W}_{c,i})^{\top}\tilde W_{c,i}\right),\vspace{1mm}\\
						V_{a,i}=\text{tr}\left( 1/\gamma_{a,i}(\tilde{W}_{a,i})^{\top}\tilde W_{a,i}\right).
					\end{array}
				\end{equation*}
				In view of the update rule~\eqref{Eqn:wc-o}, letting $\Delta V_{c,i}(\tau)=V_{c,i}(\tau+1)-V_{c,i}(\tau)$, one  writes
				\begin{equation}\label{Eqn:deltaVc(k)-o}
					\Delta V_{c,i}=\text{tr}\left( 2(\tilde{W}_{c,i})^{\top}\frac{\partial \delta_{c,i}}{\partial W_{c,i}}+\gamma_{c,i}\|\frac{\partial \delta_{c,i}}{\partial W_{c,i}}\|_F^2\right).
				\end{equation}
				First note that 
				\begin{equation}\label{Eqn:delta(k)-o}
					\frac{\partial \delta_{c,i}}{\partial W_{c,i}}=-2\sigma_{c,i}\varepsilon_{c,i}^{\top}+2\sigma_{c,i}^{+}\varepsilon_{c,i}^{\top}\vec f_{\bar{\mathcal{N}}_i},
				\end{equation}
				where $\vec f_{\bar{\mathcal{N}}_i}$ is defined previously in~\eqref{equ:assum:model}.
				
				%
				%where $\Delta h_{c,k}=\gamma^Lh_{c,k+L}-h_{c,k}$.

				Moreover, in view of the definition of $\varepsilon_{c,i}$ and of Assumption~\ref{assum:network-o}, it follows that
				\begin{equation}\label{Eqn:epsilonc(k)-o}
					\begin{array}{ll}
						\varepsilon_{c,i}&=\lambda_i^d-\lambda_i^{\ast}+ \lambda_i^{\ast}-\hat\lambda_i\vspace{1mm}\\
						&=\xi_{c,i}+\Delta \kappa_{c,i},
					\end{array}
				\end{equation}
				where $\xi_{c,i}=-\vec f_{\bar{\mathcal{N}}_i}\tilde{W}_{c,i}^{\top} h_{c,i}^{+}+\tilde{W}_{c,i}^{\top} h_{c,i}$,
				$\Delta \kappa_{c,i}={\kappa}_{c,i}^{[m]}-\vec f_{\bar{\mathcal{N}}_i}({\kappa}_{c,i}^{+})^{[m]}$.
				
				Taking~\eqref{Eqn:delta(k)-o} with~\eqref{Eqn:epsilonc(k)-o} into~\eqref{Eqn:deltaVc(k)-o},  in view of Assumption~\ref{assum:network-o}, one promptly has
				\begin{equation}\label{Eqn:delta_Vc_final(k)-o}
					\begin{array}{lll}
						\hspace{-2mm}\Delta V_{c,i}
						&\hspace{-2mm}\leq&\hspace{-2mm}-4\xi_{c,i}^{\top}(\xi_{c,i}+\Delta \kappa_{c,i})+\\
						&&\hspace{1mm}4\gamma_{c,i}\Gamma_{c,i}({\sigma}_{c,i})\|\xi_{c,i}+\Delta \kappa_{c,i}\|^2\vspace{1mm}\\
						\hspace{-2mm}&\hspace{-2mm}\leq&\hspace{-2mm}-c_{c,i}\|\xi_{c,i}\|^2+\epsilon_{c,i},
					\end{array}
				\end{equation}
				where $c_{c,i}=4-4\gamma_{c,i}\Gamma_{c,i}({\sigma}_{c,i})-\beta_{c,i}$,  $\epsilon_{c,i}=1/\beta_{c,i}(1+(4\beta_{c,i}-4)\gamma_{c,i}\Gamma_{c,i}({\sigma}_{c,i})$, $\beta_{c,i}>0$ is a tuning constant. The last inequality in~\eqref{Eqn:delta_Vc_final(k)-o} is due to Young's inequality property. 
				
				To compute $\Delta V_{a,i}$, we first write 
				\begin{equation}\label{Eqn:delta_Va_final(k)-o}
					\Delta V_{a,i}=\text{tr}\left( 2(\tilde{W}_{a,i})^{\top}\frac{\partial \delta_{a,i}}{\partial W_{a,i}}+\gamma_{a,i}\|\frac{\partial \delta_{a,i}}{\partial W_{a,i}}\|_F^2\right).
				\end{equation}
				Inline with~\eqref{Eqn:delta(k)-o}, one has
				\begin{equation}\label{Eqn:delta_wA-o}
					\frac{\partial \delta_{a,i}}{\partial W_{a,i}}
					=-2\sigma_{a,i}\varepsilon_{a,i}^{\top}\bar R_i,
				\end{equation}
				where $\bar R_i=2R_i$.
				Taking~\eqref{Eqn:delta_wA-o} into~\eqref{Eqn:delta_Va_final(k)-o}, it holds that
				\begin{equation}\label{Eqn:delta_va-o}
					\Delta V_{a,i}=-c_{a,i}\|\varepsilon_{a,i}\|_{\bar R_i}^2,
				\end{equation}
				where $c_{a,i}=4-4\lambda_{\rm max}(\bar R_i)\cdot\gamma_{a,i} {\|\sigma}_{a,i}\|^2$.
				
				Combining~\eqref{Eqn:delta_Vc_final(k)-o} and~\eqref{Eqn:delta_va-o}, leads to
				\begin{equation}
					\Delta V=\sum_{i=1}^M-(c_{c,i}\|\xi_{c,i}\|^2+c_{a,i}\|\varepsilon_{a,i}\|_{\bar R_i}^2)+\epsilon_{c}.
				\end{equation}
				where $\epsilon_{c}=\sum_{i=1}^M\epsilon_{c,i}$.
				Hence, in view of~\eqref{learning rate-inequa-o} and setting $\beta_{c,i}$ small,  for any $i\in\mathbb{N}_1^M$, it follows that
				\begin{equation}
					\|\xi_{c,i}\|\leq \sqrt{\frac{\epsilon_{c}}{c_{c,i}}}\ \ \text{and}\ \ \|\varepsilon_{a,i}\|\leq \sqrt{\frac{\epsilon_{c}}{c_{a,i}}}, 
				\end{equation}
				as the iteration step $t\rightarrow +\infty$.
				Note that one has
				\begin{equation*}
					\begin{array}{lll}
						\varepsilon_{a,i}&\hspace{-2mm}=&\hspace{-2mm}u_{o,i}^d-u_{o,i}^{\ast}+u_{o,i}^{\ast}-u_{o,i}\\
						&\hspace{-2mm}=&\hspace{-2mm}u_{o,i}^d-u_{o,i}^{\ast}+\bar R_{i} ({\xi}_{a,i}+\kappa_{a,i}).\end{array}
				\end{equation*}
				Note that
				\begin{equation}\label{Eqn: u0d-u0-o}
					\begin{array}{ll}
						\|u_{o,i}^d-u_{o,i}^{\ast}\|&\hspace{-2mm}=\|-\sum_{j\in\bar{\mathcal{N}}_i}g_{i}^{\top}(e_i) (\tilde{\lambda}^{[i]}_j)^+\|,\vspace{2mm}\\
						&\hspace{-2mm}\leq\sum_{j\in\bar{\mathcal{N}}_i}\|g_{i}(e_i)\| (\sqrt{\frac{\epsilon_{c}}{c_{a,i}}}+{\kappa}_{c,i}^{\scriptscriptstyle [m]}):=Y_{u,i},
					\end{array}
					%\sum_{j\in \bar{\mathcal{N}}_i}g_i^{\top}\tilde \lambda^{[j]}_i(\tau+1)
				\end{equation} where $\tilde{\lambda}^{[i]}_j={\lambda^{\ast}_j}^{[i]}-\hat{\lambda}^{[i]}_j$. 
				
				In view of~\eqref{Eqn: u0d-u0-o}, for any $i\in\mathbb{N}_1^M$, one consequently has
				\begin{equation*}
					\begin{array}{ll}
						\|\xi_{a,i}\|&\hspace{-2mm}\leq \frac{1}{|\rho(\bar R_{i})|}(\|\bar R_{i}\kappa_{a,i}\|+\|u_{o,i}^d-u_{o,i}^{\ast}\|+\sqrt{\frac{\epsilon_{c}}{c_{a,i}}})\vspace{2mm}\\
						&\hspace{-2mm}\leq \frac{1}{|\rho(\bar R_{i})|}(\|\bar R_{i}\|{\kappa}_{a,i}^{\scriptscriptstyle [m]}+Y_{u,i}+\sqrt{\frac{\epsilon_{c}}{c_{a,i}}}),
					\end{array}
				\end{equation*}
				as the iteration step $t\rightarrow +\infty$, where $\rho(\bar R_{i})$ is the minimal (maximal) eigenvalue of $\bar R_{i}$ if it is positive-definite (negative-definite).
				
				Consequently, if $\kappa_{a,i},\, \kappa_{c,i} \rightarrow 0$, it promptly follows that
				\begin{equation*}
					\xi_{a,i}\rightarrow 0\ \text{and}\ \xi_{c,i}\rightarrow 0,
				\end{equation*}
				as $t\rightarrow +\infty$.
				\hfill $\square$
				\begin{figure*}[h]
					\centerline{\includegraphics[scale=0.48]{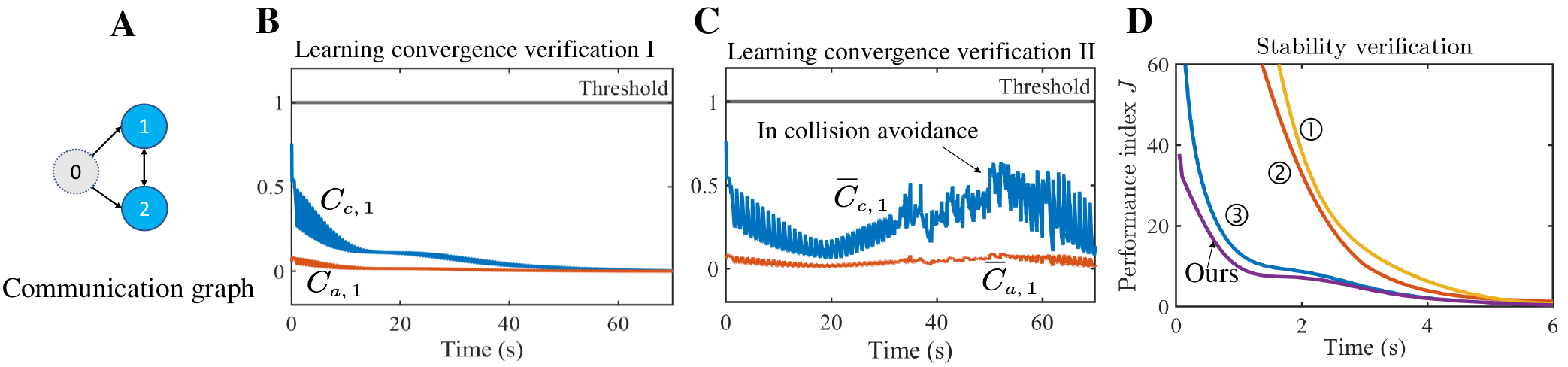}}
					\caption{ {\color{black}A: Communication graph of $M=2$ robots. B and C: Learning convergence condition of the 1st robot to verify Theorems~\ref{THEOM:convergence-o} and~\ref{THEOM:convergence} (see Appendix~\ref{appendix B}), respectively. The weights for the actor and critic were initialized with uniformly distributed random values in the range $[0,\,0.1]$.  D: Stability verification for Theorem~\ref{theo-stability-o}. \textcircled{1}, \textcircled{2}, and \textcircled{3} are the costs associated with three different baseline stabilizing policies. Various stabilizing policies can be used for stability verification.}}
					\label{fig:safety_verification} 
				\end{figure*}
				\begin{theorem}[Closed-loop stability under Algorithm~\ref{alg:d-lpc-AC-o}]\label{theo-stability-o}
					Under Assumptions~\ref{assum:stabilizing_control}-\ref{ASSUM: INPUT MAPPING-o}, if condition~\eqref{Eqn:sta-con-o} is fulfilled,  then the global state and control under Algorithm~\ref{alg:d-lpc-AC-o}, i.e., $e$ and $u$, converge to the origin asymptotically.   \hfill $\blacksquare$
				\end{theorem}
				%\textbf{Proof.} Please refer to Appendix~\ref{app-e}.   \hfill $\square$
				\textbf{Proof.}  %First, note that starting from a stabilizing control policy sequence $\bm u^{b}_i(0)$, one can promptly design the control policy $\bm u^{b}_i(k)$, $\forall\, k\in\mathbb{N}$ such that the control problem is recursively feasible. Consequently, inline with Theorem~\ref{theo:stability}, one has
				Note that
				%\begin{equation*}%\label{Eqn:stabilizing-converge-o}
				$J^b(k+1|k)- J^b(k|k)\leq-s(e^{b}(k),u^{b}(k)).$
				%\end{equation*}
				In view of Theorem~\ref{theo:safety_convergence-o}, it holds that $e^b$ and $u^b$ converge to the origin asymptotically.
				By condition~\eqref{Eqn:sta-con-o}, one has 
				\begin{equation}
					\begin{array}{ll}
						J(k+1|k)&\leq  J^b(k+1|k)\\
						&\leq  J^b(k|k)-s(e^{b}(k),u^{b}(k)).
					\end{array}
				\end{equation}
				As $e^b(k)\rightarrow 0$ and $u^b(k)\rightarrow 0$ as $k\rightarrow +\infty$, it follows that $ J(k|k)\rightarrow 0$ as $k\rightarrow+\infty$. Consequently,  $e$ and $u$ converge to the origin asymptotically.\hfill $\square$
				
				\textbf{\emph{3) Closed-loop robustness in perturbed scenario}:}  Consider that the overall model~\eqref{Eqn:C full model} is influenced by a norm-bounded additive disturbance, i.e., 
				\begin{equation}\label{eqn:perturbedsys}
					e_o(k+1)=F_c(e_o(k))+G_c(e_o(k))u(k)+w(k), 
				\end{equation}
				where $e_o(k)$ is the real state and the additive disturbance $w(k)$ satisfies $\|w(k)\|\leq \varepsilon_w$, $\varepsilon_w>0$.
				\begin{assumption}[Lipschitz continuous]\label{ASSM:LIPSCHITZ}
					There exists a finite Lipschitz constant $L_p$ such that for any states $z,y\in \mathcal{E}$ and $C^1$ control policies $u(y),u(z)\in \mathcal{U}$, one has
					\begin{equation}\label{Eqn:lipschitz}
						\|F_c(y)+G_c(y)u(y)-F_c(z)-G_c(z)u(z)\|\leq L_p\|y-z\|.
					\end{equation}
				\end{assumption}
				Let $e(k+j|k)$ be the predicted global state at time $k$ with model~\eqref{Eqn:C full model} under the control $u(e(k)),\cdots, u(e(k+N-1))$. Then the deviation between the real state $e_o(k+j)$ under $u(e_o)$ and $e(k+j|k)$ under $u(e)$ satisfies (\cite{marruedo2002input}) 
				\begin{equation}
					\|e(k+j|k)-e_o(k+j)\|\leq \frac{L_p^j-1}{L_p-1}\varepsilon_w:=\vartheta_j,
				\end{equation}%
				where $e(k|k)=e_o(k)$.
				
				%In the following, we prove the closed-loop robustness when applying the offline learned policy to the MRS. 
				The nominal model~\eqref{Eqn:C full model} can be used for offline policy learning. During the offline learning stage, the terminal state constraint is shrunken as $e(k+N|k)\in{\mathcal{E}}_D$ to ensure the constraint satisfaction, where ${\mathcal{E}}_D(k+N)=\mathcal{E}\ominus \mathcal{D}_{\varepsilon_w}^N$, $\mathcal{D}_{\varepsilon_w}^N=\{y\in\mathbb{R}^n|\|y\|\leq \vartheta_N\}$.  
				In line with~\cite{zhang2021model}, the following robustness property is stated when applying the offline learned policy to the MRS.
				\begin{theorem}[Closed-loop robustness]\label{theorem:2-o} Under Assumptions~\ref{assum:stabilizing_control}-\ref{ASSM:LIPSCHITZ},
					the state evolution, by applying the offline learned stabilizing control policy to~\eqref{eqn:perturbedsys},   converges to the set $\mathcal{D}_{\varepsilon_w}^{\infty}$ as $k\rightarrow +\infty$, i.e., $\lim_{k\rightarrow +\infty} e_o(k)\rightarrow \mathcal{D}_{\varepsilon_w}^{\infty}$. 
				\end{theorem}
				%	\textbf{Proof}. Please refer to Appendix~\ref{app-c}.  \hfill$\square$	
				
				\textbf{Proof}. Let the learned stabilizing control policy be $u^{L}(e)$. In line with~\cite{zhang2021model} and in view of the Lipschitz continuity condition~\eqref{Eqn:lipschitz}, the deviation of the real state $e_o$ under $u^{L}(e_o)$ and the nominal one $e$ under $u^{L}(e)$ is calculated as$\|e_o(1)-e(1|0)\|=\|w(0)\|\leq \varepsilon_w,$ since $e(0|0)=e_o(0)$.
				Then, by induction, one has
				$$\|e_o(j)-e(j|0)\|\leq \|e_o(j-1)-e(j-1|0)\|+\varepsilon_w\leq \frac{L_f^j-1}{L_f-1}\varepsilon_w.$$
				Hence, the real state $e_o(k)$ converges to $\mathcal{D}_{\varepsilon_w}^{\infty}$ as $k\rightarrow +\infty$ since $e(k)$ converges to the origin as $k\rightarrow +\infty$. \hfill$\square$
				
				%Let the learned stabilizing control policy be $\bar u^{L}(e)$. In line with~\cite{zhang2021model} and in view of the Lipschitz continuity condition~\eqref{Eqn:lipschitz}, the deviation of the real state $e_o$ under $\bar u^{L}(e_o)$ and the nominal one $e$ under $\bar u^{L}(e)$ can be computed as 
				%$\|e_o(1)-e(1|0)\|=\|w(0)\|\leq \varepsilon_w,$ since $e(0|0)=e_o(0)$.
				%Then, by induction, one has
				%$$\|e_o(j)-e(j|0)\|\leq \|e_o(j-1)-e(j-1|0)\|+\varepsilon_w\leq \frac{L_f^j-1}{L_f-1}\varepsilon_w.$$
				%Hence, the real state $e_o(k)$ converges to $\mathcal{D}_{\varepsilon_w}^{\infty}$ as $k\rightarrow +\infty$ since $e(k)$ converges to the origin as $k\rightarrow +\infty$. \hfill$\square$
				
				%
				%
				\subsection{Theoretical Analysis of Safe Policy Learning for DMPC}\label{sec:saferl}
				Note that, the implementation details and theoretical results of safe policy learning are omitted due to space limitations. Please refer to Appendix~\ref{appendix B} in the attached material for comprehensive descriptions. We briefly state the main results as follows. Under certain mild assumptions, we derive the condition of learning convergence under the distributed actor-critic implementation. That is, if the learning rates are designed such that
				\begin{subequations}\label{learning rate-inequa}
					\begin{align}
						&\begin{array}{l}
							\bar C_{c,i}:=\gamma_{c,i}^{[1]}\Gamma_{c,i}({\sigma}_{c,i})+
							\gamma_{c,i}^{[2]}\Gamma_{c,i}({\triangledown\mathcal{B}}_{e,i})<1,
						\end{array}\\
						&\begin{array}{l}
							\bar C_{a,i}:=\lambda_{\rm max}(\tilde R_i)\cdot(\gamma_{a,i}^{[1]}  \|{\sigma}_{a,i}\|^2+\\
							\hspace{20mm}\gamma_{a,i}^{[2]} \|{\triangledown\mathcal{B}}_{e,i}\|^2+\gamma_{a,i}^{[3]} \|{\triangledown\mathcal{B}}_{\nu,i}\|^2)<1, 
						\end{array}
					\end{align}
					where $\tilde R_i=2R_i+\mu\triangledown^2\mathcal{B}_{u,i}(\hat {\bar u}_i)$;
				\end{subequations}
				then the approximation errors associated with the actor and critic networks are uniformly ultimate bounded. 
				
				We also provide a practical condition for closed-loop stability verification and prove the robustness of online policy deployment under bounded disturbances.
				
				\subsection{Learning Convergence and Stability Verification}\label{sec:verification}
				
				We have verified the learning convergence and closed-loop stability conditions under the unconstrained and constrained scenarios, for formation control of two mobile robots. Note that the derivation of the convergence and stability conditions in the constrained scenario is deferred in Appendix~\ref{appendix B} within the attached "auxiliary-material.pdf". %Theorems~\ref{THEOM:convergence-o},~\ref{theo-stability-o} as well as Theorem~\ref{THEOM:convergence} (see Appendix~\ref{appendix B} in the auxiliary material) in a formation control scenario of two mobile robots. 
				{\color{black}The communication graph in verification is presented in Panel A of Fig.~\ref{fig:safety_verification}.  The control constraints were limited in the constrained scenario as $-(5,5)\leq u_i\leq (5,5)$, and the collision avoidance constraint was considered. 
					The terminal penalty matrices used in the constrained and unconstrained scenarios were calculated with~\eqref{Eqn:Lya-mod-o} and~\eqref{Eqn:Lya-mod}, respectively. In verification, the weights for the actor and critic were initialized with uniformly distributed random values in the range $[0,\,0.1]$. The maximum iteration $t_{\rm max}$  was 10.
					During the learning process, the values of $C_{c,i}$ and $C_{a,i}$ ($\bar C_{c,i}$ and $\bar C_{a,i}$ in~\eqref{learning rate-inequa}) for $i=1,2$ were smaller than 1, which verified the convergence condition in Theorems~\ref{THEOM:convergence-o} and \ref{THEOM:convergence} (see Appendix~\ref{appendix B}), as shown in Panels B and C of Fig.~\ref{fig:safety_verification}. In addition, the stability condition was verified using various stabilizing control policies, as shown in Panel D of Fig.~\ref{fig:safety_verification}, which reveals that the proposed stability condition is mild and reasonable.
					We also performed 20 repetitive tests to demonstrate the successive learning capacity of our approach. The implementing steps and results have been omitted due to space limitations. Readers may refer to Appendix~\ref{sec:verification-aux} (see ``auxiliary-results.pdf"). }
				\clearpage
				
				\section{Auxiliary Materials}\label{appendix B}
				\subsection{Safe Policy Learning Algorithm and Its Implementation}
				
				\textbf{\emph{1) Safe policy learning}:} In line with~\eqref{Eqn:safempc-o}, we present the safe policy learning counterpart for DMPC in each prediction interval $[k,k+N-1]$. Given an initial control $\bar{\bm u}^0$, for each robot $i\in\mathbb{N}_1^M$, $\tau\in[k,k+N-1]$, the value function and control policy are updated distributedly with iteration step  $t=1,\cdots$:
				
				%			$\forall i\in\mathbb{N}_1^M$
				%			$\tau=k,\cdots,k+N-1$
				%  Compute $e_i(\tau+1)$ with $u_i^t(e_{\scriptscriptstyle \mathcal{N}_i}(\tau))$;
				\begin{enumerate}[(i)]
					\item Parallel value update:
					\begin{subequations}\label{Eqn:safempc}
						%				\scalebox{0.88}{\parbox{0.88\columnwidth}{
								\begin{align}\label{Eqn:value-up-dhp}
									\hspace{-16mm}\begin{array}{ll}
										\hspace{-9mm}	\bar J^{t+1}_i(e_{\scriptscriptstyle \mathcal{N}_i}(\tau))
										=&\bar r_i(\tau)+\bar J_i^{t}(e_{\scriptscriptstyle \mathcal{N}_i}(\tau+1)).
									\end{array}
								\end{align}
								%   }}
						\item	Safe policy update:
						%				\scalebox{0.88}{\parbox{0.88\columnwidth}{%
								\begin{align}\label{Eqn:policy-im-DHP}
									\hspace{-9mm}\begin{array}{ll}
										\hspace{0mm}	(\nu_i^{t+1}(\tau),L_i^{t+1})=
										\argmin{\nu_i(\tau),L_i}\left\{\bar r_i(\tau)+\sum_{j\in\bar{\mathcal{N}}_i}\bar J_j^{t+1}\big(e_{\scriptscriptstyle \mathcal{N}_j}(\tau+1)\big)\right\},\\
										\hspace{0mm} \bar u_i^{t+1}(e_{\scriptscriptstyle \mathcal{N}_i})=\nu_i^{t+1}(e_{\scriptscriptstyle \mathcal{N}_i})+L_i^{t+1}\cdot(\triangledown\mathcal{B}_{e,i}(e_{\scriptscriptstyle\mathcal{N}_i}),\triangledown\mathcal{B}_{\nu,i}(\nu_i)).
									\end{array}	
								\end{align}
								%			}}
					\end{subequations}
				\end{enumerate}
				
				\textbf{\emph{2) Safe policy learning implementation}:} The learning steps of the distributed safe actor-critic implementation are summarized in Algorithm~\ref{alg:d-lpc-AC}.
				After the learning process in the prediction interval $[k,k+N-1]$ being completed, the first control action $\bar u_i(e_{\scriptscriptstyle \mathcal{N}_i}(k))$ computed with~\eqref{Eqn:act-d} is applied to~\eqref{Eqn:LL-nei}. Then, the above learning process is repeated at the subsequent prediction interval $[k+1,k+N]$.
				
				\begin{algorithm}[h]
					\caption{ Safe policy learning implementation.}
					\label{alg:d-lpc-AC}
					\begin{algorithmic}[1]
						\REQUIRE  {	\STATE Initialize $\bar W_{c,i}$ and $\bar W_{a,i}$ with uniformly distributed random matrices,  $i\in\mathbb{N}_1^M$;
							\STATE Set $ \bar{\epsilon}>0$, ${\rm Err}\geq \bar{\epsilon}$, $t=0$,  $t_{\rm max}$;}\\
						%\textbf{for} {$k=1,2,\cdots$} \textbf{do loop}
						\FOR{$k=1,2,\cdots$}
						\STATE Set $\bar W_{c,i}^0=\bar W_{c,i}$ and $\bar W_{a,i}^0=\bar W_{a,i}$, $\forall i\in\mathbb{N}_1^M$;
						\WHILE{${\rm Err}\geq  {\epsilon}\,\vee\, t\leq t_{\rm max}$}
						%		\STATE \textbf{1)} Policy $\pi^t=u^t(k),\cdots,u^t(k+N-1)$;
						\FOR{$\tau=k,\cdots,k+N-1$}
						\STATE  Compute $e_i(\tau+1)$ with $\hat {\bar{u}}_i(e_{\scriptscriptstyle \mathcal{N}_i}(\tau))$ using~\eqref{Eqn:LL-nei}, $\forall\, i\in\mathbb{N}_1^M$; 
						\STATE  Derive $\hat{\bar \lambda}_i(\tau)$ using $e_{\scriptscriptstyle\mathcal{N}_i}(\tau)$ and $\hat{\bar \lambda}_i(\tau+1)$ using the one-step-ahead prediction $e_{\scriptscriptstyle\mathcal{N}_i}(\tau+1)$ with \eqref{eqn:critic}, $\forall\, i\in\mathbb{N}_1^M$;
						\STATE   Calculate $\bar\lambda_i^{d}(\tau)$ with~\eqref{Eqn:lam_d} and $\bar u_{o,i}^d(\tau)$ with~\eqref{Eqn:act-d}, $\forall\, i\in\mathbb{N}_1^M$;
						\STATE Update $\bar W_{c,i}$  with \eqref{Eqn:wc} and $\bar W_{a,i}$ with~\eqref{Eqn:wa}, $\forall\, i\in\mathbb{N}_1^M$;
						\ENDFOR
						\STATE Set $\bar W_{c,i}^{t+1}=\bar W_{c,i}$ and $\bar W_{a,i}^{t+1}=\bar W_{a,i}$, $\forall i\in\mathbb{N}_1^M$;
						\STATE Compute $${\rm Err}=
						\sum_{i=1}^M\|\bar W_{c,i}^{t+1}-\bar W_{c,i}^{t}\|+ \|\bar W_{a,i}^{t+1}-\bar W_{a,i}^{t}\|;$$
						\STATE $t\leftarrow t+1$;
						\ENDWHILE
						\STATE Calculate cost $\bar J(e(k))$ with~\eqref{Eqn:re-cost};
						\IF{Condition deferred in~\eqref{Eqn:sta-con} is violated}
						\STATE Re-initialize $\bar W_{c,i}$ and $\bar W_{a,i}$,  and repeat steps 3-15;
						\ENDIF
						\STATE Update $e_i(k+1)$, $i\in\mathbb{N}_1^M$, by applying $\hat{\bar u}_i(e_{\scriptscriptstyle \mathcal{N}_i}(k))$ to~\eqref{Eqn:LL-nei}.		\ENDFOR
					\end{algorithmic}
				\end{algorithm}
				%
				%    \subsection{Practical Stability Verification Condition}
				\begin{remark}
					In line with the unconstrained scenario, we introduce the following condition for verifying the closed-loop stability, i.e.,
					\begin{equation}\label{Eqn:sta-con}
						\bar J(e(k))-\bar J^b(e^b(k))\leq 0,
					\end{equation}
					$k\in\mathbb{N}$,
					where $\bar J^b(e^b(k))$ is the cost value of~\eqref{Eqn:re-cost} under a baseline $\bar u^{b}(k)$ such that $\bar J^b(e^b(k+1))-\bar J^b(e^b(k))<-\bar s(e^b(k),\bar u^b(k)),$
					where $\bar s(\cdot,\cdot)$ is a class $\mathcal{K}$ function, $\bar J^b(e^b)=\sum_{i=1}^{M}\bar J_i^b(e_{\scriptscriptstyle \mathcal{N}_i}^b)$.
					The design of  $\bar J^b(e^b(k))$ is similar to Remark~\ref{rem:baseline} and is neglected for space limitation.  
				\end{remark}
				\subsection{Theoretical Results of Safe Policy Learning for DMPC}
				In this subsection, we first provide the safety and closed-loop stability guarantees under procedure~\eqref{Eqn:safempc}. Then, practical conditions for the convergence and stability under the distributed actor-critic implementation, i.e., Algorithm~\ref{alg:d-lpc-AC}, are established. Finally, the closed-loop robustness of online policy deployment is proven under bounded disturbances.
				
				\textbf{\emph{1) Safety and stability guarantees under procedure~\eqref{Eqn:safempc}}:}
				To this end, we first introduce a definition of safe control. 
				\begin{definition}[Safe control]\label{def:safe-control}
					At a generic time $k$, a control policy denoted as $\bar{\bm u}(k)$ is safe for~\eqref{Eqn:C full model} if $\bar{\bm u}(k)\in\mathcal{U}^{N}$ and the resulting state evolutions satisfy $e(k+1),\cdots,e(k+N)\in\mathcal{E}^{N-1}\times \mathcal{E}_f$.  
				\end{definition}
				
				In what follows, we show that the control policy in~\eqref{Eqn:safempc} is safe, and the control policy and value function eventually converge to the optimal values respectively, i.e.,  $\bar u^t(e(\tau))={\rm col}_{i\in\mathbb{N}_1^M}\bar u_i^t(e_{\scriptscriptstyle \mathcal{N}_i}(\tau))\rightarrow \bar u^{\ast}(e(\tau))$ and  $\bar J^t(\tau)=\sum_{i=1}^M\bar J_i^t(\tau)\rightarrow \bar J^{\ast}(\tau)$ as $t\rightarrow+\infty$. %, $\forall i\in\mathbb{N}_1^N$.   
				\begin{theorem}[Safety and convergence]\label{theo:safety_convergence}
					Let $\bar{\bm u}^0(k)$ be a safe policy and the initial value function $\bar J^0(e(\tau))\geq\bar r(e(\tau),\bar u^0(\tau))+\bar J^{0}(e(\tau+1))$, $\tau\in[k,k+N-1]$; then under iteration~\eqref{Eqn:safempc}, it holds that
					\begin{enumerate}[(i)]
						\item $\bar J^{t+1}(e(\tau))\leq \bar J^{t}(e(\tau))$;
						\item $\bar{\bm u}^t(k)$ is a safe control policy;
						\item $\bar J^t(e(\tau))\rightarrow \bar J^{\ast}(e(\tau))$ and $\bar u^t(\tau)\rightarrow \bar u^{\ast}(\tau)$ for all $\tau\in[k,k+N]$, as $t\rightarrow +\infty$. \hfill $\blacksquare$
					\end{enumerate}
				\end{theorem}
				
				\textbf{Proof}. The proof arguments are similar to those in Theorem~\ref{theo:safety_convergence-o}. 
				\hfill $\square$
				
				We introduce the following standard assumption for deriving closed-loop stability.
				\begin{assumption}\label{assump:control-invariant}
					There exists a control invariant set of~\eqref{Eqn:LL-nei}  that satisfies constraint~\eqref{eqn:constraints} and contains the origin in the interior. 
				\end{assumption} 
				
				Let $\mathcal{E}_{f,i}$ (i.e., $S_i$)  be selected as a subset  of the control invariant set of~\eqref{Eqn:LL-nei} under~\eqref{Eqn:Lya-mod-o} in the neighbor of the origin and let $\beta_i$ and $P_i$ be such that such that 
				\begin{equation}\label{Eqn:q_in}
					\|P_i\|L_{\phi,i}^2+2\|P_iF_i\|L_{\phi,i}<(\beta_i-1)\lambda_{\rm min}(\bar Q_{i,H}),
				\end{equation} where $\bar Q_{i,H}=\mu (H_{e,i}+K_{\scriptscriptstyle\mathcal{N}_i}^{\top} H_{u,i} K_{\scriptscriptstyle\mathcal{N}_i})+Q_{i}+K_{\scriptscriptstyle\mathcal{N}_i}^{\top} R_{i} K_{\scriptscriptstyle\mathcal{N}_i}$.   
				\begin{theorem} [Closed-loop stability]\label{theo:stability}
					Suppose the prediction horizon $N$ has been selected such that at time $k=0$, the optimal control  $\bar{\bm u}_i^{\ast}(0)\in\mathcal{U}_i^N$, $\forall i\in\mathbb{N}_1^M$ satisfy $e_i(\tau)\in\mathcal{E}_i$, $\tau\in\mathbb{N}_1^{N-1}$, and $e_i(N)\in\mathcal{E}_{f,i}$. Under Assumptions~\ref{assum:stabilizing_control}-\ref{assump:control-invariant}, if, for any $e\in\mathcal{E}_f$,  the next local state evolution,
					denoted as $e^+_{i}$, under control $\bar u_i(e_{\scriptscriptstyle \mathcal{N}_i})=\bar u_i^{\ast}(e_{\scriptscriptstyle \mathcal{N}_i})\in\mathcal{U}_i$ is such that $e^+_{i}\in\mathcal{E}_{f,i}$ $\forall i\in\mathbb{N}_1^M,$ then the global state and control, i.e., $e$ and $\bar u$, converge to the origin asymptotically. \hfill $\blacksquare$
				\end{theorem}
				%\textbf{Proof}. Please refer to Appendix~\ref{app-b}.
				%\hfill$\square$
				\textbf{Proof}. At time $0$, the optimal control policies  $\bar{\bm u}_i^{\ast}(0)$, $\forall i\in\mathbb{N}_1^M$ are feasible. Let at the subsequent time $k=1$, $\bar{\bm u}_i^{\ast}(1)=\bar u_i^{\ast}(e_{\scriptscriptstyle \mathcal{N}_i}(1)),\cdots,\bar u_i^{\ast}(e_{\scriptscriptstyle \mathcal{N}_i}(N-1)),\bar u_i(e_{\scriptscriptstyle \mathcal{N}_i}(N))$ such that $e_{i}(N+1)\in\mathcal{E}_i$. Denoting $\bar J^{f}(e(1))$ as the cost associated with $\bar{\bm u}_i^{f}(1)$, $\forall i\in\mathbb{N}_1^M$, one has
				\begin{equation}\label{Eqn:V-OPT}
					\begin{array}{ll}
						\bar J^{f}(e(1))- \bar J^{\ast}(e(0))=-\bar D(e(0),\bar u^{\ast}(0))+ \bar{\chi}(e(N)),
					\end{array}
				\end{equation}
				where $\bar D(e(0),\bar u(0))=\sum_{i=1}^M(\|e_{\scriptscriptstyle\mathcal{N}_i}(0)\|_{Q_i}^2+\|\bar u_i^{\ast}(0)\|_{R_i}^2
				+\mu \mathcal{B}_{e,i}(e_{\scriptscriptstyle\mathcal{N}_i}(0))+\mu \mathcal{B}_{u,i}(\bar u_i(0)))$, $\bar{\chi}(e(N))=\sum_{i=1}^M\|e_{\scriptscriptstyle\mathcal{N}_i}(N)\|_{Q_i}^2+ \|\bar u_i(N)\|_{R_i}^2+ \mu \mathcal{B}^{[{\rm T}]}_{e,i}(e_i(N+1))-\mu \mathcal{B}^{[{\rm T}]}_{e,i}(e_i(N))+\mu \mathcal{B}_{e,i}(e_{\scriptscriptstyle\mathcal{N}_i}(N))+\mu \mathcal{B}_{u,i}(\bar u_i(N))+\|e_i(N+1)\|_{P_i}^2-\|e_i(N)\|_{P_i}^2$.
				%Specifically, replacing $u_i^{\ast}(e_{\scriptscriptstyle \mathcal{N}_i}(N))$ with $K_{\scriptscriptstyle \mathcal{N}_i}e_{\scriptscriptstyle \mathcal{N}_i}(N)$ in $\bm u_i^{\ast}(1)$ results in feasible control policies $\bm u_i^{f}(1)$, $i\in\mathbb{N}_1^M$. 
				Then, in view of the definition of $\phi_i$ and inline with~\eqref{Eqn:Vf-0}, one has 
				\begin{equation}\label{Eqn:Vf-1}
					\|e_i(N+1)\|_{P_i}^2
					\leq\|e_{\scriptscriptstyle \mathcal{N}_i}(N)\|^2_{F_i^{\top}P_iF_i+(\beta_i-1)\bar Q_{i,H}}.
				\end{equation}
				In view of~\eqref{Eqn:Lya-mod},~\eqref{Eqn:q_in}, and~\eqref{Eqn:Vf-1}, it holds that   
				\begin{equation}\label{Eqn:V-MONO}
					\begin{array}{ll}
						\bar{\chi}(e(N))\leq \sum_{i=1}^M\mu(\mathcal{B}^{[{\rm T}]}_{e,i}(e_i(N+1))- \mathcal{B}^{[{\rm T}]}_{e,i}(e_i(N))). %+\\
						%\sum_{i=1}^M\|e_i(N)\|^2_{F_i^{\top} P_{i}F_i-\bar{P}_{i}+\beta_i(\mu H_{\scriptscriptstyle\mathcal{N}_i}+Q_{i}+K_{\scriptscriptstyle\mathcal{N}_i}^{\top} R_{i} K_{\scriptscriptstyle\mathcal{N}_i})-\Gamma_{\scriptscriptstyle\mathcal{N}_i}=0}
					\end{array}
				\end{equation}
				Note that, for $e_i(N),e_i(N+1)\in\mathcal{E}_{f,i}$, it follows that
				\begin{equation}
					\begin{array}{ll}
						\sum_{i=1}^M(\mathcal{B}^{[{\rm T}]}_{e,i}(e_i(N+1))- \mathcal{B}^{[{\rm T}]}_{e,i}(e_i(N)))\\
						%=
						%\mathcal{B}_f(e(N+1))- \mathcal{B}_f(e(N))\\
						=-\text{log}(1-{e(N)^{\top}}F^{\top}SFe(N))-\text{log}(1-e(N)^{\top}Se(N))\\
						<0,
					\end{array}
				\end{equation}
				since $F^{\top}SF\leq S$, where $S=\text{diag}\{S_1,\cdots,S_M\}$ and $F$ is such that $Fx=\text{col}_{i\in\mathbb{N}_1^M}(F_{\scriptscriptstyle \mathcal{N}_i}e_{\scriptscriptstyle \mathcal{N}_i})$.
				in view of~\eqref{Eqn:Lya-mod} one has
				%\begin{equation}\label{Eqn:V-MONO-1}
				%	\begin{array}{ll}
					$$\bar J^{f}(e(1))- \bar J^{\ast}(e(0))\leq-\bar D(e(0),\bar u^{\ast}(0)), $$
					%	\end{array}
				%\end{equation}
				which by induction leads to $\bar J^{f}(k+1)- \bar J^{f}(k)\rightarrow0$ as $k\rightarrow+\infty$. Hence, $e$, $\bar u$ converge to the origin asymptotically. 
				\hfill$\square$
				
				\textbf{\emph{2) Convergence and stability of Algorithm~\ref{alg:d-lpc-AC}}:}
				We first prove the convergence of Algorithm~\ref{alg:d-lpc-AC} in each prediction interval. To this end,  we write the local optimal costate and control policy  for all $\tau\in[k,k+N-1]$ and $i\in\mathbb{N}_1^M$ as
				$$\bar{\lambda}_i^{\ast}(e_{\scriptscriptstyle\mathcal{N}_i}(\tau))=({\bar W_{c,i}^{\ast})}^{\top}h_{c,i}(e_{\scriptscriptstyle\mathcal{N}_i}(\tau),\tau)+\bar{\kappa}_{c,i}(\tau)$$
				$$\bar u_i^{\ast}(e_{\scriptscriptstyle\mathcal{N}_i}(\tau))=({\bar W_{a,i}^{\ast})}^{\top}h_{a,i}(e_{\scriptscriptstyle\mathcal{N}_i}(\tau),\tau)+\bar{\kappa}_{a,i}(\tau),$$ where $\bar W_{c,i}^{\ast}$ and $\bar W_{a,i}^{\ast}$ are the optimal weights of $\bar W_{c,i}$ and $\bar W_{a,i}$, $\bar{\kappa}_{c,i}$ and $\bar{\kappa}_{a,i}$ are the associated reconstruction errors.
				\begin{assumption}[Weights and reconstruction errors]\label{assum:network} For all $i\in\mathbb{N}_1^M$,  it holds that
					\begin{enumerate}[(i)]
						\item $\|\bar W_{c,i}^{\ast}\|\leq \bar{W}_{c,i}^{\scriptscriptstyle[m]}$,  $\|\triangledown {\mathcal{B}}_{e,i}\|\leq  {\mathcal{B}}_{e,i}^{\scriptscriptstyle[m]}$,  $\|\bar{\kappa}_{c,i}\|\leq  \bar{\kappa}_{c,i}^{\scriptscriptstyle[m]}$;
						\item $\|\bar W_{a,i}^{\ast}\|\leq \bar W_{a,i}^{\scriptscriptstyle[m]}$,  $\|\triangledown {\mathcal{B}}_{\nu,i}\|\leq  {\mathcal{B}}_{\nu,i}^{\scriptscriptstyle[m]}$,  $\|\bar{\kappa}_{a,i}\|\leq  \bar{\kappa}_{a,i}^{\scriptscriptstyle[m]}$. 
					\end{enumerate} 
				\end{assumption}
				Define $\tilde {\bar W}_{\star,i}= \bar W^{\ast}_{\star,i}- \bar W_{\star,i}$, $\star=a,c$ in turns.  For the sake of simplicity, we consider the control scenario under a linear control constraint, i.e., ${\Xi_{u,i}^j}(\bar u_i)=E_{u,i}^j\bar u_i$ for all $j\in\mathbb{N}_1^{q_{u,i}}$.

				\begin{theorem}[Convergence of Algorithm~\ref{alg:d-lpc-AC}]\label{THEOM:convergence}
					Under Assumptions~\ref{ASSUM: INPUT MAPPING-o}-\ref{assum:network},  if  the learning rates are designed such that
					\begin{align*}
						&\begin{array}{l}
							\bar C_{c,i}:=\gamma_{c,i}^{[1]}\Gamma_{c,i}({\sigma}_{c,i})+
							\gamma_{c,i}^{[2]}\Gamma_{c,i}({\triangledown\mathcal{B}}_{e,i})<1,
						\end{array}\\
						&\begin{array}{l}
							\bar C_{a,i}:=\lambda_{\rm max}(\tilde R_i)\cdot(\gamma_{a,i}^{[1]}  \|{\sigma}_{a,i}\|^2+\\
							\hspace{20mm}\gamma_{a,i}^{[2]} \|{\triangledown\mathcal{B}}_{e,i}\|^2+\gamma_{a,i}^{[3]} \|{\triangledown\mathcal{B}}_{\nu,i}\|^2)<1, 
						\end{array}
					\end{align*}
					where $\tilde R_i=2R_i+\mu\triangledown^2\mathcal{B}_{u,i}(\hat {\bar u}_i)$;
					then the terms $$
					\begin{array}{ll}
						\bar{\xi}_{a,i}(\tau)=\tilde {\bar W}_{a,i}^{\top}(\tau)h_{a,i}(\tau),\vspace{1mm}\\ \bar{\xi}_{c,i}(\tau)=-\vec f_{\bar{\mathcal{N}}_i}\tilde{\bar W}_{c,i}^{\top}(\tau) h_{c,i}^{+}(\tau)+\tilde{\bar W}_{c,i}^{\top}(\tau) h_{c,i}(\tau),
					\end{array}$$ are uniformly ultimate bounded,  as the iteration step $t\rightarrow +\infty$ (see Algorithm~\ref{alg:d-lpc-AC}). Moreover, if $\bar{\kappa}_{a,i},\,\bar{\kappa}_{c,i}\rightarrow 0$,  
					\begin{equation*}
						\bar{\xi}_{a,i}\rightarrow 0\ \text{and}\ \bar {\xi}_{c,i}\rightarrow  0,
					\end{equation*}
					as $t\rightarrow +\infty$.  \hfill $\blacksquare$
				\end{theorem}
				%	\textbf{Proof}. Please refer to Appendix~\ref{app-d}.
				%  \hfill $\square$
				
				\textbf{Proof}.
				Define a collective Lyapunov function as
				\begin{equation}\label{Eqn:v_{k}}
					\begin{array}{ll}
						\bar V(\tau)=\sum_{i=1}^M\bar V_{c,i}(\tau)+\bar V_{a,i}(\tau),
					\end{array}
				\end{equation}
				where
				\begin{equation*}
					\begin{array}{ll}
						\bar V_{c,i}=\text{tr}\left(\sum_{j=1}^2 1/\gamma_{c,i}^{[j]}(\tilde{W}_{c,i}^{[j]})^{\top}\tilde W_{c,i}^{[j]}\right),\vspace{1mm}\\
						\bar V_{a,i}=\text{tr}\left(\sum_{j=1}^3 1/\gamma_{a,i}^{[j]}(\tilde{\bar W}_{a,i}^{[j]})^{\top}\tilde {\bar W}_{a,i}^{[j]}\right).
					\end{array}
				\end{equation*}
				In view of the update rule~\eqref{Eqn:wc}, letting $\Delta \bar V_{c,i}(\tau)=\bar V_{c,i}(\tau+1)-\bar V_{c,i}(\tau)$, one writes
				\begin{equation}\label{Eqn:deltaVc(k)}
					\Delta \bar V_{c,i}=\text{tr}\left(\sum_{j=1}^2 2(\tilde{\bar W}_{c,i}^{[j]})^{\top}\frac{\partial \bar{\delta}_{c,i}}{\partial \bar W_{c,i}^{[j]}}+\gamma_{c,i}^{[j]}\|\frac{\partial \bar{\delta}_{c,i}}{\partial \bar W_{c,i}^{[j]}}\|_F^2\right).
				\end{equation}
				First note that 
				\begin{subequations}\label{Eqn:delta(k)}
					\begin{align}
						\frac{\partial \bar{\delta}_{c,i}}{\partial \bar W_{c,i}^{[1]}}&=-2\sigma_{c,i}\bar{\varepsilon}_{c,i}^{\top}+2\sigma_{c,i}^{+}\bar{\varepsilon}_{c,i}^{\top}\vec f_{\bar{\mathcal{N}}_i},\\
						\frac{\partial \bar{\delta}_{c,i}}{\partial \bar W_{c,i}^{[2]}}&=-2\triangledown\mathcal{B}_{e,i}\bar{\varepsilon}_{c,i}^{\top}+2\triangledown\mathcal{B}_{e,i}^{+}\bar{\varepsilon}_{c,i}^{\top}\vec f_{\bar{\mathcal{N}}_i},
					\end{align}
					where $\vec f_{\bar{\mathcal{N}}_i}$ is defined previously in~\eqref{equ:assum:model}.
				\end{subequations}
				%
				%where $\Delta h_{c,k}=\gamma^Lh_{c,k+L}-h_{c,k}$.
				
				Moreover, in view of the definition of $\varepsilon_{c,i}$ and of Assumption~\ref{assum:network}, it follows that
				\begin{equation}\label{Eqn:epsilonc(k)}
					\begin{array}{ll}
						\bar{\varepsilon}_{c,i}&=\bar{\lambda}_i^d-\bar{\lambda}_i^{\ast}+ \bar{\lambda}_i^{\ast}-\hat{\bar{\lambda}}_i\vspace{1mm}\\
						&=\bar{\xi}_{c,i}+\Delta \bar{\kappa}_{c,i},
					\end{array}
				\end{equation}
				where $\bar{\xi}_{c,i}=-\vec f_{\bar{\mathcal{N}}_i}\tilde{\bar W}_{c,i}^{\top} h_{c,i}^{+}+\tilde{\bar W}_{c,i}^{\top} h_{c,i}$,
				$\Delta \bar{\kappa}_{c,i}=\bar{\kappa}_{c,i}^{[m]}-\vec f_{\bar{\mathcal{N}}_i}(\bar{\kappa}_{c,i}^{+})^{[m]}$. 
				
				Taking~\eqref{Eqn:delta(k)} with~\eqref{Eqn:epsilonc(k)} into~\eqref{Eqn:deltaVc(k)},  in view of Assumption~\ref{assum:network}, one promptly has
				\begin{equation}\label{Eqn:delta_Vc_final(k)}
					\begin{array}{lll}
						\hspace{-2mm}\Delta \bar V_{c,i}
						&\hspace{-2mm}\leq&\hspace{-2mm}-4\bar{\xi}_{c,i}^{\top}(\bar{\xi}_{c,i}+\Delta \bar{\kappa}_{c,i})+\\
						&&\hspace{1mm}4(\gamma_{c,i}^{[1]}\Gamma_{c,i}({\sigma}_{c,i})+\gamma_{c,i}^{[2]}\Gamma_{c,i}({\triangledown\mathcal{B}}_{e,i}))\|\bar{\xi}_{c,i}+\Delta \bar{\kappa}_{c,i}\|^2\vspace{1mm}\\
						\hspace{-2mm}&\hspace{-2mm}\leq&\hspace{-2mm}-\bar c_{c,i}\|\bar{\xi}_{c,i}\|^2+\bar{\epsilon}_{c,i},
					\end{array}
				\end{equation}
				where $\bar c_{c,i}=4-4(\gamma_{c,i}^{[1]}\Gamma_{c,i}({\sigma}_{c,i})+\gamma_{c,i}^{[2]}\Gamma_{c,i}(\triangledown{\mathcal{B}}_{e,i}))-\beta_{c,i}$,  $\bar{\epsilon}_{c,i}=1/\beta_{c,i}(1+(4\beta_{c,i}-4)\gamma_{c,i}^{[1]}\Gamma_{c,i}({\sigma}_{c,i})+(4\beta_{c,i}-4)\gamma_{c,i}^{[2]}\Gamma_{c,i}({\triangledown\mathcal{B}}_{e,i}))\|\Delta \bar{\kappa}_{c,i}\|^2$, $\beta_{c,i}>0$ is a tuning constant. The last inequality in~\eqref{Eqn:delta_Vc_final(k)} is due to Young's inequality property.

				\begin{figure*}[h]
					\centerline{\includegraphics[scale=0.53]{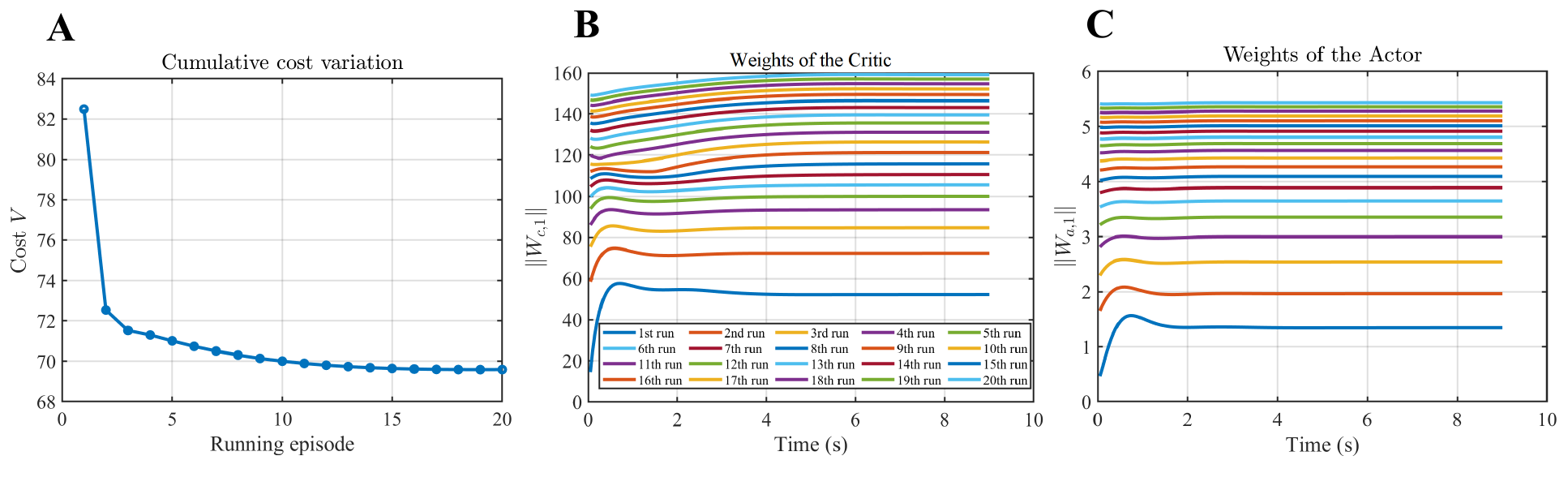}}
					\caption{{\color{black}Learning convergence under the random initialization. A: The cumulative cost $V=1/M\sum_{j=1}^{N_{\rm sim}}r_i(e_{\scriptscriptstyle \mathcal{N}_i}(j),u_i(j))$ with running episodes from 1 to 20. B: The weights of the critic with running episodes from 1 to 20.  C: The weights of the actor with running episodes from 1 to 20. In each running experiment, the weights of the critics and actors are updated and converge rapidly to constant values. Moreover, the weights can be persistently updated under repetitive tests to achieve better performance.}}
					\label{fig:convergence_verification} 
				\end{figure*}
				\begin{figure*}[h]
					\centerline{\includegraphics[scale=0.53]{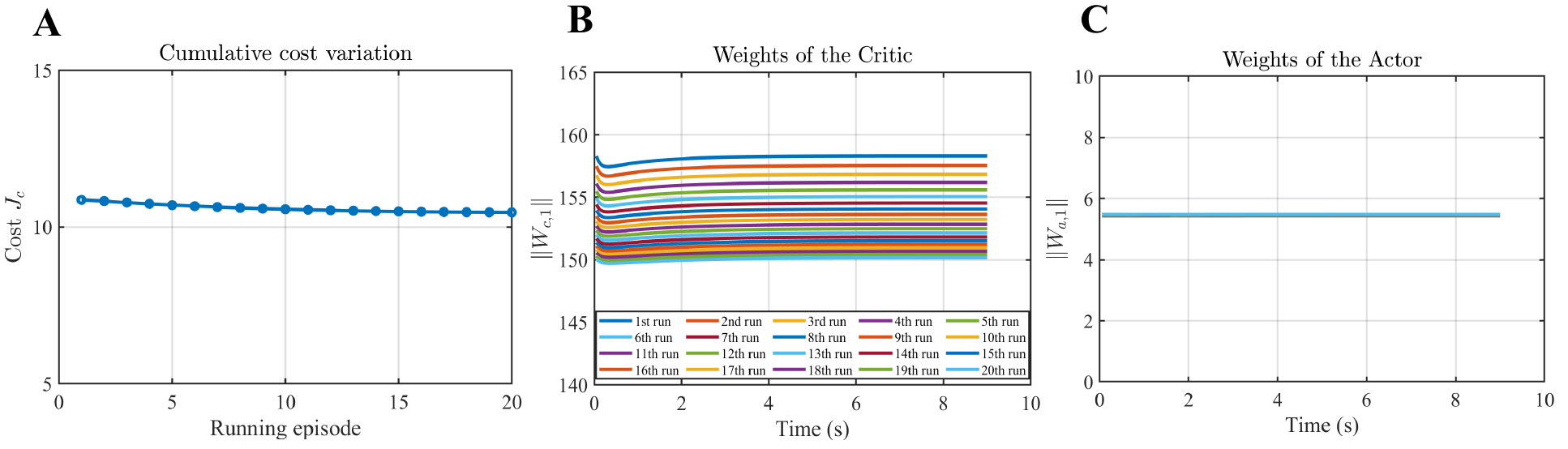}}
					\caption{{\color{black}Learning convergence using the learned weights as initialization. A: The cumulative cost $V=1/M\sum_{j=1}^{N_{\rm sim}}r_i(e_{\scriptscriptstyle \mathcal{N}_i}(j),u_i(j))$ with running episodes from 1 to 20. B: The weights of critics with running episodes from 1 to 20.  C: The weights of actors with running episodes from 1 to 20.}}
					\label{fig:convergence_verification_ii} 
				\end{figure*}
				
				\begin{figure*}[h!]
					\centerline{\includegraphics[scale=0.5]{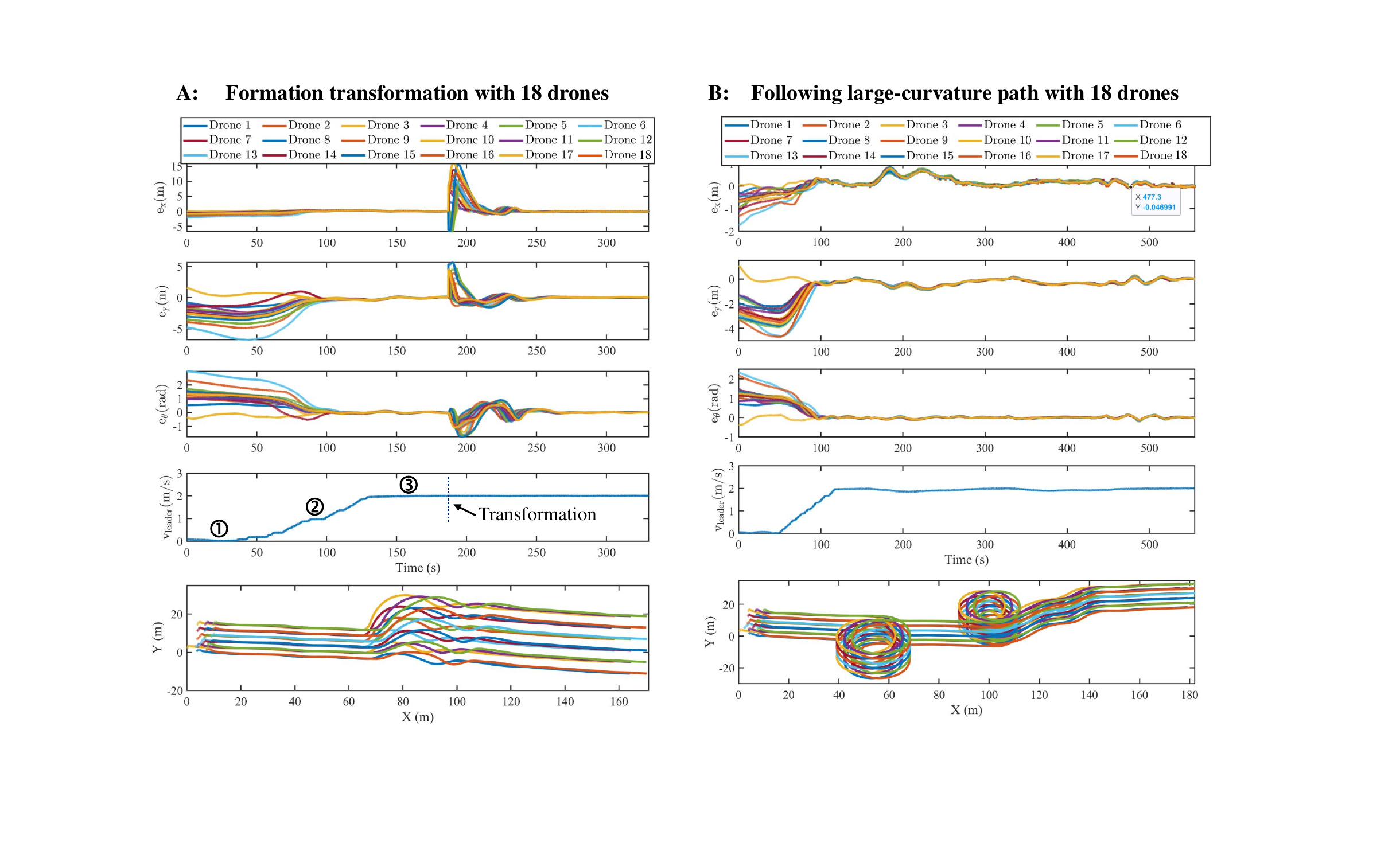}}
					\caption{State errors and paths of the multirotor drones in Gazebo. We directly deployed the learned control policy to the formation control of multirotor drones with $M=18$.  Stage \textcircled{1}: Leader in hover mode; Stage \textcircled{2}: Leader speeds up with keyboard control; Stage \textcircled{3}: Leader at a constant speed of 2 m/s. %; Stage D: Change to a new formation shape.
					}
					\label{fig:drone_experi-18} 
				\end{figure*}
				To  compute $\Delta \bar{V}_{a,i}$, we first write 
				\begin{equation}\label{Eqn:delta_Va_final(k)}
					\Delta \bar{V}_{a,i}=\text{tr}\left(\sum_{j=1}^3 2(\tilde{\bar W}_{a,i}^{[j]})^{\top}\frac{\partial \bar{\delta}_{a,i}}{\partial \bar W_{a,i}^{[j]}}+\gamma_{a,i}^{[j]}\|\frac{\partial \bar{\delta}_{a,i}}{\partial \bar W_{a,i}^{[j]}}\|_F^2\right).
				\end{equation}
				Inline with~\eqref{Eqn:delta(k)}, one has
				\begin{subequations}\label{Eqn:delta_wA}
					\begin{align}
						\frac{\partial \bar{\delta}_{a,i}}{\partial \bar W_{a,i}^{[1]}}
						&=-2\sigma_{a,i}\bar{\varepsilon}_{a,i}^{\top}\tilde R_i,\\
						\frac{\partial \bar{\delta}_{a,i}}{\partial \bar W_{a,i}^{[2]}}
						&=-2\triangledown\mathcal{B}_{e,i}\bar{\varepsilon}_{a,i}^{\top}\tilde R_i,\\
						\frac{\partial \bar{\delta}_{a,i}}{\partial \bar W_{a,i}^{[3]}}
						&=-2\triangledown\mathcal{B}_{\nu,i}\bar{\varepsilon}_{a,i}^{\top}\tilde R_i.
					\end{align}
				\end{subequations}
				%where $\bar R_i=2R_i+\mu\triangledown^2\mathcal{B}_{u,i}(\hat u_i)$.
				
				Taking~\eqref{Eqn:delta_wA} into~\eqref{Eqn:delta_Va_final(k)}, it holds that
				\begin{equation}\label{Eqn:delta_va}
					\Delta \bar V_{a,i}=-\bar c_{a,i}\|\bar{\varepsilon}_{a,i}\|_{\tilde R_i}^2,
				\end{equation}
				where $\bar c_{a,i}=4-4\lambda_{\rm max}(\tilde R_i)\cdot(\gamma_{a,i}^{[1]} {\|\sigma}_{a,i}\|^2+\gamma_{a,i}^{[2]} {\|\triangledown\mathcal{B}}_{e,i}\|^2+\gamma_{a,i}^{[3]} {\|\triangledown\mathcal{B}}_{\nu,i}\|^2)$.
				
				Combining~\eqref{Eqn:delta_Vc_final(k)} and~\eqref{Eqn:delta_va}, leads to
				\begin{equation}
					\Delta \bar V=\sum_{i=1}^M-(\bar c_{c,i}\|\bar{\xi}_{c,i}\|^2+\bar c_{a,i}\|\bar{\varepsilon}_{a,i}\|_{\tilde R_i}^2)+\bar{\epsilon}_{c}.
				\end{equation}
				where $\bar{\epsilon}_{c}=\sum_{i=1}^M\bar{\epsilon}_{c,i}$.
				Hence, in view of~\eqref{learning rate-inequa} and setting $\beta_{c,i}$ small,  for any $i\in\mathbb{N}_1^M$, it follows that
				\begin{equation}
					\|\bar{\xi}_{c,i}\|\leq \sqrt{\frac{\bar{\epsilon}_{c}}{\bar c_{c,i}}}\ \ \text{and}\ \ \|\bar{\varepsilon}_{a,i}\|\leq \sqrt{\frac{\bar{\epsilon}_{c}}{\bar c_{a,i}}}, 
				\end{equation}
				as the iteration step $t\rightarrow +\infty$.
				Note that one has
				\begin{equation*}
					\begin{array}{lll}
						\bar{\varepsilon}_{a,i}&\hspace{-2mm}=&\hspace{-2mm}\bar u_{o,i}^d-\bar u_{o,i}^{\ast}+\bar u_{o,i}^{\ast}-\bar u_{o,i}\\
						&\hspace{-2mm}=&\hspace{-2mm}\bar u_{o,i}^d-\bar u_{o,i}^{\ast}+2R_i (\bar{\xi}_{a,i}+\bar{\kappa}_{a,i})+\\
						&&\hspace{20mm}\mu(\triangledown\mathcal{B}_{u,i}(\bar u^{\ast}_{i})-\triangledown\mathcal{B}_{u,i}(\hat {\bar u}_i)),
					\end{array}
				\end{equation*}
				
				%where $g_c=\triangledown_u f(e,u)^{\top}(\triangledown h_c^+)^{\top}$,  $\bar \kappa_1=\triangledown_u f(e,u)^{\top}\triangledown\kappa_c^++2R\kappa_a$. 
				Moreover, in view of the definition of $\mathcal{B}_{u,i}$ and ${\Xi_{u,i}^j}(\bar u_i)=E_{u,i}^j\bar u_i$, it holds that
				\begin{equation}\label{Eqn:barrier-diff}
					\bar{\mathcal{B}}_{u,i}=\triangledown\mathcal{B}_{u,i}(\bar u_i^{\ast})-\triangledown \mathcal{B}_{u,i}(\hat {\bar u}_i)=-C_i(\bar{\xi}_{a,i}+\bar{\kappa}_{a,i}),
				\end{equation}
				where
				$C_i=\sum_{i=1}^{q_{u,i}}(E_{u,i}^j)^{\top}E_{u,i}^j/(G^i(\bar u_i^{\ast})G^i(\bar u_i))$ for $\bar\sigma\geq\kappa_{i}$ and $C_i=2H_{u,i}$ otherwise. 
				
				Hence, with~\eqref{Eqn:barrier-diff}, one has
				\begin{equation}\label{Eqn:epsilona(k)}
					\begin{array}{lll}
						\varepsilon_{a,i}=\bar u_{o,i}^d-\bar u_{o,i}^{\ast}+R_{d,i}\bar{\xi}_{a,i}+R_{d,i}\bar{\kappa}_{a,i},
					\end{array}
				\end{equation}
				where $R_{d,i}=2R_i-\mu C_i$.
				
				Note that
				\begin{equation}\label{Eqn: u0d-u0}
					\begin{array}{ll}
						\|\bar u_{o,i}^d-\bar u_{o,i}^{\ast}\|&\hspace{-2mm}=\|-\sum_{j\in\bar{\mathcal{N}}_i}g_{i}^{\top}(e_i) (\tilde{\lambda}^{[i]}_j)^+\|,\vspace{2mm}\\
						&\hspace{-2mm}\leq\sum_{j\in\bar{\mathcal{N}}_i}\|g_{i}(e_i)\| (\sqrt{\frac{\bar{\epsilon}_{c}}{\bar c_{a,i}}}+\bar{\kappa}_{c,i}^{[m]}):=\bar Y_{u,i},
					\end{array}
					%\sum_{j\in \bar{\mathcal{N}}_i}g_i^{\top}\tilde \lambda^{[j]}_i(\tau+1)
				\end{equation} where $\tilde{\bar{\lambda}}^{[i]}_j=(\bar{\lambda}^{\ast}_j)^{[i]}-\hat{\bar{\lambda}}^{[i]}_j$. 
				
				In view of~\eqref{Eqn:epsilona(k)} and~\eqref{Eqn: u0d-u0}, for any $i\in\mathbb{N}_1^M$, one consequently has
				\begin{equation*}
					\begin{array}{ll}
						\|\bar{\xi}_{a,i}\|&\hspace{-2mm}\leq \frac{1}{|\rho(R_{d,i})|}(\|R_{d,i}\bar{\kappa}_{a,i}\|+\|\bar u_{o,i}^d-\bar u_{o,i}^{\ast}\|+\sqrt{\frac{\bar{\epsilon}_{c}}{\bar c_{a,i}}})\vspace{2mm}\\
						&\hspace{-2mm}\leq \frac{1}{|\rho(R_{d,i})|}(\|R_{d,i}\|\bar{\kappa}_{a,i}^{[m]}+\bar Y_{u,i}+\sqrt{\frac{\bar{\epsilon}_{c}}{\bar c_{a,i}}}),
					\end{array}
				\end{equation*}
				as the iteration step $t\rightarrow +\infty$, where $\rho(R_{d,i})$ is the minimal (maximal) eigenvalue of $R_{d,i}$ if it is positive-definite (negative-definite).
				
				Consequently, if $\bar{\kappa}_{a,i},\, \bar{\kappa}_{c,i} \rightarrow 0$, it immediately follows that
				\begin{equation*}
					\bar{\xi}_{a,i}\rightarrow 0\ \text{and}\ \bar{\xi}_{c,i}\rightarrow 0,
				\end{equation*}
				as $t\rightarrow +\infty$.
				\hfill $\square$  
				\begin{figure*}[h!]
					\centerline{\includegraphics[scale=0.32]{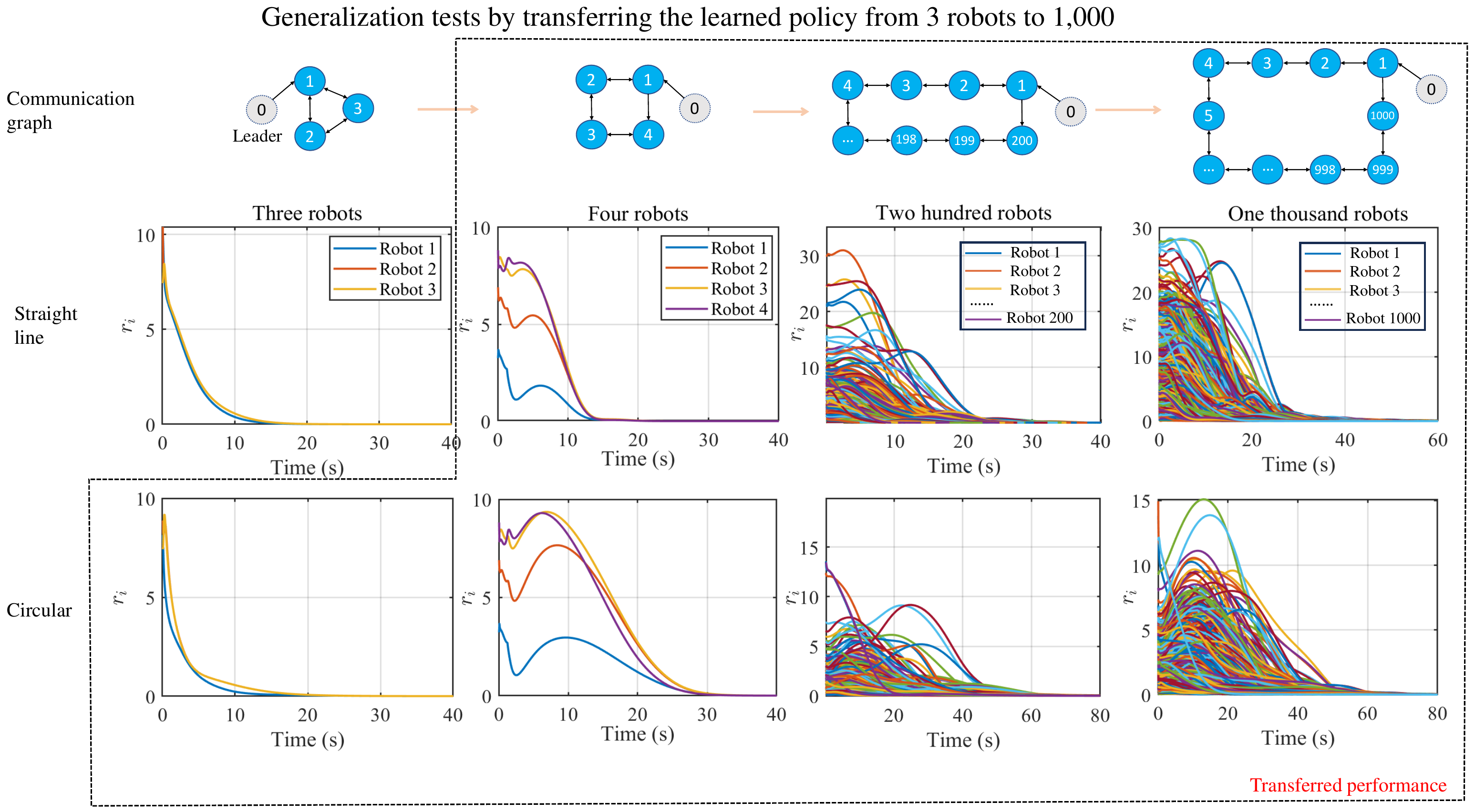}}
					\caption{{\color{black}Transferred performance of straight-line formation of 3 robots to the circular formation of 3 robots and different formation scenarios of 4, 200, and 1,000 robots.}}
					\label{fig:trajectory-gen-aux} 
				\end{figure*}
				\begin{theorem}[Closed-loop stability under Algorithm~\ref{alg:d-lpc-AC}]\label{theo-stability}
					Under Assumptions~\ref{assum:stabilizing_control}-\ref{assump:control-invariant}, if condition~\eqref{Eqn:sta-con} is fulfilled,  then the global state and control under Algorithm~\ref{alg:d-lpc-AC}, i.e., $e$ and $\bar u$, converge to the origin asymptotically.   \hfill $\blacksquare$
				\end{theorem}
				%\textbf{Proof.} Please refer to Appendix~\ref{app-e}.   \hfill $\square$
				%
				\textbf{Proof.}  %First, note that starting from a stabilizing control policy sequence $\bm u^{b}_i(0)$, one can promptly design the control policy $\bm u^{b}_i(k)$, $\forall\, k\in\mathbb{N}$ such that the control problem is recursively feasible. Consequently, inline with Theorem~\ref{theo:stability}, one has
				Note that
				%\begin{equation}\label{Eqn:stabilizing-converge}
				$\bar J^b(k+1|k)-\bar J^b(k|k)\leq-\bar s(e^{b}(k),\bar u^{b}(k)).$
				%\end{equation}
				Hence, it holds that $e^b$ and $\bar u^b$ converge to the origin asymptotically.
				By condition~\eqref{Eqn:sta-con}, one has 
				\begin{equation}
					\begin{array}{ll}
						\bar J(k+1|k)&\leq \bar J^b(k+1|k)\\
						&\leq \bar J^b(k|k)-\bar s(e^{b}(k),\bar u^{b}(k)).
					\end{array}
				\end{equation}
				As $e^b(k)\rightarrow 0$ and $\bar u^b(k)\rightarrow 0$ as $k\rightarrow +\infty$, it follows that $\bar J(k|k)\rightarrow 0$ as $k\rightarrow+\infty$. Consequently,  $e$ and $\bar u$ converge to the origin asymptotically.\hfill $\square$
				
				\textbf{\emph{3) Closed-loop robustness in perturbed scenario}:}   During the offline learning stage, the state constraint is shrunken as $e(k+j|k)\in\bar{\mathcal{E}}$ to ensure the constraint satisfaction, where $\bar{\mathcal{E}}(k+j)=\mathcal{E}\ominus \mathcal{D}_{\varepsilon_w}^j$, $\mathcal{D}_{\varepsilon_w}^j=\{y\in\mathbb{R}^n|\|y\|\leq \vartheta_j\}$.  
				The following results are stated in line with~\cite{zhang2021model}.
				\begin{theorem}[Closed-loop robustness]\label{theorem:2} Under Assumptions~\ref{assum:stabilizing_control}-\ref{ASSM:LIPSCHITZ},
					the state evolution, by applying the offline learned stabilizing control policy to~\eqref{eqn:perturbedsys},   converges to the set $\mathcal{D}_{\varepsilon_w}^{\infty}$ as $k\rightarrow +\infty$, i.e., $\lim_{k\rightarrow +\infty} e_o(k)\rightarrow \mathcal{D}_{\varepsilon_w}^{\infty}$. \hfill $\blacksquare$
				\end{theorem}
				\textbf{Proof}. The proof argument is similar to that in Theorem~\ref{theorem:2-o}. \hfill$\square$

				{\color{black}\subsection{Auxiliary Results for Learning Convergence and Stability Verification}\label{sec:verification-aux}
					
					In addition to learning convergence and stability verification in Appendix~\ref{sec:verification}, 
					we further conducted 20 repetitive tests to demonstrate the successive learning capacity of our approach. In these tests, the weights generated from the latest experiment were used as initial values for a new experimental run. The results, shown in Fig.~\ref{fig:convergence_verification}, indicate that both the cumulative cost and the weights of the actor and critic converge to the vicinity of a local optimum after 15 episodes.
					
					Furthermore, we carried out 20 more experiments to evaluate the generalizability of our learning policy under different initial state conditions. In this scenario, the weights learned from the last experimental run in Fig.~\ref{fig:convergence_verification} were used for weight initialization and a maximum iteration of $t_{\rm max}=1$ was set to accelerate the learning process. The results in Fig.~\ref{fig:convergence_verification_ii} reveal that both the cost and weights converge, showing no significant changes during learning. This observation suggests that using the learned weights for initialization can reduce the number of iterations in the prediction horizon, enabling more efficient online policy learning.}

				\subsection{Auxiliary Results for Policy Deployment to Multirotor Drones in Gazebo }\label{sec:gazebo-auxiliary}
				The auxiliary results for deploying the learned control policy to 18 drones are displayed in Fig.~\ref{fig:drone_experi-18}. In the formation transformation scenario (Scenario A in Fig.~\ref{fig:drone_experi-18}), the state errors approach the origin together with the speed-up process of the leader and then recover promptly from a short transient formation transformation. The state errors remain close to the origin in the subsequent scenario with a large-curvature path (Scenario B in Fig.~\ref{fig:drone_experi-18}).
				{\color{black}\subsection{Auxiliary Results on Transferability from 3 Robots to 1,000}\label{sec:transfer}
					In addition to the transferability test from 2 robots to 1,000 in Section~\ref{sec:simu}, we also verified the transferability from 3 robots to 1,000. The training was performed for 3 robots' formation control in a straight-line formation scenario. The learned weighting matrix of the actor for the second robot is $W_{a,2}=[w_{1}\quad w_{2} \quad w_{3}]^{\top}$, where 
					\begin{equation*}
						\begin{array}{ll}
							w_{1}=\begin{bmatrix}
								2.79&1.07&-0.08&7.78\\
								-0.46&0.76&3.05&0.75
						\end{bmatrix}\end{array}
					\end{equation*}
					\begin{equation*}
						\begin{array}{ll}
							w_{2}=\begin{bmatrix}
								1.52&-0.1&0.16&4.36\\
								0.4&0.67&0.07&-0.59
						\end{bmatrix}\end{array}
					\end{equation*}
					\begin{equation*}
						\begin{array}{ll}
							w_{3}=\begin{bmatrix}
								0.38&0.06&-0.42&-0.15\\
								-0.1&0.1&-0.11&-0.26
							\end{bmatrix}.
						\end{array}
					\end{equation*}
					The weighting matrix $W_{a,2}$ was then used directly to construct control policies for formation control with 4, 200, and 1,000 robots in both straight-line and circular formation scenarios. In scenarios with 4, 200, and 1,000 robots, we adopted a similar communication network for the 3-robot scenario where the first robot has two neighbors while other robots have three neighbors. Therefore, the weighting matrix used for policy deployment was constructed as $W_{a,1}=[w_{1}\quad w_{2}]^{\top}$ for the first robot and $W_{a,i}=[w_{1}\quad w_{2}\quad w_{3}]^{\top}$ for the other robots. 
					%Since each robot in the 3-robot scenario has three neighbors, the input of each actor is in the $\mathbb{R}^{12}$ space. In the policy deployment to scenarios with 4, 200, 1,000 robots, we adopted a similar communication network where each robot receives information from three neighbors. As a result, the input dimension of each local actor equals that in the 3 robots scenario. Then, we utilized the learned weight of the first actor to generate all the local control policies in scenarios with 4, 200, 1,000 robots.
					As shown in Fig.~\ref{fig:trajectory-gen-aux}, the transferred policy stabilizes the formation control system on various scales of robots.
					\begin{figure}[h] \centering\includegraphics[width=0.45\textwidth]{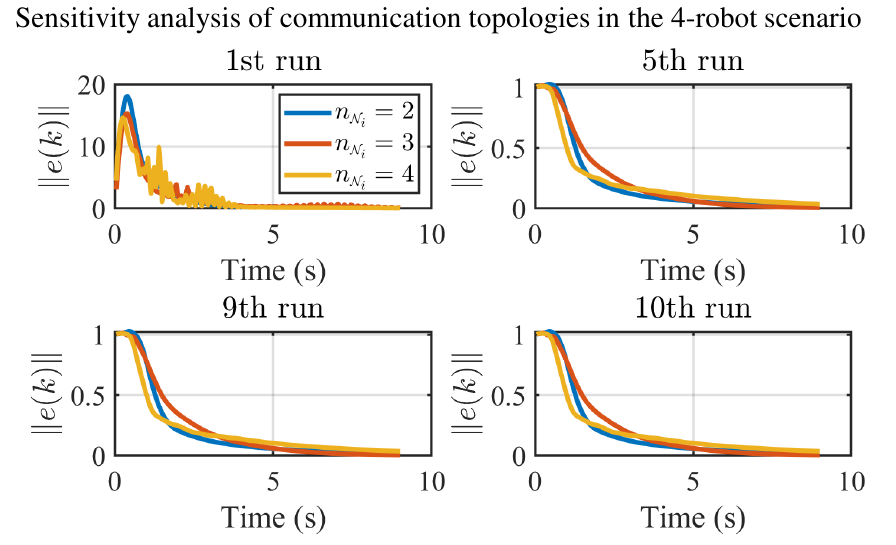} \caption{{\color{black}Sensitivity analysis of communication strategies in a four-robot scenario, where $e=(e_1,e_2,e_3,e_4)$ represents the overall state error.}} \label{fig sensitivity} \end{figure}
					
					{\color{black}\subsection{Sensitivity Analysis of Communication Strategies}\label{sec:sensitivity}
						We conducted sensitivity analysis studies in a four-robot scenario, focusing on performance comparison under different communication strategies, particularly with the number of neighbors $n_{\scriptscriptstyle \mathcal{N}_i},\, \forall i\in\mathbb{N}_1^{M}$ being 2, 3, and 4, respectively. We have run 10 tests using these communication strategies, and the results are shown in Fig.~\ref{fig sensitivity}. In the first experimental run for each communication strategy, the weights in the actor and critic were initialized with uniformly distributed random values in the range $[0,\,0.1]$. As shown in Fig.~\ref{fig sensitivity}, the state error under $n_{\scriptscriptstyle \mathcal{N}_i}=3,\,4$ converged more rapidly than the one under $n_{\scriptscriptstyle \mathcal{N}_i}=2$ within the time period $[0,\,\,1\,{\rm s}]$ but then exhibited a slight oscillating behavior. This oscillating behavior occurred because the weights under more connected neighbors were more susceptible to the random behavior of their neighbors. However, when subsequent experiments were performed using the learned weights of the previous one, the oscillating behavior of state errors was eliminated, and a notable acceleration in convergence was observed under the scenario with $n_{\scriptscriptstyle N_i}=4$ neighbors.}
					
				\end{document}